\documentclass[review]{elsarticle}

\usepackage[margin=1in]{geometry}
\usepackage{hyperref}

\usepackage{outlines} 
\usepackage{amsmath} 
\usepackage[dvipsnames]{xcolor} 
\usepackage{amsfonts} 
\usepackage{amsthm} 
\newtheorem{theorem}{Proposition} 
\usepackage{lipsum}  

\usepackage[ruled,vlined,linesnumbered]{algorithm2e}
\IncMargin{0.7cm} 

\SetKwComment{Comment}{// }{}
\SetKwProg{Fn}{Function}{}{}
\SetKwFunction{recursetree}{recursive\_tree\_expansion}%

\usepackage{siunitx}

\usepackage{bbm}
\usepackage{float}
\usepackage[labelformat=simple]{subcaption}

\newcommand{\R}{\mathbb{R}}

\newcommand{\regtext}[1]{\mathrm{\textnormal{#1}}}

\newcommand{\defemph}[1]{\textbf{#1}}

\newcommand{\mc}[1]{\mathcal{#1}}

\newcommand{\complement}{\mathsf{c}}

\newcommand{\AOIvertex}{\mc{A}} 
\newcommand{\groundplane}{\mc{G}}

\newcommand{\sats}{\mc{S}}

\newcommand{\shadowVertex}[1]{\mc{C}_{#1}}

\newcommand{\nodepoly}[1]{\lozenge_{#1}}
\newcommand{\leafnode}[1]{\lozenge^{\regtext{leaf}}_{#1}}
\newcommand{\parentnode}[1]{\lozenge^{\regtext{parent}}_{#1}}
\newcommand{\LOSchildnode}[1]{\lozenge^{\regtext{LOS child}}_{#1}}
\newcommand{\NLOSchildnode}[1]{\lozenge^{\regtext{NLOS child}}_{#1}}

\newcommand{\nodeleaf}{\lozenge_{\regtext{leaf}}}

\newcommand{\nooverlap}{No Shadow Overlap}
\newcommand{\partoverlap}{Split Overlap}
\newcommand{\fulloverlap}{Full Shadow Overlap}

\newcommand{\nooverlaplower}{no shadow overlap}
\newcommand{\partoverlaplower}{split overlap}
\newcommand{\fulloverlaplower}{full shadow overlap}

\newcommand{\confselect}{\mc{L}}
\newcommand{\confobj}{\mc{O}}
\newcommand{\conflevel}{\gamma}
\newcommand{\confcollect}{\mc{K}_\gamma}
\newcommand{\confcollectnum}[1]{\mc{K}_{\gamma_{#1}}}
\newcommand{\confcollectval}[1]{\mc{K}_{\gamma = {#1}}}

\newcommand{\mzsmraw}{\mc{M}}

\newcommand{\pmatch}[1]{p^{\text{match}}_{#1}}

\newcommand{\gmm}[1]{f_{\text{GMM}, \, {#1}}}
\newcommand{\gmmk}{f_{\text{GMM}, \, k}}
\newcommand{\mzsm}{f_{\mc{M}}}
\newcommand{\percenterror}{\delta_\%}

\newcommand{\aoi}{_{\regtext{\tiny{AOI}}}}

\newcommand{\sat}{_{\regtext{sat}}}

\newcommand{\leaves}{_{\regtext{leaves}}}

\newcommand{\cno}{C/N_0}   
\newcommand{\los}{\regtext{LOS}} 
\newcommand{\nlos}{\regtext{NLOS}} 

\begin{document}

\begin{frontmatter}

\title{Mosaic Zonotope Shadow Matching for Risk-Aware Autonomous Localization in Harsh Urban Environments}

\author{Daniel Neamati, Sriramya Bhamidipati and Grace Gao}
\address{Stanford University}




\begin{abstract}
Risk-aware urban localization with the Global Navigation Satellite System (GNSS) remains an unsolved problem with frequent misdetection of the user's street or side of the street. Significant advances in 3D map-aided GNSS use grid-based GNSS shadow matching alongside AI-driven line-of-sight (LOS) classifiers and server-based processing to improve localization accuracy, especially in the cross-street direction. Our prior work introduces a new paradigm for shadow matching that proposes set-valued localization with computationally efficient zonotope set representations. While existing literature improved accuracy and efficiency, the current state of shadow matching theory does not address the needs of risk-aware autonomous systems. We extend our prior work to propose Mosaic Zonotope Shadow Matching (MZSM) that employs a classifier-agnostic polytope mosaic architecture to provide risk-awareness and certifiable guarantees on urban positioning. We formulate a recursively expanding binary tree that refines an initial location estimate with set operations into smaller polytopes. Together, the smaller polytopes form a mosaic. We weight the tree branches with the probability that the user is in line of sight of the satellite and expand the tree with each new satellite observation. Our method yields an exact shadow matching distribution from which we guarantee uncertainty bounds on the user localization. We perform high-fidelity simulations using a 3D building map of San Francisco to validate our algorithm's risk-aware improvements. We demonstrate that MZSM provides certifiable guarantees across varied data-driven LOS classifier accuracies and yields a more precise understanding of the uncertainty over existing methods. We validate that our tree-based construction is efficient and tractable, computing a mosaic from 14 satellites in 0.63 seconds and growing quadratically in the satellite number.
\end{abstract}


\end{frontmatter}



\section{Introduction}
\label{sec:introduction}

A growing number of autonomous transportation systems are being developed for safety-critical urban applications~\cite{zhang2017current}, including traffic management, lane keeping, ride-share services, cargo transport, and medical-aid delivery.
With recent advances in sensor technologies and computing resources, there is an increasing interest in using Artificial Intelligence~(AI)-driven tools for localization of autonomous vehicles, such as self-driving cars and unmanned aerial vehicles~(UAVs)~\cite{ma2020artificial, xin2019review,anderson2005ai}. 
Specifically, AI tools are popular in localization frameworks that rely on data-driven classifiers to identify distinct features in sensor measurements.
For example, AI tools have been successful in object detection and classification for computer vision~\cite{Huang_2018} and, similarly, for satellite LOS or NLOS classification for Global Navigation Satellite System~(GNSS)~\cite{Xu2018classifier,ruan2020learning}.
However, a lack of widespread risk-aware and certifiable localization inhibits the AI decision-making agents~(in this case autonomous systems) from making optimal decisions in safety-critical settings, thus leading to potentially catastrophic consequences \cite{Abdar_2021}.

Of many possible sensor choices, GNSS receivers have emerged as a standard choice for the outdoor localization of many autonomous systems since these receivers provide absolute user location~\cite{misra2006global}.
However, many tall buildings in close proximity form an urban canyon that degrades the performance of GNSS-based localization, often incorrectly locating the user's street or side of the street.
Specifically, tall buildings block, reflect, scatter, and diffract the direct Line-of-Sight (LOS) GNSS signals, thus inducing Non-Line-of-Sight~(NLOS) and multipath effects~\cite{zhu2018gnss}.
The broad field of 3D Map-Aided GNSS~(3DMA-GNSS) uses 3D building maps and LOS classifiers to detect, mitigate, and correct the NLOS and multipath effects that degrade GNSS performance~\cite{groves2011shadow,miura2015gps,groves2019performance}. 
A 3D building map encodes the 3D structure of the city's infrastructure and a LOS classifier analyzes signal properties to predict the probability with which the user is in line of sight of a satellite.
However, the complexity of the real-world impacts both the realism of 3D building maps and the accuracy of LOS/NLOS classifiers.
For example, differences in the material of the building's facade, which are difficult to include in the 3D building map, change the way that the GNSS signals reflect, scatter, and diffract off the building \cite{groves2015gnss, Adjrad2019performance2nd}. 
The local environment also changes across seasons and with dynamic obstacles.
Various methods augment 3D building maps with digital surface models, antenna hardware advances, on-board imaging, satellite imagery, and LiDAR~\cite{chen2008shaping,chen2021reflection,dostal2018photogrammetric,bai2020using}.
However, current 3DMA-GNSS works struggle to meet the computational efficiency needed to include these real-time 3D building map augmentations on board.
New data-driven/AI-driven LOS classifiers employ decision trees, support vector machines, or neural networks on many signal features including carrier-to-noise density ratio ($\cno$) and pseudorange residuals to estimate the probability that the user is in line of sight of the satellite \cite{Xu2018classifier, xu2020machine}. 
However, environmental complexities make classification difficult, with past data-driven works reporting accuracies of roughly 80\% to 90\%  \cite{Xu2018classifier, xu2020machine}.

Advancements in computational capabilities and increasing numbers of GNSS signals have steadily helped improve both the 3D building map processing and AI-driven LOS classifier accuracy.  
The resources for computation processing (e.g., RAM) and storage memory (e.g., flash memory) have increased considerably since 3DMA-GNSS was first proposed. 
In 2013, authors used previous top-of-the-line smartphones, such as the Samsung Galaxy S3 with up to 2 GB RAM and 16 GB to 64 GB flash memory \cite{Wang2013samsungbuildingboundaries}. 
In comparison, the recent Samsung Galaxy S22 Ultra has up to 12 GB RAM and 128 GB to 1 TB flash memory. 
On-board processing on autonomous systems, such as cars and drones, has correspondingly improved. 
In 2013, authors from \cite{Wang2013samsungbuildingboundaries, Wang2013scoringbuildingboundaries} primarily used GPS and GLONASS signals from one frequency band (e.g., GPS L1 C/A signal). 
More modern antennas in phones and autonomous systems now use multiple recently-available constellations, such as BeiDou and Galileo, and multiple frequencies, such as the GPS L1 and L5 signals~\cite{Ng2021dualfreq}. 

At the methodological level, there are two primary categories of 3DMA-GNSS: ray-tracing and shadow matching. That said, we also acknowledge that recent end-to-end machine learning~(ML) methods from Google~\cite{van_diggelen_android_improving_urban_gps} take a fundamentally different approach to 3DMA-GNSS, though this approach has not exhibited much available literature and we cannot assess its risk-awareness. 
The first primary category is ray-tracing, which considers a large sample of possible signal paths between the user and the satellite.
By computing possible scattering and reflections off buildings, ray-tracing uses the computed distribution in signal path length to recover a corrected GNSS signal.
For a sufficient number of samples, ray-tracing provides an accurate and precise understanding of the distribution of possible signals and the associated uncertainties \cite{miura2015gps,Suzuki2012,miura2013gps}. 
However, ray-tracing is too computationally expensive even for a single satellite and cannot be used in real-time applications without incurring large server overhead. 

The second primary approach to 3DMA-GNSS is shadow matching.
Grid-Based Shadow Matching (GBSM) is the current state-of-the-art for shadow matching and is often mixed with 3DMA GNSS ranging in the Intelligent Urban Positioning~(IUP) methodology \cite{Wang2015probabilistic, Adjrad2016IUP, Adjrad2018IUP}. 
GBSM starts with a grid overlaid over the entire city at a fixed resolution where 
each grid point represents a possible receiver location. 
GBSM derives an offline building boundary look-up table for each grid point that maps elevation and azimuth with prespecified angular discretization to satellite visibility. Specifically, a satellite would be visible at a given elevation and azimuth if no buildings in the 3D building map would block that satellite's LOS signal.
The look-up table method provides online efficiency for a fixed map across large numbers of satellites and has been consistently demonstrated in real-time settings \cite{groves2019performance, Wang2013samsungbuildingboundaries, Wang2015probabilistic, Adjrad2016IUP,Adjrad2018IUP, Groves2018gridfilter, Zhong2021gridfilter}. 
Unlike ray-tracing, shadow matching can provide localization even when the user does not receive a GNSS signal from a satellite (e.g., if the buildings fully blocked the satellite). 
Works such as \cite{Xu2018classifier, xu2020machine} have successfully incorporated AI-driven classifiers with the shadow matching building boundary look-up tables to improve localization. Hence, shadow matching has been a successful method to improve urban GNSS accuracy in real-time.

However, urban GNSS localization accuracy is not the only objective.
Safety-critical autonomous systems require further guarantees on the urban localization solution to ensure the understanding of risk necessary for downstream AI decision-making \cite{Abdar_2021}.
From a bigger picture outlook, this gap is not unique to GNSS localization.
Authors in \cite{Abdar_2021} present a thorough survey of uncertainty quantification in AI and learning-based tools. 
A large volume of work has addressed uncertainty for single specific components such as uncertainty quantification in computer vision or reinforcement learning.
However, we must also provide the theoretical tools for transitioning uncertainty quantification in perception systems to AI decision-making agents in a risk-aware manner.
As throughout risk-aware AI, the push toward risk-aware urban GNSS localization requires new theoretical developments.
To reach risk-aware 3DMA-GNSS, one must meet the following two objectives:
(1)~formulate precise models of the uncertainty in real-time while tractably scaling with increasing numbers of available signals and
(2)~provide guarantees on the uncertainty bounds of the urban localization solution across a vast space of AI-driven LOS classifier designs.


Several works in GBSM have considered the implications of these risk-aware objectives on 3DMA-GNSS.
In \cite{Wang2013samsungbuildingboundaries, Wang2013scoringbuildingboundaries}, the authors discussed their method to compute the building boundaries, which can compute a 1-meter grid resolution for a 500 m by 500 m AOI in less than 4 days. 
This method of computing building boundaries cannot adapt to real-time changes. 
Moreover, with the grid-based approach, one cannot assess if two highly weighted grid points use the same or different satellite visibility information.
This raises the concern that a fixed number of top-scoring output grid cells could contain identical misclassification, thereby conflating the shadow matching uncertainty with the grid resolution.
GBSM also did not provide self-monitoring on the initial localization. 
Instead, they assumed large areas of interest to begin processing and temporally filter from the initial localization assuming a lock on the user location.
While many recent works in IUP use a grid-based filter to combine information and handle the highly non-Gaussian uncertainties in urban localization, many of the output statistics use a weighted average approach with an implicit Gaussian approximation. 
Lastly, the fixed grid spacing and fixed elevation/azimuth spacing enforce a fixed precision of the overall GBSM approach. 
Improving the precision requires finer grid spacing and finer elevation/azimuth spacing which come at an increased computational cost. 
To the authors' knowledge, GBSM cannot yet provide certifiable guarantees on the user uncertainty bounds. 
In our results (Section~\ref{sec:experiments}), we illustrate the how the lack of risk-awareness in GBSM can impede an accuracy and precision understanding of the shadow matching uncertainty.

\begin{figure}
    \centering
    \includegraphics[width=0.6\textwidth]{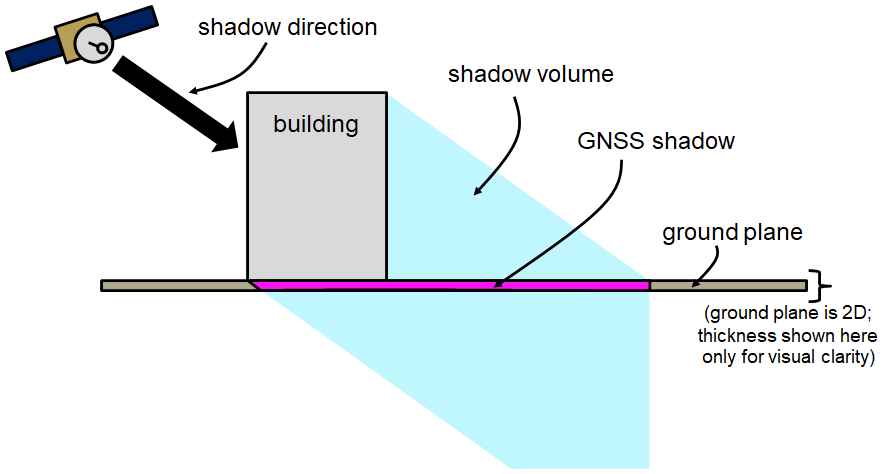}
    \caption{Our prior work on ZSM introduced set-based shadow matching, which efficiently extracts a GNSS shadow~(magenta) using set-valued representations, namely constrained zonotopes~\cite{bhamidipati2021set}. 
    We denote the building in gray, the shadow direction with the black thick arrow, the shadow volume in cyan, the ground plane in brown, and the extracted GNSS shadow in magenta. 
    ZSM recursively refines the area of interest with GNSS shadows, but assumes an ideal LOS classifier, and is thus brittle against any LOS classifier inaccuracies. 
    (Figure adapted from~\cite{bhamidipati2021set}).}
    \label{fig:GNSSshadow}
\end{figure}
Our recent work proposes a new shadow matching paradigm: set-based shadow matching \cite{bhamidipati2021set}. 
A set-valued approach computes a continuum of state estimates rather than considering a grid of position candidates. 
In related set-based works, authors use an initial set of states and measurement bounds to provide certifiable guarantees on the state estimation by checking if the final estimated set lies within user-specified safety bounds~\cite{shi2014set,scott2016constrained_zonotopes}. 
In literature~\cite{shi2014set,althoff2014online,combettes1993foundations}, many set-valued representations exist for parameterizing the states.  
The zonotope is a widely popular set-representation in robotics for computing reachable sets of dynamical systems that enable formally-verified path planning and collision avoidance~\cite{althoff2014online}. 
Zonotopes are convex, symmetrical polytopes that can propagate set-valued state estimates using fast vector concatenation operations. 
Furthermore, constrained zonotopes generalize zonotopes to represent any convex polytopes, thereby avoiding the limitation of symmetry~\cite{scott2016constrained_zonotopes}. 
Our prior work~\cite{bhamidipati2021set} developed a novel Zonotope Shadow Matching~(ZSM) algorithm that computes the set-based GNSS shadow based on the 3D building map, as shown in Figure~\ref{fig:GNSSshadow}. 
ZSM parameterizes each satellite/building pair with constrained zonotopes, extends the building along shadow direction from the GNSS satellite, and performs set intersection with the ground plane. 
Thereafter, this ZSM work iteratively refines an area of interest~(AOI) with an ideal LOS classifier to judge if the receiver is inside or outside the GNSS shadow; if outside, set subtraction is executed, otherwise set intersection.
While this work demonstrates great success in achieving positioning accuracy with high certainty for various simulation scenarios, there still exists a major shortcoming: ZSM is brittle to misclassification. 
In iteratively processing the GNSS shadows, even a single incorrect LOS/NLOS classification can lead to divergence of the algorithm, resulting in incorrect localization, i.e., a null set for the user position.
Simply, ZSM alone is not risk-aware.

In this work, we extend the set-based shadow matching paradigm to allow for risk-aware 3DMA-GNSS.
Specifically, we propose \defemph{Mosaic Zonotope Shadow Matching (MZSM)}, which uses the shadow extraction portion of ZSM as a front-end processing unit and a novel polytope mosaic as a back-end risk-aware localization unit.
Effectively, we use the portion of ZSM that extracts set-based GNSS shadows from the GNSS satellite position and 3D building map. 
Our novel polytope mosaic stitches the GNSS shadows and their complements~(the rest of the AOI not within GNSS shadow) with the LOS probabilities from the AI-driven LOS classifier into a risk-aware framework.
MZSM exhaustively considers all possible classifier misclassifications 
to guarantee uncertainty bounds on the output set-based localization solution. 
To ensure that this exhaustive mapping of classifier misclassifications is tractable and efficient, we employ a memory-efficient binary tree that grows quadratically with the number of satellites rather than the exponential growth of fully-branching binary trees. 
To ensure MZSM can operate with off-the-shelf AI-driven classifiers and adapt to future classifier designs, we take a novel classifier-agnostic approach.
Lastly, we show how our tree-based architecture can predict the expected value shadow matching distribution based on the classifier accuracy statistics.
This prediction capability will allow AI decision-making agents to consider the expected uncertainty on the shadow matching distribution in advance, which enables risk-awareness in downstream autonomous planning.
Altogether, we aim to further the state-of-the-art in shadow matching to include risk-aware urban localization via MZSM.

\subsection{Contributions and Paper Organization}
Our key contributions are as follows: 
\begin{enumerate}
    \item We 
    develop the theoretical foundations for a polytope mosaic 
    that sequential processes LOS classifier probabilities and 
    GNSS shadow polytopes for each GNSS satellite
    to provide classifier-agnostic certifiable guarantees on the user position.
	\item We design a recursively expanding binary tree that performs set operations to bifurcate the area of interest into smaller polytopes at each tree level, and thereafter, probabilistically weighs each tree branch based on the LOS classifier probability. 
	To ensure tractability, we special case different shadow overlap conditions to produce a memory-efficient binary tree.
	Our method computes an exact shadow matching distribution from which we determine a confidence collection of smaller polytopes that collectively satisfies a user confidence level.
	\item We produce an analysis method to estimate the expected shadow matching distribution from a classifier's accuracy statistics (more broadly, its uncertainty quantification statistics).
	We demonstrate that this analysis method arises from the information organization in our binary tree and the set-based localization in our mosaic. We discuss how a system designer and an AI decision-making agent can leverage our analysis method to forecast the shadow matching mosaic in design or online planning. 
	\item We experimental validate our risk-aware shadow matching focusing on the improvements of classifier-agnostic evaluation, set-based bounds, and efficiency with tree-based processing. 
\end{enumerate}

The remainder of the paper is organized as follows.
Section~\ref{sec:proposed_method} describes our proposed MZSM implementation through the tree-based MZSM construction. 
We end Section~\ref{sec:proposed_method} with a discussion of set-based uncertainty bounds and an illustrative example. 
We validate our key risk-aware contributions of MZSM in Section~\ref{sec:experiments} via simulated experimental results.
Finally, Section~\ref{sec:conclusions_and_future_work} provides concluding remarks.

\section{Proposed Mosaic Zonotope Shadow Matching Algorithm} \label{sec:proposed_method}



\begin{figure}
    \centering
    \includegraphics[width = 0.8\textwidth, trim={0cm 1.1cm 2.1cm 0cm}, clip]{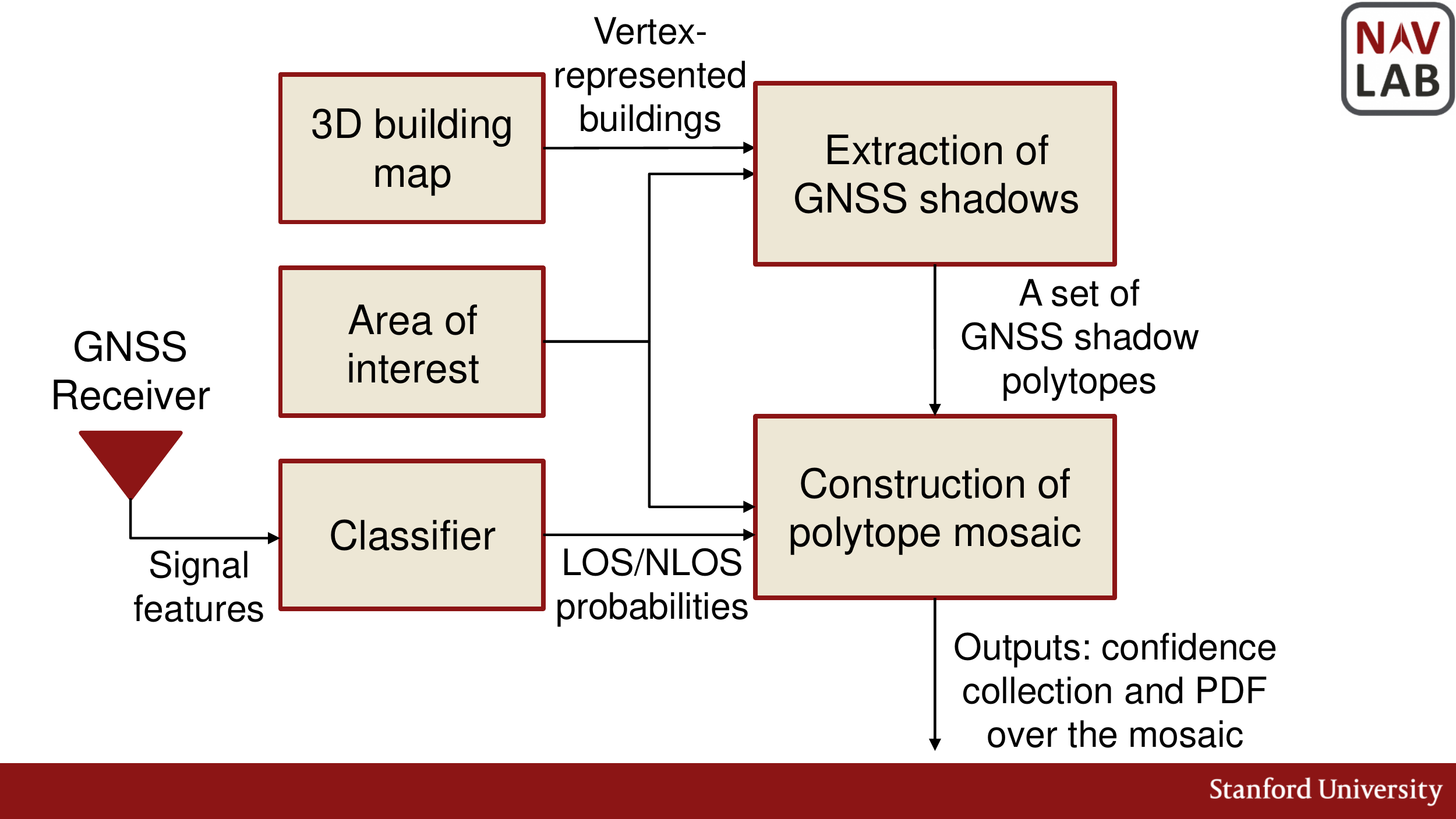}
    \caption{Our MZSM algorithm with two key modules: a) extraction of GNSS shadows, which is based on our prior work~\cite{bhamidipati2021set}; and b) construction of a polytope mosaic. 
    We take the 3D building map, the area of interest, and the known GNSS satellite positions to compute a set collection of GNSS shadow polytopes. 
    We construct our polytope mosaic by taking in the area of interest, the LOS classifier probabilities, and the set of GNSS shadow polytopes. 
    Our MZSM provides risk-aware 3DMA-GNSS localization, wherein the outputs include a confidence collection set of receiver positions that collectively satisfy a user confidence level and a PDF of user position over the mosaic conditioned on the AOI.
    }
    \label{fig:methodsflowchart}
\end{figure}

We first describe our model and assumptions of the GNSS satellite signals, the initial area of interest, and the 3D building map. 
We then summarize the procedure for extracting GNSS shadows, which is based on our prior work~\cite{bhamidipati2021set}. 
We finally explain the details of our tree data structure that takes in the GNSS shadows as input to construct the exact shadow matching distribution (i.e., no discretization), which provides certifiable guarantees on the receiver position.
Figure~\ref{fig:methodsflowchart} provides a roadmap for the methods. 


\subsection{Notation}
Throughout, scalars are italic lowercase. Sets and sets of sets are in script uppercase. We use a subscript for indexing and a text superscript for direct labeling. Probability is denoted $p(\cdot)$. We use the shorthand nation $\{\cdot_i\}_{i = 1}^n$ to construct a set of $n$ elements where each element is indexed with $i$.

\subsection{System Inputs}
MZSM takes in three inputs: (1) an AOI, (2) GNSS shadow polytopes, and (3) LOS probabilities from the classifier. We briefly discuss each input below.

First, the AOI is an initial (often coarse) bounded estimate of the 2D receiver location. As in past works \cite{bhamidipati2021set, Adjrad2017Terrain}, we use a terrain model from the 3D building map to enforce that the 2D AOI properly accounts for the city topography. 
Mathematically, we parameterize the AOI ($\AOIvertex$) by either a single 2D polytope or a collection of 2D polytopes to match the terrain.
\begin{equation}
    \AOIvertex = \{A_k\}_{k=1}^{n\aoi} \subseteq \groundplane,
\end{equation}
where $A_k$ denotes a single polytope among $n\aoi$ total that form $\AOIvertex$, and $\groundplane$ represents the 2D ground plane. We remove the building footprints from the AOI, effectively leaving the polytopes with holes. 
If no prior estimate is available, we can set the AOI ($\AOIvertex$) equal to the ground plane ($\groundplane$), where $\groundplane$ is considered to be bounded. 
Since the AOI is a coarse estimate, risk-aware shadow matching must self-monitor the AOI to determine if the AOI is consistent with the incoming GNSS information. 
Our MZSM method is the first to incorporate this AOI self-monitoring feature (elaborated upon in Section~\ref{sec:memory-efficient-and-self-monitoring-method} and quantitatively assessed in Section~\ref{sec:selfmonitoring}). 

Second, we compute the GNSS shadows using the 3D building map per the method in our prior ZSM work \cite{bhamidipati2021set}. We briefly summarize the important aspects of ZSM for MZSM. 
Standard 3D city maps usually comprise a collection of 3D vertices. 
We convert the point-based 3D city map to a constrained zonotope city map via triangulation, as described further in \cite{bhamidipati2021set}. 
We assume that the user has access to measurements from $n\sat$ GNSS satellites. 
We denote the observed set of GNSS satellites by $\sats = \{s_j\}_{j=1}^{n\sat}$, where each $s_j \in \R^3$ represents the satellite location and $n\sat$ is the total number. 
For each satellite-building pair, we compute the shadow volume (cyan in Figure~\ref{fig:GNSSshadow}) and project the shadow onto the ground plane. 
We convert the constrained zonotope shadow for each satellite-building pair into a vertex representation and take the union across buildings. 
This yields one GNSS shadow ($\shadowVertex{j}$, magenta in Figure~\ref{fig:GNSSshadow}) as a vertex representation for each satellite~$j$. 
Full details on the shadow extraction algorithm and online processing capabilities are available in~\cite{bhamidipati2021set}. 
Importantly for the mosaic, the GNSS shadows are completely independent of the receiver location. 
In other words, the shadows are fixed with respect to the buildings and satellites such that they are the same for all receivers in the scene and change only as the satellite orbits. 

Third, the user must determine the probability with which they are in the LOS of the satellite or in the satellite's GNSS shadow. 
We leverage LOS/NLOS classifiers to determine the probability $p(s_j = \los)$ that the user is in the LOS of the $j$-th satellite, $\forall s_{j}\in \sats$. 
By extension, we express the probability that the user is in the 2D GNSS shadow as $p(s_j = \nlos)$, where $p(s_j = \los) = 1-p(s_j = \nlos)$. Our proposed algorithm is independent of the choice of classifier utilized, but, in practice, system designer can largely select any classifier from literature~\cite{Xu2018classifier, xu2020machine}.

\subsection{Tree-based processing and expansion from a collection of GNSS shadows}


We now proceed to build a risk-aware polytope mosaic from the AOI denoted by $\AOIvertex$, the set of GNSS shadows denoted by $\{\shadowVertex{j}\}_{j=1}^{n\sat}$, and the LOS classifier probability values denoted by $p(s_j = \los), \,\forall s_{j}\in \sats$. 
We sequentially process the extracted GNSS shadow polytopes~$\shadowVertex{j}$ to split the AOI polytope~$\AOIvertex$ into various smaller polytopes. 
Together the smaller polytopes form a mosaic over the AOI. 
Simply as a polytope mosaic, it is not straightforward to estimate which of the smaller polytopes is the receiver most likely to be in. 
To address this, we design a probabilistic polytope mosaic that sequentially processes the extracted GNSS shadow polytopes~$\shadowVertex{j}$ with the LOS/NLOS probabilities~$p(s_j = \los)$ or $p(s_j = \nlos)$ associated with the $j$-th GNSS satellites. 
Thus, each smaller polytope has a unique identification based on the associated joint LOS/NLOS probabilities from the set of all observed GNSS satellites~$\sats$.

\begin{figure}
    \centering
    \includegraphics[width=\textwidth]{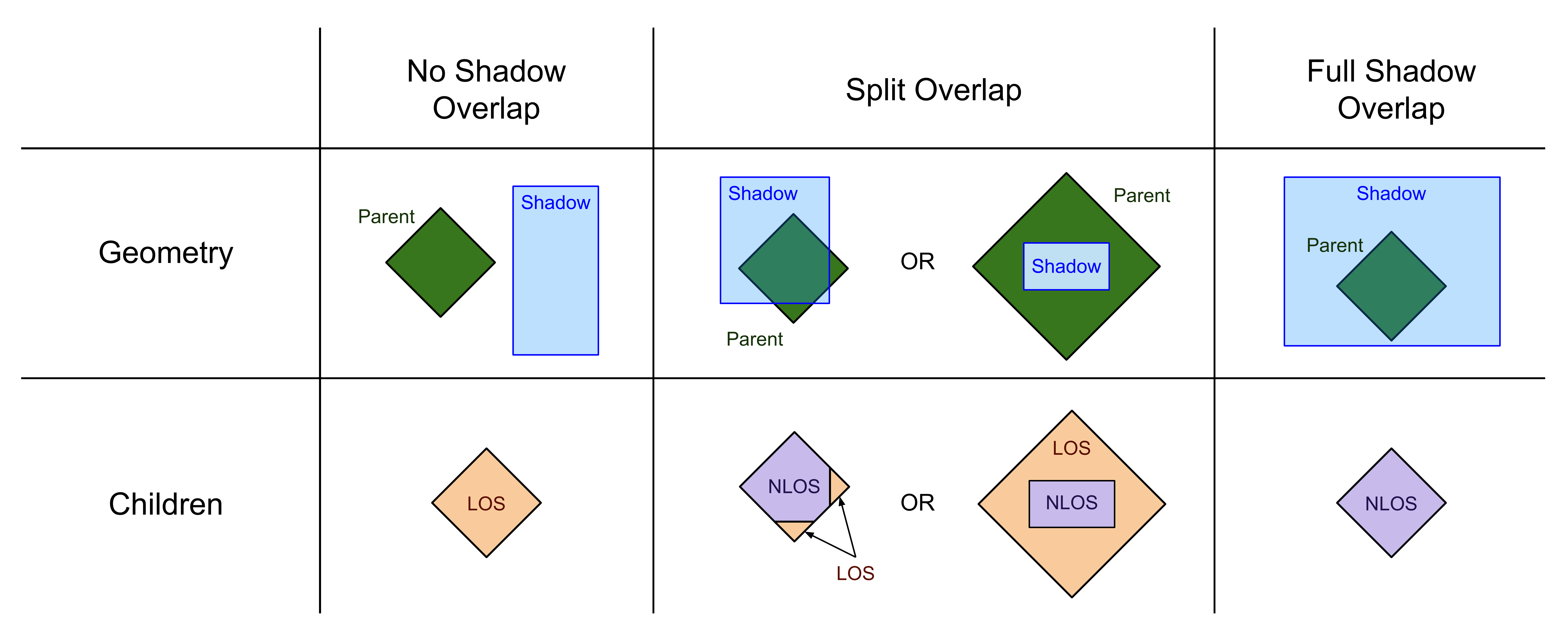}
    \caption{Geometry of the three overlap cases: \nooverlaplower, \partoverlaplower, and \fulloverlaplower. 
    For each case, we illustrate how the overlap between the shadow (blue) with the parent node polytope (green) dictates the children node polytopes. 
    The \nooverlaplower \ case generates an LOS child polytope (orange) but no NLOS child polytope (purple). 
    Equivalently, the NLOS child polytope is the empty set. 
    The \fulloverlaplower \ case generates an NLOS child but no LOS child. 
    Equivalently, the LOS child polytope is the empty set. Only the \partoverlaplower \ case generates non-empty LOS and NLOS children. 
    The \partoverlaplower \ can generate single polytopes (such as the either of the NLOS child polytope), disjoint sets (such as the left LOS child polytope), or sets with holes (such as the right LOS child polytope). } 
    \label{fig:overlap_cases}
\end{figure}


We organize these sequential operations as a recursively-expanding binary tree to connect information across satellites and ensure computational efficiency.
The tree root is the AOI polytope~$\AOIvertex$ with probability $p(\AOIvertex)$ that the user is in the AOI. 
Generally, $p(\AOIvertex) = 1$. We discuss how we monitor if the AOI is violated later in the methods (Section~\ref{sec:memory-efficient-and-self-monitoring-method}).
Each node in the tree encodes a set-based location and associated probability that the user is in that set. Each set-based location is a polytope or a collection of disjoint polytopes, which is collectively denoted by~$\lozenge$. The associated probability is denoted by~$p(\lozenge)$. Each tree branch represents a LOS- or an NLOS-based bifurcation of the parent node polytope into two children nodes. The LOS/NLOS bifurcations and the root AOI uniquely describe a node in the tree. 
A node with two empty subtrees~(no further branching) is called a leaf.
Each new GNSS satellite allows us to expand the tree, such that the maximum depth of the tree equals the number of GNSS satellites~$n\sat$.



When incorporating another GNSS shadow to expand a depth $(j-1)$ tree to the subsequent depth $j$ tree, there are three possible overlap cases between the GNSS shadow and a node polytope. 
These three overlap cases illustrated in Figure~\ref{fig:overlap_cases} are listed as follows: (1) the GNSS shadow and parent node polytope do not overlap, (2) the GNSS shadow either partially or fully lies within the parent node polytope (thereby causing a split in the parent node polytope), and (3) the parent node polytope fully lies within the GNSS shadow. 
We can express the overlap conditions mathematically via Eqs.~\eqref{eq:no_overlap}-\eqref{eq:full_overlap} where we label 
$\nodepoly{j-1,n}$ as the $n$-th node polytope at the $(j-1)$-th tree level. The $0$-th node polytope at the $0$-th tree level represents the tree's root, and thus~$\nodepoly{0,0}$ equals $\AOIvertex$.

\begin{align}
    \text{\nooverlap:} \quad& \nodepoly{j-1,n} \cap \shadowVertex{j} = \emptyset \label{eq:no_overlap} \\
    \text{\partoverlap:} \quad& (\nodepoly{j-1,n} \cap \shadowVertex{j} \neq \emptyset) \text{ and } (\nodepoly{j-1,n} \cap \shadowVertex{j} \neq \nodepoly{j-1,n}) \label{eq:partial_overlap} \\ 
    \text{\fulloverlap:} \quad& \nodepoly{j-1,n} \cap \shadowVertex{j} = \nodepoly{j-1,n}  \label{eq:full_overlap} 
\end{align}
Note that the condition specified for full shadow overlap in Eq.~\eqref{eq:full_overlap} essentially implies~$ \nodepoly{j-1,n} \subseteq \shadowVertex{j}$. These three cases provide a distinct and complete (though non-unique) categorization of all possible overlap cases (\textbf{Proposition}~\ref{theo:propositionA}).


\begin{theorem} \label{theo:propositionA}
In branching any $(j-1)$-th tree level to the next based on the $j$-th GNSS shadow~$\shadowVertex{j} \neq \emptyset$, the outcome of overlaying $\shadowVertex{j}$ onto any parent node polytope~$\nodepoly{j-1,n} \neq \emptyset,~\forall n,$ in the $(j-1)$-th tree level can be uniquely characterized as either (1) \nooverlaplower, (2) \partoverlaplower, or (3) \fulloverlaplower. That is, these cases are distinct and complete.
\end{theorem}
\begin{proof}
See \ref{app:AppendixA}.
\end{proof}

As a result of the overlap between a GNSS shadow and any position node polytope designed as a \texttt{parent}, we generate a maximum of two \texttt{children} node polytopes in the next tree level. 
One child, designated as \texttt{LOS child}, represents the portion of the parent node polytope that is in the LOS of the satellite. The other child, namely \texttt{NLOS child}, indicates the portion that is not in the LOS of the satellite or equivalently, the portion that is in the satellite's GNSS shadow. 
As in \cite{bhamidipati2021set}, we calculate the LOS case with set subtraction in Eq.~\eqref{eq:loschild} and the NLOS case with set intersection in Eq.~\eqref{eq:nloschild}. 
However, the existence~(not empty) or feasibility~(non-zero probability) of these children node polytopes depend on the overlap condition~(from \textbf{Proposition}~\ref{theo:propositionA}) satisfied by the parent node and the GNSS shadow.
\begin{align}
    \LOSchildnode{j,m} &=\parentnode{j-1,n} - \shadowVertex{j} = \parentnode{j-1,n} \cap \shadowVertex{j}^\complement \label{eq:loschild} \\
    \NLOSchildnode{j,l} &= \parentnode{j-1,n} \cap \shadowVertex{j} \label{eq:nloschild}
\end{align}
where $\LOSchildnode{j,m}$, $\NLOSchildnode{j,l}$, and $\parentnode{j-1,n}$ denotes the node polytopes, such that $\LOSchildnode{j,m}$, $\NLOSchildnode{j,l}$ are the children nodes of the parent node~$\parentnode{j-1,n}$. 
Additionally, $\shadowVertex{j}^\complement$ represents the complement of the $j$-th GNSS shadow polytope with $\shadowVertex{j}^\complement=\AOIvertex-\shadowVertex{j}$. Eqs.~\eqref{eq:loschild} and \eqref{eq:nloschild} recursively produce the children from the parent.

The \nooverlaplower \ and \fulloverlaplower \ cases generate empty child nodes.
When the GNSS shadow and the parent node polytope do not overlap, $\parentnode{j-1,n} \cap \shadowVertex{j} = \emptyset$, as seen in Eq.~\eqref{eq:no_overlap} and the \nooverlaplower \ case of Figure~\ref{fig:overlap_cases}. Thus $\LOSchildnode{j,m} = \parentnode{j-1,n}$ and $\NLOSchildnode{j,l} = \emptyset$. 
On the other hand, when the parent node polytope fully overlaps with the GNSS shadow polytope, $\parentnode{j-1,n} \cap \shadowVertex{j} = \parentnode{j-1,n}$, as seen in Eq.~\ref{eq:full_overlap} and the \fulloverlaplower \ case of Figure~\ref{fig:overlap_cases}. Thus, $\LOSchildnode{j,m} = \emptyset$ and $\NLOSchildnode{j,l} = \parentnode{j-1,n}$. 
During a split overlap case seen in Eq.~\eqref{eq:partial_overlap}, both children node polytopes exist as long as $\parentnode{j-1,n} \neq \emptyset$ and $\shadowVertex{j} \neq \emptyset$. 
Intuitively, the empty children nodes in any of these two cases imply redundant or inconsistent measurements. 
For instance, consider the \nooverlaplower \ case at any $j$-th tree level. Given that the receiver lies within the parent node polytope, no overlap means that all locations within the parent node polytope are in LOS of the $j$-th GNSS satellite.
The LOS child node thereby reflects redundant information while the NLOS child reflects incompatible information. 
The reverse relationship holds true for \fulloverlaplower, i.e., the LOS child node reflects incompatible information, and the NLOS child node reflects redundant information. 

We assign a probability score to the children corresponding to $p(s_{j} = \los)$ for the LOS child node polytope and $p(s_{j} = \nlos) = 1 - p(s_{j} = \los)$ for the NLOS child node polytope. 
Starting with a pre-defined value of $p(\parentnode{0,0}) = p(\AOIvertex)$, we arrive at recursive equations (Eq.~\eqref{eq:losprob}-\eqref{eq:nlosprob}) with conditional independence between the GNSS satellite signals. The probability scores decrease monotonically for nodes deeper in the tree, since probabilities are between zero and one.
\begin{align}
    p\Big(\LOSchildnode{j,m}\Big) &= p(s_j = \los) \cdot p\Big(\parentnode{j-1,n}\Big) \label{eq:losprob} \\
    p\Big(\NLOSchildnode{j,l}\Big) &=  p(s_j = \nlos) \cdot p\Big(\parentnode{j-1,n}\Big)\label{eq:nlosprob}
\end{align}

Starting with the AOI polytope, we sequentially process the GNSS shadows and recursively expand the tree to arrive at a completed binary tree.
The leaves of this tree form a probabilistic mosaic of polytopes.
Our binary tree is commutative, the mathematical proof of which is discussed in \textbf{Proposition}~\ref{theo:propositionB}. 
Intuitively, commutativity implies that we arrive at the same completed binary tree irrespective of the order in which we overlay the GNSS shadows onto the AOI. 
This is a crucial property in ensuring risk-aware 3DMA-GNSS localization since an observation error from the AI-driven LOS classifier in the LOS/NLOS probability of a GNSS shadow processed early in the ordering will not disrupt the tree at later operations. That is, we preserve the conditional independence between GNSS signals.

\begin{theorem}  \label{theo:propositionB}
Given a set of GNSS shadow polytopes~$\{\shadowVertex{j}\}_{j=1}^{n\sat}$ and an AOI polytope~$\AOIvertex$, the leaf polytopes~(node polytopes with no further branching) generated from our completed binary tree are the same for any ordering of the AOI and the GNSS shadows.
\end{theorem}
\begin{proof}
See \ref{app:AppendixB}.
\end{proof}

\subsection{Memory-efficient binary tree for a tractable mosaic}
\label{sec:memory-efficient-and-self-monitoring-method}

During the \nooverlaplower \ and \fulloverlaplower \ cases explained earlier, the NLOS child and the LOS child polytope are empty, respectively. 
Expanding the tree with an empty set as the parent node polytope leads to empty sets as the children nodes as well. 
Mathematically, if $\parentnode{j,n} = \emptyset$, then $\LOSchildnode{j+1,l} = \emptyset - \shadowVertex{k} = \emptyset$ via Eq.~\eqref{eq:loschild} and $\NLOSchildnode{j+1,m} = \emptyset \cap \shadowVertex{k} = \emptyset$ via Eq.~\eqref{eq:nloschild}. 
In \textbf{Proposition}~\ref{theo:propositionC} we demonstrate that the non-empty leaf polytopes are sufficient to produce the complete mosaic over the AOI. 


\begin{theorem} \label{theo:propositionC}
Given a set collection of leaf polytopes, the union of the non-empty leaf polytopes exactly matches the tree's root, i.e., AOI polytope ($\AOIvertex$). 
\end{theorem}
\begin{proof}
See \ref{app:AppendixC}.
\end{proof}

However, the probabilities of the empty parent and empty children are non-zero.
These probabilities represent the possibility that one of the parent's branches is not compatible. The probability is sizeable when the AOI does not include the true receiver location. 
Thus, we do not need to propagate an empty set in the tree from a geometric sense, but we do need to keep track of the total probability of the empty nodes to self-monitor the probability that our AOI is violated. 
We use this to our advantage: we avoid expanding the tree when we detect that a non-empty parent node will yield an empty child polytope. 
Instead of tree expansion, we update the probability score of the parent node with on LOS/NLOS probability of the non-empty child node polytope. 
In particular, we identify the empty children nodes under each overlap condition by assessing the area estimates of the \texttt{LOS child} and \texttt{NLOS child}. 
We then update the probabilities using the same method as Eqs.~\eqref{eq:losprob} and \eqref{eq:nlosprob}, but assigning the probability update to the parent. We demonstrate this self-monitoring capability across classifiers later in Section~\ref{sec:selfmonitoring} (Figure~\ref{fig:violation_probability}).

We primarily exclude the empty sets to maintain a low and tractable memory footprint for the tree. 
For instance, if we divided every parent node polytope into two children nodes~(without excluding empty sets) for each GNSS shadow polytope, the binary tree would grow to have $2^n$ leaf polytopes whose sum of probability scores satisfy $\sum\limits_{n}p(\nodepoly{n\sat,n})=\sum\limits_{n}p(\leafnode{n})=p(\AOIvertex)$. 
Geometrically, the total number of leaf polytopes will be far fewer than the $2^n$ size for a full binary tree if we exclude the empty sets. 
This mathematically follows from the Lazy Caterer's sequence (integer sequence A000124 of OEIS \cite{OEISFoundationInc.}) where the number of polygonal sections formed from $n$ cuts grows quadratically with the number of cuts. 
Intuitively, given a sufficiently deep tree, processing the next GNSS shadow polytope cannot result in the split overlap case for all the leaf nodes. 
Instead, the GNSS shadow will fully overlap many nodes and will not overlap with many other nodes. 
We experimentally validate this quadratic complexity later in Section~\ref{sec:selfmonitoring} (Figures~\ref{fig:leaf_analysis} and \ref{fig:tree_timing}).




Lastly, we must ensure that our memory-efficient binary tree maintains the risk-aware self-monitoring of the full binary tree. 
In our memory-efficient binary tree, the AOI violation probability~(i.e., probability of inconsistent information) is directly calculable from the root and leaf probabilities with Eq.~\eqref{eq:pempty}. 
\begin{equation}
    p_{\emptyset} = p(\AOIvertex) - \sum\limits_{l}p(\leafnode{l}) \label{eq:pempty}
\end{equation}
where $\sum\limits_{l}p(\leafnode{l})$ denotes the sum of probability scores at the non-empty leaf nodes in the memory-efficient binary tree.
By construction, $p_{\emptyset}$ is the total probability of all the empty leaf nodes in the full tree. 
We can use $p_{\emptyset}$ to diagnose the posterior confidence in our AOI. 
If $p_{\emptyset} \approx 1 - p(\AOIvertex)$, then the mismatch in $p_{\emptyset}$ is more attributed to inaccuracies in the LOS classifier. 
For example, given $4$ GNSS satellite signals with $p(s_j = \nlos) = 0.1~\forall j\in\{1,\cdots,4\}$, the leaf polytope that represents lying within the GNSS shadow polytope four consecutive times has a probability score of $(0.1)^4 = 0.0001$. 
If this position leaf polytope is an empty set~(i.e., area equals zero), then either the AOI is incorrect or at least one of the satellites should have a higher LOS probability. 
Further, if this leaf with four NLOS branches is the only empty leaf polytope in the entire binary tree, then $p_{\emptyset} = 0.0001$. 
On the other hand, if $p_{\emptyset} \approx p(\AOIvertex)$, then the mismatch is more attributed to an incorrect AOI. 
To understand this, let's consider a different case for the same example as earlier with $4$ GNSS satellites. 
Suppose that the position leaf polytope that represents lying within the complement of the GNSS shadow polytope four consecutive times is an empty set. 
However, the probability score of this position leaf polytope exhibits a significant value of $(0.9)^4 = 0.6561$. 
In this case, we are more likely to trust that the GNSS signals were classified correctly as LOS, but the initial AOI polytope did not include the true receiver position. 
If this leaf with four LOS branches is the only empty leaf polytope in the entire binary tree, then $p_{\emptyset} = 0.6561$. 
Hence, at low $p_{\emptyset}$ values the tree's self-monitor is confident in the AOI. At moderate to high $p_{\emptyset}$ values the tree's self-monitor is skeptical of the AOI and the AOI should be enlarged accordingly. 
Therefore, our memory-efficient binary tree maintains a worst-case quadratic complexity, only stores information of non-empty nodes, and features built-in self-monitoring on the AOI root node. 

We summarize the recursive expansion process to build the memory-efficient binary tree in Algorithm~\ref{alg:tree_expansion}.

\begin{algorithm}[ht]
\caption{Memory-Efficient Recursive Binary Tree Expansion (Snapshot)}\label{alg:tree_expansion}
\Fn(){\recursetree{{\normalfont \texttt{node}}$_{k,n}$, $\shadowVertex{j}$, $p(s_j = \los)$}}{
{\tcc*[h]{Inputs: 
(1)~node$_{k,n}$, (2)~a GNSS shadow $\shadowVertex{j}$, and (3)~probability of LOS $p(s_j = \los)$ for GNSS satellite $j$. The mosiac tree node stores: $\nodepoly{k, n}$, $p(\nodepoly{k,n})$, its \texttt{LOS\_child} node, and its \texttt{NLOS\_child} node.
}
}


\If{split overlap between $\nodepoly{k,n}$ and $\shadowVertex{j}$ (Eq.~\eqref{eq:partial_overlap})}{
    \If{{\normalfont \texttt{node}}$_{k,n}$ has no {\normalfont \texttt{LOS\_child}} and {\normalfont \texttt{NLOS\_child}}}{
    \Comment{Expanded tree at this node}
    add \texttt{LOS\_child} with polytope per Eq.~\eqref{eq:loschild} and probability per Eq.~\eqref{eq:losprob} \\
    add \texttt{NLOS\_child} with polytope per Eq.~\eqref{eq:nloschild} and probability per Eq.~\eqref{eq:nlosprob}
    }\Else{
    \recursetree{{\normalfont \texttt{LOS\_child}} of $\nodepoly{k,n}, \ \shadowVertex{j}, \ p(s_j = \los)$} \\
    \recursetree{{\normalfont \texttt{NLOS\_child}} of $\nodepoly{k,n}, \ \shadowVertex{j}, \ p(s_j = \los)$}
    }
}
\ElseIf{no shadow overlap between $\nodepoly{j-1,n}$ and $\shadowVertex{j}$ (Eq.~\eqref{eq:no_overlap})}{
recursively propagate the LOS score ($p(s_j = \los)$) to leaves (no further expansion)
}
\Else{
\Comment{Remaining case is full shadow overlap (Eq.~\eqref{eq:full_overlap})}
recursively propagate the NLOS score ($1 - p(s_j = \los)$) to leaves (no further expansion)
}
\Return $\nodepoly{k,n}$ \Comment{updated mosaic tree node} \label{line:return_tree}
}
\end{algorithm}

\subsection{MZSM Outputs: Shadow Matching Mosaic, AOI-conditioned PMF, and PDF over the Mosaic}
\label{sec:outputs}

The most immediate output of the memory-efficient binary tree is the shadow matching mosaic ($\mzsmraw$, Eq.~\eqref{eq:mosaic_def}), which denotes the set of all $n\leaves$ leaf polytopes in the memory-efficient binary tree. 
\begin{equation}
    \mzsmraw = \{\leafnode{l}\}_{l=1}^{n\leaves} \label{eq:mosaic_def}
\end{equation}
From $\mzsmraw$, we produce an overall probability mass function~(PMF) over the mosaic ($\mzsmraw$) and a separate self-monitor of AOI violation ($p_\emptyset$, Eq.~\eqref{eq:pempty}). 
Mathematically, we can condition the leaf node probabilities $p(\leafnode{l})$ for each $l$-th leaf to the AOI $p(\leafnode{j} \mid \AOIvertex )$ as shown in Eq.~\eqref{eq:AOI_prob_normalization}.
Together, the $p(\leafnode{j} \mid \AOIvertex )$ probabilities from the PMF over $\mzsmraw$.
\begin{equation}
    p(\leafnode{l} \mid \AOIvertex ) = \frac{p(\leafnode{l})}{\sum_{l = 1}^n p(\leafnode{l})} \label{eq:AOI_prob_normalization}
\end{equation}

The user requires a probabilistic guarantee that they are within a bounded region at a desired confidence to ensure risk-aware localization. 
From the mosaic $\mzsmraw$ and the PMF over the mosaic ($p(\leafnode{j} \mid \AOIvertex )$), we can generate two outputs that achieve this goal: (1)~a probability density function over the user location and (2)~a set-based confidence collection. 
To have certifiable guarantees, the latter is the preferred output. 

In certain cases, we may need to work with the user location directly. 
In position space, each leaf polytope is a uniform distribution ($\mathcal{U}_{\leafnode{l}}$) with an irregular shape.
Namely, the position $(x, y)$ only has non-zero probability in the polytope's enclosed area, where $x$ and $y$ are any two orthogonal 2D coordinates (e.g., North and East or cross-street and along-street). 
We normalize the weights to produce a well-defined probability density function~(PDF) over the mosaic
denoted by $\mzsm(x, y)$, per Eq.~\eqref{eq:mzsm_pdf}. 
\begin{equation}
    \mzsm(x, y) = \left(\sum_{l=1}^{n\leaves} \mathcal{U}_{\leafnode{l}}(x, y) \right) \left(\iint_{\text{AOI}} \sum_{l=1}^{n\leaves} \mathcal{U}_{\leafnode{l}}(x, y) \ dx \  dy \right)^{-1} \label{eq:mzsm_pdf}
\end{equation}

In contrast, another output of interest is the confidence collection, explained in the subsequent subsection.

\subsubsection{MZSM Output: Certifiable Guarantees via Confidence Collections of Set-Based Distributions} \label{sec:confidence_collection}

To motivate the set-based confidence collection, notice that a single set in the mosaic is very unlikely to hold the desired user confidence, even with a high accuracy classifier. 
For example, if a classifier outputs $\{p(s_j = \los)\}_{j = 1}^3 = \{0.9, 0.15, 0.95\}$ for three satellites. 
Then, the leaf node matching (LOS, NLOS, LOS) would have a probability of $p(\leafnode{(\los, \nlos, \los)}) = 0.9 \cdot 0.85 \cdot 0.95 = 0.73 $. 
This leaf node probability of 0.73 is much lower than a desired confidence level of 0.95, as an example. 
Instead, we need to form a collection of leaf polytopes that together achieve a user-defined confidence level ($\gamma$, e.g., $\gamma = 0.95$). 
We call the collection at user confidence level $\gamma$ the confidence collection ($\confcollect$), defined in Eq.~\eqref{eq:confcollect}.
We use an index set $\confselect \subseteq \{l\}_{l = 1}^{n\leaves}$ to select the set of leaves in the confidence collection.
\begin{equation}
    \confcollect = \{\leafnode{l}\}_{l \in \confselect} \subseteq \mzsmraw \label{eq:confcollect}
\end{equation}
We intentionally avoid calling it a confidence interval. A confidence interval is well defined for many unimodal distributions, such as a Gaussian distribution. However, the symmetries of buildings in an urban plan result in a multimodal distribution of shadow matching predictions \cite{groves2019performance, groves2015gnss, Adjrad2019performance2nd, Wang2014kinematic}. We can write the process to compute the confidence collection as an optimization problem over the index set $\confselect$ as in Eq.~\eqref{eq:confcollectoptimization}. 
If the user accepts the AOI based on the self-monitor, then we can certifiably guarantee that the user is in the confidence collection to the user's specified confidence with the optimization problem in Eq.~\eqref{eq:confcollectoptimization}.

\begin{equation}
    \begin{aligned}
        \min_{\confselect} \quad & \sum_{l \in \confselect}\confobj(\leafnode{l}) \\
        \textrm{s.t.} \quad & \sum_{l \in \confselect} p(\leafnode{l} \mid \AOIvertex ) \geq \conflevel \label{eq:confcollectoptimization}
    \end{aligned}
\end{equation}


The optimization objective $\confobj$ will depend on the user application and the goals of the system designer.
One simple choice is $\confobj(\leafnode{j}) = 1$, such that the objective function counts the number of leaf nodes in the confidence collection. 
Such an objective function seeks to find the minimal set of leaf nodes that meets the confidence level. 
Intuitively, the minimal set confidence collection obtains the level sets of the shadow matching mosaic conditioned to the AOI. 
Mathematically, for the minimal set confidence collection, $\confcollectnum{1} \subseteq \confcollectnum{2}$ if $\conflevel_1 < \conflevel_2$.
In contrast, more advanced techniques can be explored to compute confidence collections. 
For instance, one can use the mosaic's tree structure to cluster leaf polytopes or even select leaves the minimize the area~(i.e., $\confobj(\leafnode{j}) = \text{Area}(\leafnode{j})$) based on the Knapsack problem \cite{Martello1990knapsack}. 

We use a greedy algorithm to find the minimal set confidence collection.
First, we take the mosaic $\mzsmraw$ and convert it into an array. Second, we sort the array based on descending leaf node probability ($p(\leafnode{j} \mid \AOIvertex)$). 
Third, we sequentially push leaf nodes to the confidence collection $\confcollect$ from the start of the sorted array until we arrive at the required confidence level $\gamma$. 
Since $\sum_{l = 1}^{n\leaves} p(\leafnode{l} \mid \AOIvertex) = 1$, a confidence collection $\confcollect$ always exists for any confidence level $\conflevel \in [0, 1]$.
So, this greedy algorithm is simple, sufficient, and polynomial-time.


\subsection{Illustrative Example of MZSM's Memory-Efficient Tree Construction}
\label{sec:illustrative}

\begin{figure}[H]
    \centering
    \includegraphics[width=0.96\textwidth]{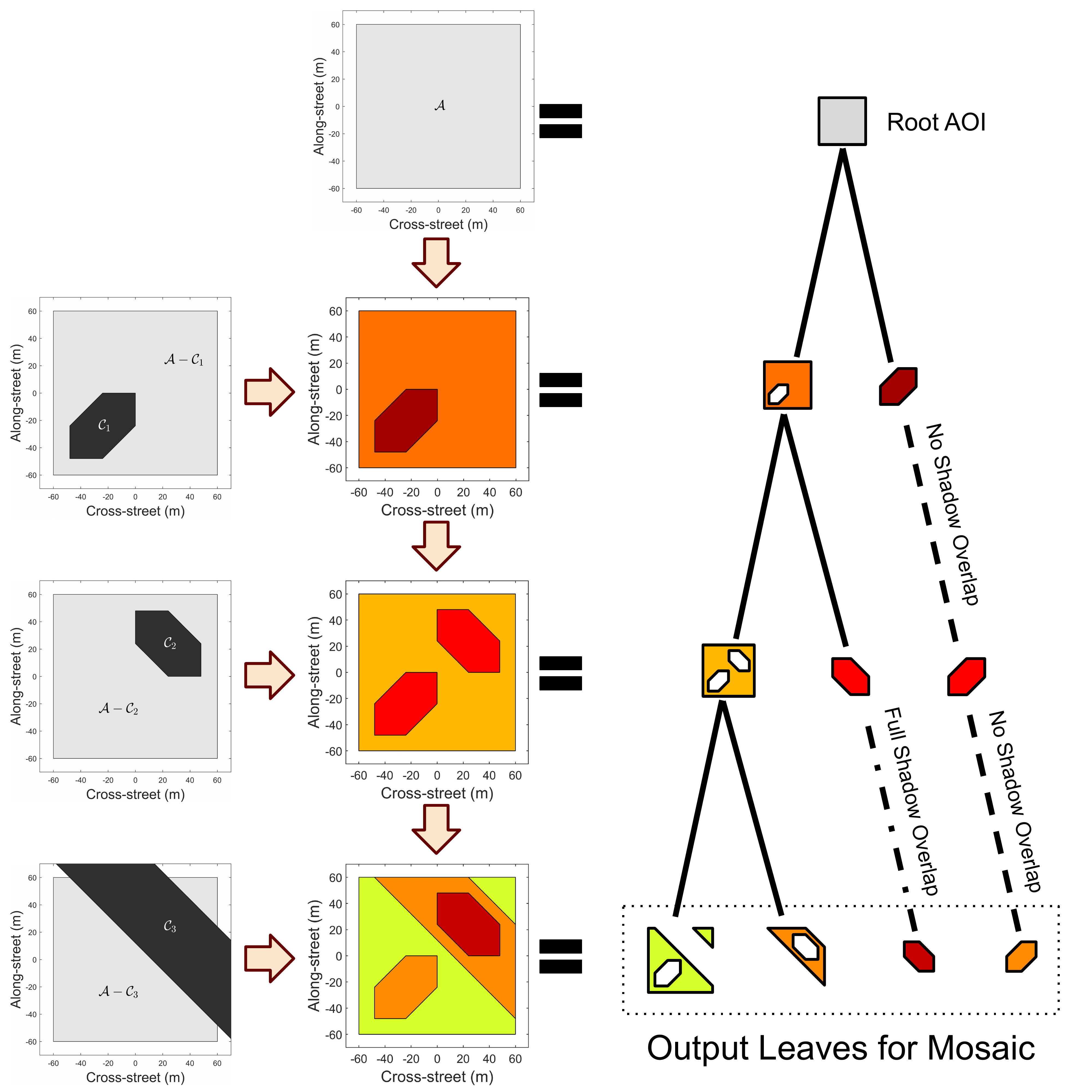}
    \caption{Illustrative example of a recursively-expanding binary tree with three GNSS shadows and an LOS satellite classifier probability $p(s_j = \nlos) = 0.8~\forall j\in\{1,\cdots,3\}$.
    The figures on the far left column show the GNSS shadows $\shadowVertex{1}, \shadowVertex{2}$, and $\shadowVertex{3}$ individually overlaid in black on the AOI $\AOIvertex$~(gray). 
    The middle column shows the polytope mosaic~(breakdown into different node polytopes) constructed after processing each shadow (namely, $\shadowVertex{1}, \shadowVertex{2}$, and $\shadowVertex{3}$ in order) based on its LOS satellite classifier probability.
    The node polytopes at each tree level are colored with the probability mass function (PMF) over the mosaic per Eq.~\eqref{eq:AOI_prob_normalization}.
    The rightmost column represents our polytope mosaic $\mzsmraw$ in terms of a memory-efficient binary tree structure, where the addition of each GNSS shadow splits the node polytope based on one of the three cases discussed in Eqs.~\eqref{eq:no_overlap}-\eqref{eq:full_overlap}, i.e., no shadow overlap~(dashed line), split overlap~(solid line) and full shadow overlap~(dotted dashed line). 
    The final layer showcases our constructed polytope mosaic, wherein we gather the leaves from the tree and plot the leaf polytopes that form the mosaic. 
    }
    \label{fig:annotated-tree-build}
\end{figure}

We illustrate a simple non-physical example of MZSM in Figure~\ref{fig:annotated-tree-build} that includes all overlap cases to build a clearer understanding of our proposed technique. 
The AOI ($\AOIvertex$) is a square of size $60~$\si{\metre}$\times 60~$\si{\metre} oriented in cross-street and along-street directions. 
We consider a total of three GNSS shadows (left side of Figure~\ref{fig:annotated-tree-build}): (1) a hexagonal shadow ($\shadowVertex{1}$) in the bottom left corner of the AOI, (2) a hexagonal shadow in the top right corner of the AOI, and (3) a long-wide hexagonal shadow ($\shadowVertex{3}$) that extends beyond the AOI. 
On the right side Figure~\ref{fig:annotated-tree-build}, we illustrate how we expand the tree as we sequentially process each shadow.
We denote branches with the \nooverlaplower \ case with a dashed line, the \partoverlaplower \ case with the solid line, and the \fulloverlaplower \ case with a dot-dash line.
The tree structure would be different with a different satellite ordering. 
However, the tree always expands with one of the three overlap cases (\textbf{Proposition}~\ref{theo:propositionA}) and always yields the same four leaves (\textbf{Proposition}~\ref{theo:propositionB}).
Lastly, we showcase the progression of the mosaic in the middle of Figure~\ref{fig:annotated-tree-build}, culminating in the tree's output $\mzsmraw$ (Eq.~\eqref{eq:mosaic_def}).
For simplicity, we assign $p(s_j = \nlos) = 0.8~\forall j\in\{1,\cdots,3\}$ and color code the leaves with the probability.
Our memory-efficient binary tree has 3 leaves at the second expansion and 4 leaves at the third expansion. 
As noted in \textbf{Proposition}~\ref{theo:propositionC}, these 4 non-empty leaves fully complete the mosaic.
The full binary tree would have 4 and 8 leaves at the second and third expansions, respectively. 
Even in this simple example, we already see the benefits of our MZSM's memory-efficient binary tree construction. 
From the output mosaic $\mzsmraw$, we can formulate the PDF~(Eq.~\ref{eq:mzsm_pdf}) or the confidence collection~(Eq.~\ref{eq:confcollect}). 
As noted before, a risk-aware autonomous agent, should use the confidence collection to have certifiable guarantees. 
From the confidence collection and self-monitoring probability, a risk-aware AI decision-making agent can properly judge the uncertainty in the scene before continuing its policy.

\section{Experiments and Results} \label{sec:experiments}
We assess the risk awareness of our proposed MZSM algorithm. We begin the section by discussing our two baseline methods. We then detail our simulation setup and move into the experimental results. We conduct four analyses: (1)~performance for different classifiers, (2)~improvements with set-based uncertainty models, (3)~improvements with set-based confidence bounds, and (4)~self-monitoring and efficiency.

\subsection{Baselines}
\label{sec:baselines}

This work is the first set-based risk-aware shadow matching framework, so we cannot directly compare the set-based outputs to other methods. 
Specifically, we tackle two objectives in risk-aware urban localization (Section~\ref{sec:introduction}): 
(1)~accurately modeling the uncertainty and
(2)~providing uncertainty bounds across different classifier designs.
GBSM is the state-of-the-art in urban localization and serves as an intuitive baseline choice. 
However, we need to isolate to what extent the risk-aware improvements from MZSM arise from improved uncertainty modeling versus from improved uncertainty bounds.
Therefore, we use two baselines to match one baseline per objective:
Gaussian mixture models~(GMMs) and set-augmented GBSM~(SA-GBSM). 

First, the GMM baseline isolates the first objective related to the difficulties in handling shadow matching's non-Gaussian uncertainties.
We provide the GMM with a precise understanding of the shadow matching distribution and demonstrate that we cannot provide certifiable guarantees with Gaussian error models since the Gaussian approximation does not hold. 
Specifically, a GMM with one mixture component preserves the Gaussian state error model that is often used in GBSM. 
For example, \cite{Wang2015probabilistic, Adjrad2016IUP, Adjrad2018IUP, Wang2014kinematic} derive a position and
covariance estimate
from the weighted average of top-scoring cells based on an implicit single component Gaussian assumption. 
The multiple mixture component cases allow us to extend beyond the current state-of-the-art GBSM approach and demonstrate more general results about the issues of Gaussian error models in risk-aware urban localization. 
We provide implementation details of GMM in Section~\ref{sec:gmm-details}.

Second, the SA-GBSM baseline allows us to isolate the second objective related to the uncertainty bounds across different classifiers.
We base SA-GBSM on the recent GBSM grid filter \cite{Groves2018gridfilter, Zhong2021gridfilter}, whereby the grid cells have different probability mass and a selection of top-scoring grid cells define the position update.
In SA-GBSM, we augment GBSM to a set-based domain where each cell represents a square area rather than a single point. 
We demonstrate how the grid-based uncertainty bounds in SA-GBSM converge to the exact polytope set-based uncertainty model in MZSM, though only at high grid resolutions that are generally unpractical to compute in real-time.
We provide implementation details of SA-GBSM in Section~\ref{sec:gbsm-details}.

\subsubsection{GMM details}
\label{sec:gmm-details}
To provide a risk-aware analogue of GBSM that preserves the Gaussian uncertainty model, we fit GMMs with different numbers of mixture components.
Since GMM fitting requires fitting to sampled data, 
we use the MZSM tree structure to sample one million points from the exact shadow matching distribution. 
We use the off-the-shelf Matlab GMM fitting function (\texttt{fitgmdist}), which initializes with the \texttt{$k$-means++} algorithm and then optimizes the GMM likelihood with the iterative Expectation-Maximization algorithm~\cite{mathworks2022fitgmm}. 
We use modest regularization to help the algorithm converge at such large numbers of samples. 
We also run five replicates of Expectation-Maximization and select the fit with the best likelihood. 
Generally, the fitting process takes about one second for the 1 mixture component case, about tens of seconds for the 2 or 3 mixture component case, about one minute for the 4 mixture component case, and about one to four minutes for the 5 mixture component case. 

\subsubsection{GBSM and SA-GBSM details}
\label{sec:gbsm-details}
To provide a risk-aware analogue of GBSM that preserves the grid-based localization, we develop SA-GBSM.
SA-GBSM is nearly the same as GBSM, except that we output a set-based localization.

We first discuss the probabilistic model shared between GBSM and SA-GBSM.
GBSM stores a building boundary look-up table for each grid cell. 
GBSM scores each cell based on the probabilistic consistency between that cell and the available satellite LOS or NLOS classification information. 
In particular, the matching probability for the $j$-th satellite ($\pmatch{j}$) is given as Eq.~\eqref{eq:PmatchNg} in works such as \cite{Ng2021dualfreq, Wang2015probabilistic}
\begin{align}
    \pmatch{j} &= p(s_j = \los \mid \cno) p(s_j = \los \mid BB) + [1 - p(s_j = \los \mid \cno)] [1 - p(s_j = \los \mid BB)] \label{eq:PmatchNg}
\end{align}
\noindent where the term $p(s_j = \los \mid \cno)$ is the probability that the satellite is in the user's LOS given the measurement carrier to noise density ratio. This term can be readily switched with a machine learning-based classifier that can use many signal features as illustrated in~\cite{Xu2018classifier, xu2020machine}. 
The second term, $p(s_j = \los \mid BB)$, is the probability that the satellite is in the LOS given the building boundary look-up table. 
In past GBSM works~\cite{Wang2015probabilistic} where real-world experiments and dynamic obstacles were considered, the authors set $p(s_j = \los \mid BB)=0.85$ if the building boundary predicts that the satellite is visible and 0.2 if the building boundary predicts that the satellite would be blocked. 
Intuitively, these constants cap the range of $\pmatch{j}$. 
Such caps make sense since the GBSM building boundaries cannot be adapted in real-time and the 3D building map does not include all the objects in the user's environment \cite{Adjrad2019performance2nd}. 
For our current analysis, we consider a simulated set-up (as detailed in the next section) with an ideal building model (i.e., no dynamic obstacles, etc.), and thus set the constants as 1 and 0, respectively. 
The final score at each grid is 
the product of the matching probabilities across satellites, namely $\prod_j \pmatch{j}$ \cite{Adjrad2016IUP, Adjrad2018IUP}. 
As noted in our methods (Section~\ref{sec:proposed_method}), our MZSM scoring reflects the product of LOS/NLOS classifier probabilities across satellites, which largely mirrors the state-of-the-art GBSM scoring \cite{Adjrad2018IUP}. 

We modify the GBSM framework into SA-GBSM to treat each grid cell as a square with side lengths matching the grid resolution following the motivation for grid filters in recent GBSM work \cite{Groves2018gridfilter, Zhong2021gridfilter}. 
Then, each grid cell's square is a set, and the squares together fully tile the AOI. 
Usually with GBSM, the user must calibrate the number of top-scoring candidates based on the grid spacing, number of satellites, and building geometry. 
With our set-based confidence collections, we 
do not require further calibration once the designer assigns the clustering objective (Section~\ref{sec:confidence_collection}). We use the same minimal set confidence collection objective for both MZSM and SA-GBSM (Section~\ref{sec:confidence_collection}).

\subsection{Simulated Experiment Set-up}
We use a publicly-available 3D building map of San Francisco~\cite{gyorgy2018sfmap}
and we isolate a relevant subsection of the 3D map near the Salesforce tower~(translucent pink box at the bottom left in Figure~\ref{fig:sensitivity_mosiac}), which is a notable landmark in San Francisco. Salesforce tower sits at the intersection of Mission and Fremont Streets, where eight large buildings (translucent pink boxes in Figure~\ref{fig:sensitivity_mosiac}) form an urban canyon. 
We consider the ground truth of receiver position at $(0,-18)~$m in local map coordinates (black circle marker in Figure~\ref{fig:sensitivity_mosiac}) and the AOI ($\AOIvertex$) to be a square of $120~$m$\times 120~$m (black box outline in Figure~\ref{fig:sensitivity_mosiac}) while excluding the building footprints of six buildings that lie within the AOI.
Using the same software simulator set-up \cite{softGNSS,gpssdrsim} as explained in Figure~9 of our prior work~\cite{bhamidipati2021set}, we generate $14$ GPS satellites. 
Based on the relevant subsection of the 3D building map and the chosen ground truth of receiver location, 10 among a total of 14 satellites are expected to exhibit NLOS characteristics. 
Next, we utilize our prior work~\cite{bhamidipati2021set} to perform the following steps: a)~convert the entire 3D building map from their standard vertex representation to constrained zonotope representation, b)~compute the 2D GNSS shadow~(a constrained zonotope) for each satellite/building pair~(from among 14~satellites and 8~buildings) by extending the 3D building constrained zonotope in the “shadow direction” from the satellite to the building, c)~convert the shadow of each satellite/building pair from constrained zonotope to a vertex representation that is used for representing 2D polytopes, and d)~concatenate the vertex-represented regions of 2D GNSS shadows across all the buildings for each satellite, and then convert them to polygons of GNSS shadow~(one for each satellite). From there, we proceed with MZSM as detailed in Section~\ref{sec:proposed_method} and as illustrated in Section~\ref{sec:illustrative}.

\subsection{Assessing and Predicting the Shadow Matching Mosaic for different classifiers}

\begin{figure}
    \centering
    \begin{subfigure}[b]{0.45\textwidth}
         \centering
         \includegraphics[width= \textwidth]{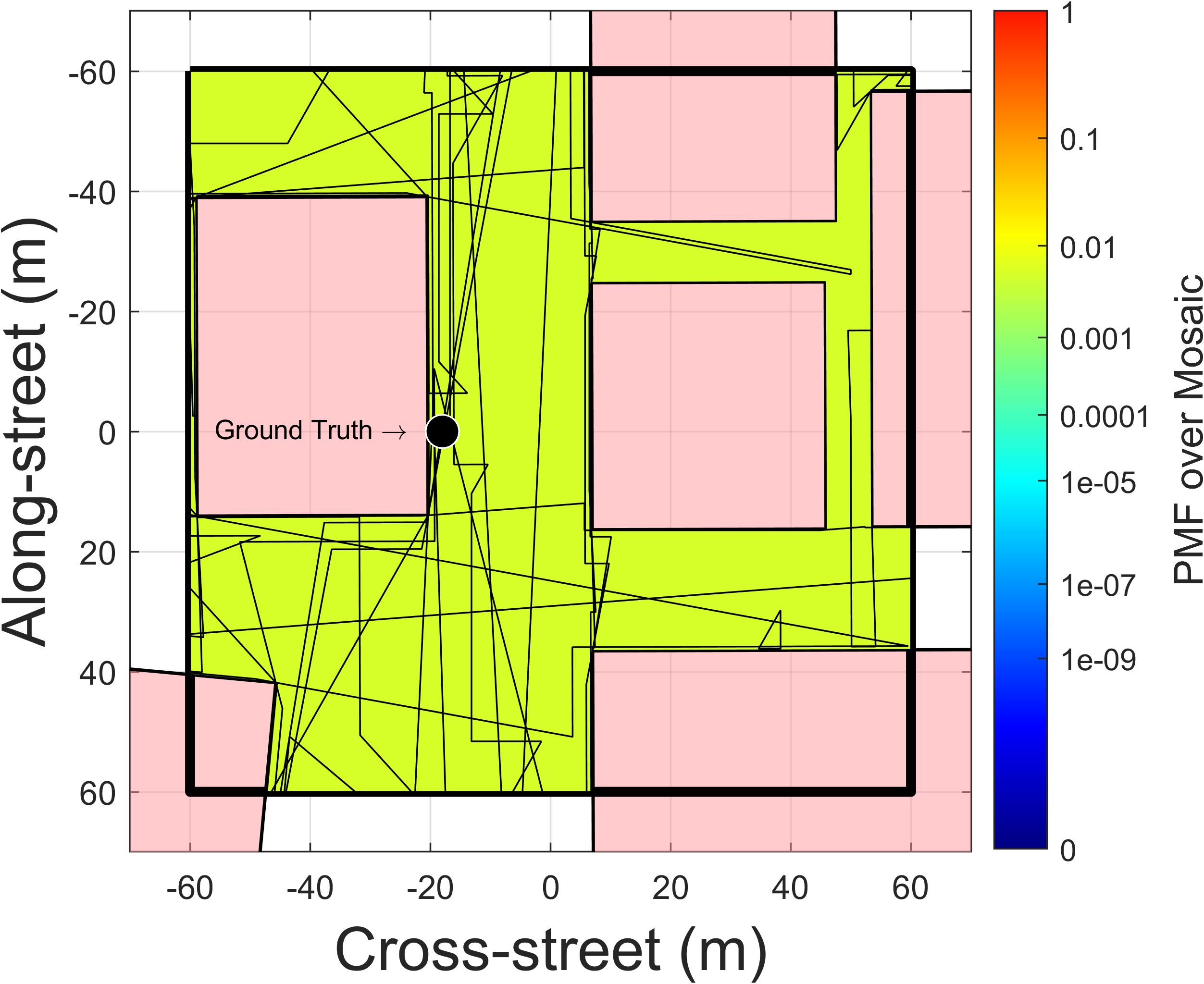}
         \caption{Classifier Posterior of 50\% \\ (Uninformative Classifier)}
         \label{subfig:sensitivity_mosiac_0_50}
     \end{subfigure}
     \hfill
    \begin{subfigure}[b]{0.45\textwidth}
         \centering
         \includegraphics[width= \textwidth]{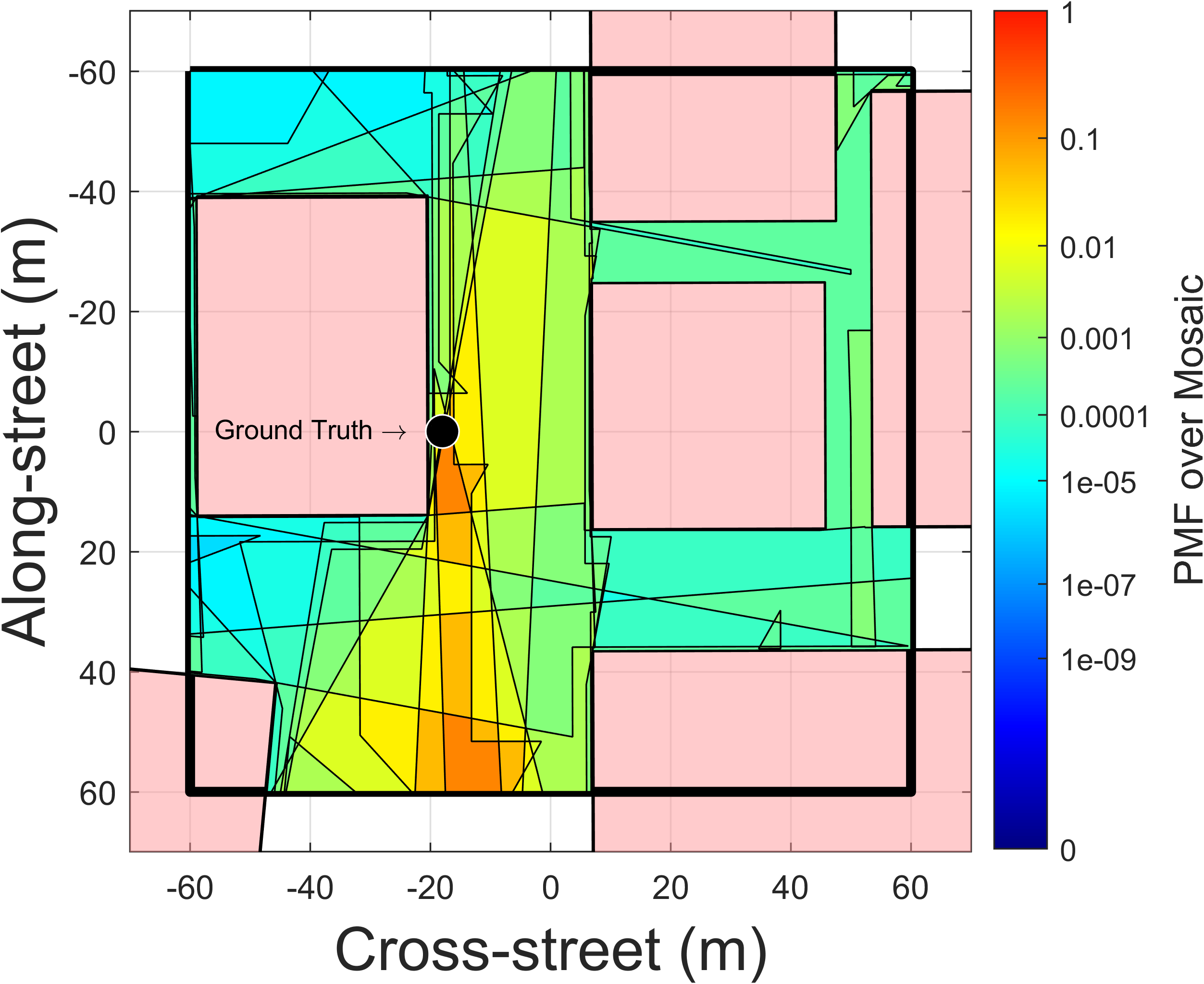}
         \captionsetup{justification=centering}
        \caption{Classifier Posterior of 75\% \\ (Typical $\cno$ Bayesian Classifier)}
     \end{subfigure}
     \par\bigskip
    \begin{subfigure}[b]{0.45\textwidth}
         \centering
         \includegraphics[width= \textwidth]{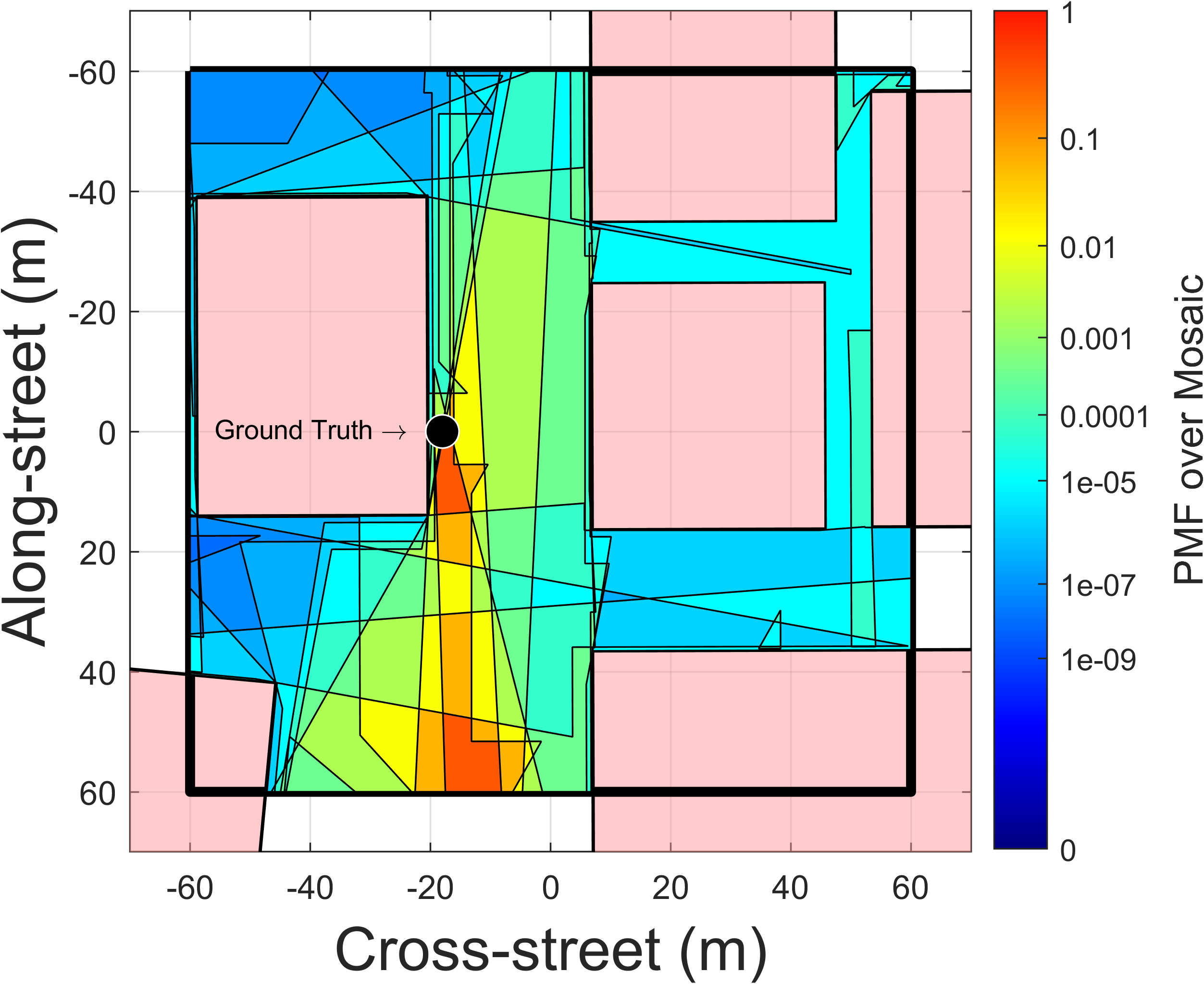}
         \captionsetup{justification=centering}
         \caption{Classifier Posterior of 85\% \\ (Typical SVM Classifiers)}
         \label{subfig:sensitivity_mosiac_0_85}
     \end{subfigure}
     \hfill
    \begin{subfigure}[b]{0.45\textwidth}
         \centering
         \includegraphics[width= \textwidth]{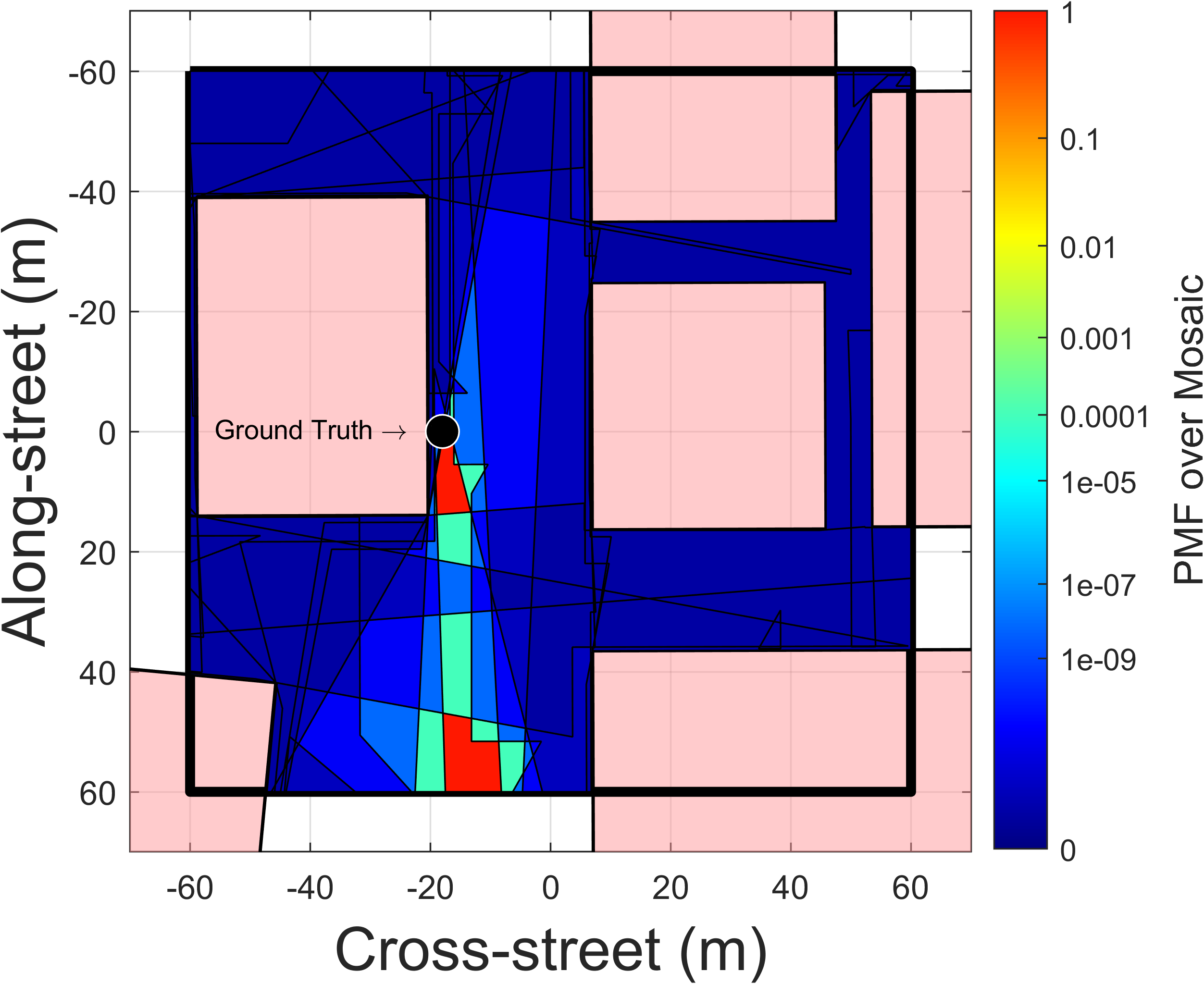}
         \caption{Classifier Posterior of 99.99\% \\ (Near-Perfect Classifier)}
         \label{subfig:sensitivity_mosiac_0_9999}
     \end{subfigure}
     \hfill
     \begin{subfigure}[b]{0.5 \textwidth}
          \includegraphics[width = \textwidth]{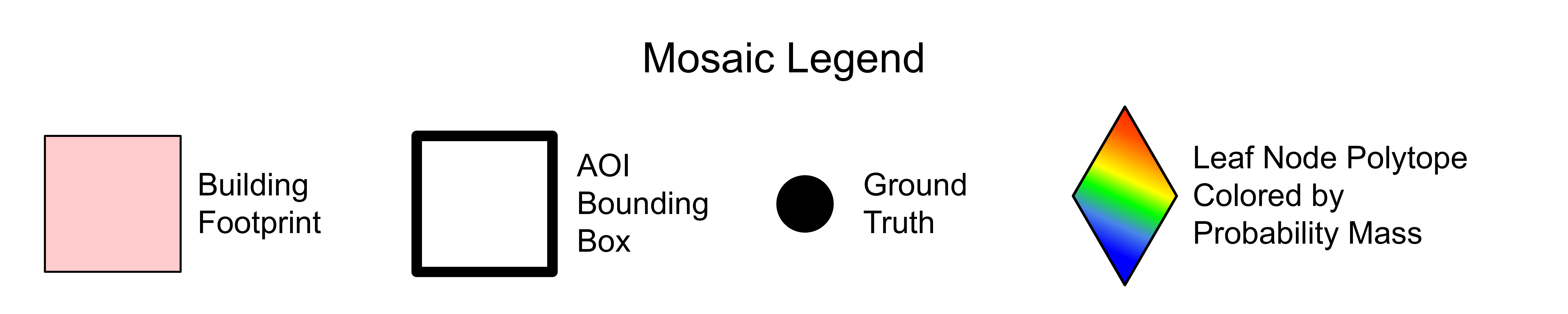}
     \end{subfigure}
    \caption{PMF of our shadow matching mosaic $ p(\leafnode{l} \mid \AOIvertex )$ (as defined in Eq.~\eqref{eq:AOI_prob_normalization}) for four classifier posteriors of 50\%, 75\%, 85\% and 99.99\%, where the classifier true positive and true negative rates are assumed equivalent. 
    The translucent magenta boxes are the building footprints,  
    the thick black box outline is the AOI bounding box, and
    the labeled black circle is the ground truth receiver location.
    Our proposed MZSM validates the presence of two distinct modes in (d) that matches the results from an ideal classifier case discussed in our prior work~\cite{bhamidipati2021set}.
    In contrast, as expected, our MZSM forms a uniform PMF over the entire mosaic using an uninformed classifier in (a), i.e., posterior of 50\%.
    Comparing the classifier posterior case of 75\% in (b) to that of 85\% in (c), we demonstrate a noticeably more precise expected localization with the concentration of the probability mass to fewer leaves.
     }
    \label{fig:sensitivity_mosiac}
\end{figure}

With the vast space of classifiers, it is not practical to evaluate the risk-awareness of our proposed algorithm for each possible classifier. 
Instead, we take the expected value of the shadow matching mosaic~($\mzsmraw$, defined in Eq.~\eqref{eq:mosaic_def}) via the classifier's true positive and negative rates. 
Specifically, in expectation, the classifier's posterior over predictions converge to its true positive rate for LOS detection and its true negative rate for NLOS detection, by definition of the true positive and negative rates. 
So, we can input the true positive and negative rates as the node probabilities into the tree expansion of the LOS and NLOS branches, per Section~\ref{sec:proposed_method}. 
We assume that the true positive rate is equal to the true negative rate throughout the subsequent results, i.e., the true positive rate and true negative rate are symmetric. 
Consistently in the results, we shorthand this approach as ``Classifier Posterior of $p\%$" to mean that we take the expected value of the mosaic given a classifier with an accuracy of $p\%$ and with symmetric true positive and negative rates. 
Note that one can extend these results to consider asymmetric rates without loss of generality. 

The classifier posterior method to evaluate the shadow matching mosaic is an enabling technique during system design and online AI decision making. 
First, the classifier posterior method allows the system designer to assess where in the city shadow matching is most or least informative. 
It also allows the designer to rapidly test different confidence collection objectives (Eq.~\ref{eq:confcollectoptimization}) for different AI-driven LOS classifiers without having to run each classifier in the loop with shadow matching. 
Most importantly, it allows a system designer to properly propagate the uncertainty in AI-driven LOS classifiers into the state uncertainty to enable downstream AI decision-making.
Second, the classifier posterior method allows an AI agent to reason about how the shadow matching uncertainty can evolve over the course of a trajectory. 
We note that our tree-based structure in concert with the set-based mosaic uniquely enables this classifier posterior method. 
In GBSM, there is no embedded structure to connect grid cells based on the available satellite information or shadow geometry. 
Grid cells with similar probability may or may not hold identical LOS/NLOS classification.
In MZSM, our tree provides this necessary structure and allows the agent to forecast the mosaic, the PMF over the mosaic, and the confidence collections.
Throughout these results we take the role of a system design for AI agents where we seek to understand the expected shadow matching mosaic and adjust the design of an AI agent with insights from MZSM.

We assess our proposed algorithm's performance under four values of classifier posteriors (Figure~\ref{fig:sensitivity_mosiac}). The lower bound on classifiers is an uninformative classier with a 50\% classifier posterior. 
An uninformative classifier produces a uniform distribution over the scene and does not locate the user. 
The upper bound on classifiers is a near-perfect classifier with a very high classifier posterior, such as 99.99\%. 
The near-perfect classifier produces a more precise user location and mirrors our past work in~\cite{bhamidipati2021set} where we used an ideal classifier with 100\% classifier posterior.
In between, we have more typical classifiers as detailed in \cite{Xu2018classifier, xu2020machine}. 
In particular, we consider a classifier posterior of 75\% that reflects a simple one-dimensional Bayesian classifier with $\cno$, whose accuracy varies between 56\% to 96\% depending on the scene and exterior building material. 
Lastly, we consider an 85\% classifier posterior to model the state-of-the-art support vector machine (SVM) approaches, which use various satellite signal characteristics to consistently achieve 80\% to 90\% classification accuracy.
A system designer should use their AI-driven classifier's performance at their city of interest in more tailored analyses. 

Our analyses in Figure~\ref{fig:sensitivity_mosiac} demonstrate that the jump from 
75\% to 85\% classifier accuracy allows for a noticeably more precise expected localization by concentrating the probability mass to the fewer leaves.
The two disjoint modes (most clearly seen as the red polytopes in the near-perfect classifier case) represent the same leaf. 
In this scenario, there is not sufficient information from shadow matching to provide a more precise localization than the near-perfect classifier. 
Our past work \cite{bhamidipati2021set} provides further insights into this disjoint set behavior. 
With MZSM, the 75\%, 85\%, and 99.99\% classifier posterior cases locate the correct street and correct side of the street, in expectation.

\subsection{Improved risk-awareness in uncertainty models via MZSM compared to GMMs}

\begin{figure}
    \centering
    \centering
    \begin{subfigure}[b]{0.32\textwidth}
         \centering
         \includegraphics[width= \textwidth, trim={1.6cm 0cm 0cm 0.3cm}, clip]{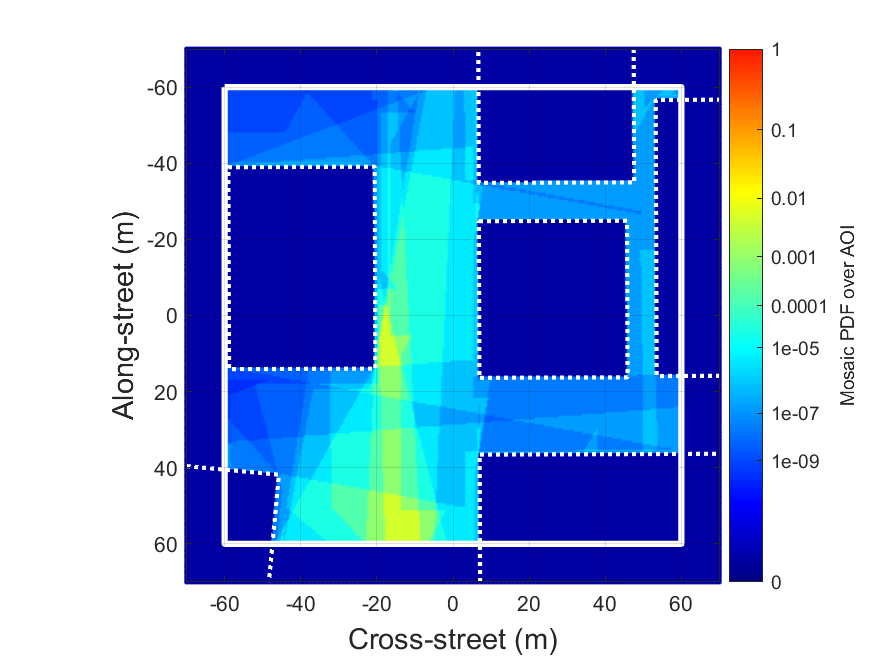}
         \caption{Mosaic PDF \\($\mzsm$)}
     \end{subfigure}
    \hfill
    \begin{subfigure}[b]{0.32\textwidth}
         \centering
         \includegraphics[width= \textwidth, trim={1.6cm 0cm 0cm 0.3cm}, clip]{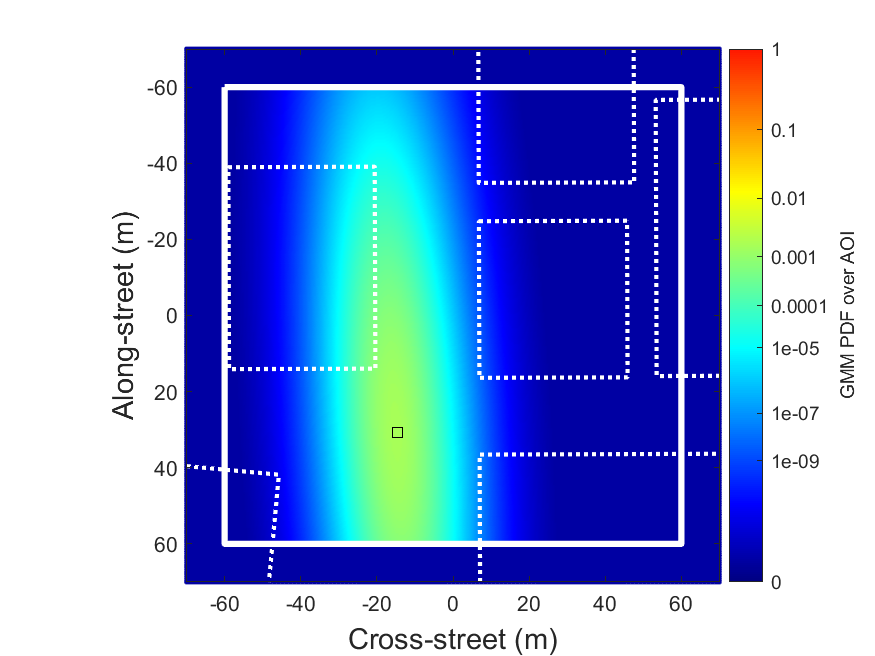}
         \caption{GMM PDF ($k=1$) \\ ($\gmm{1}$)}
     \end{subfigure}
    \hfill
    \begin{subfigure}[b]{0.32\textwidth}
         \centering
         \includegraphics[width= \textwidth, trim={1.6cm 0cm 0cm 0.3cm}, clip]{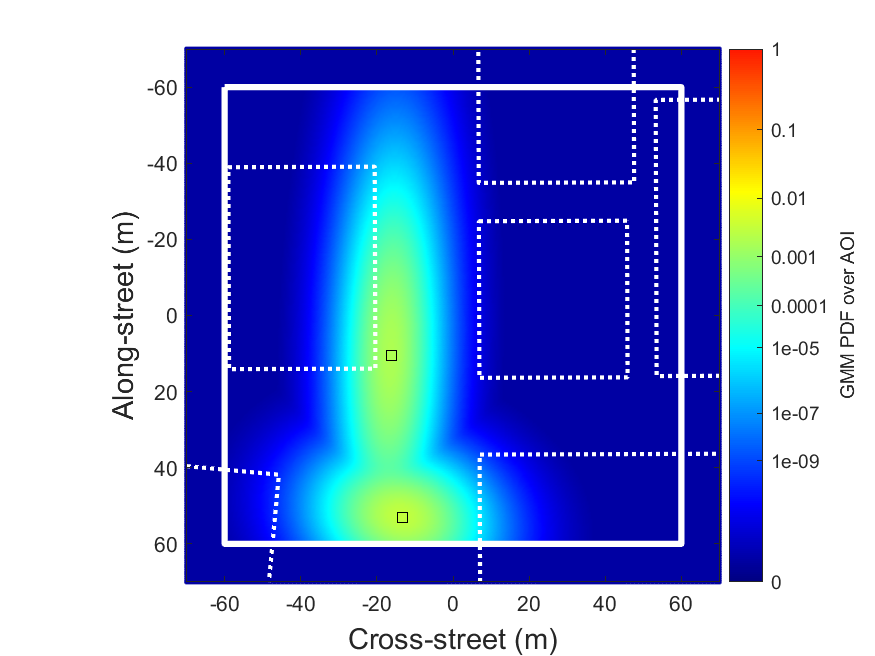}
         \caption{GMM PDF ($k=2$) \\ ($\gmm{2}$)}
     \end{subfigure}
     \par\medskip
    \begin{subfigure}[b]{0.32\textwidth}
         \centering
         \includegraphics[width= \textwidth, trim={1.6cm 0cm 0cm 0.3cm}, clip]{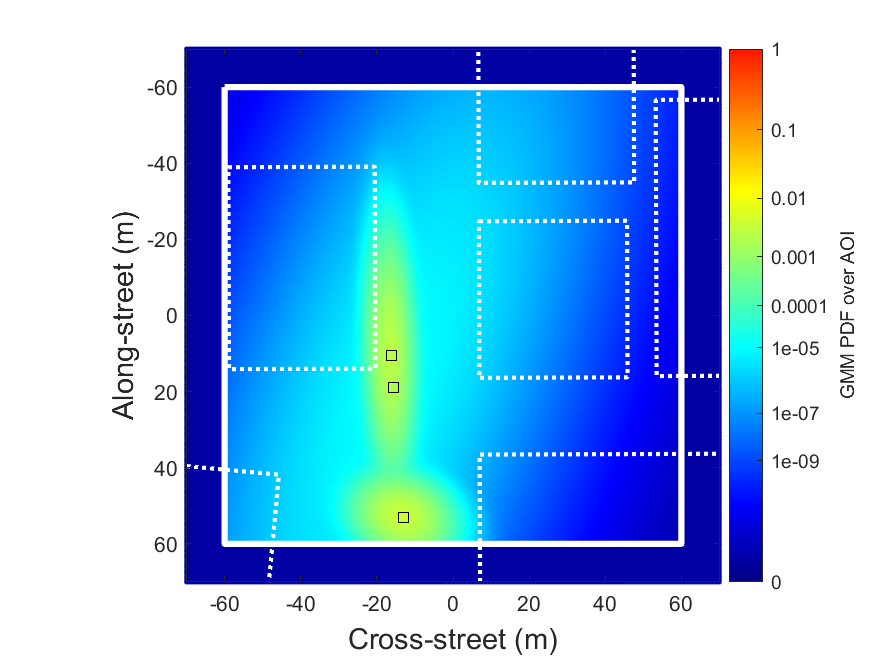}
         \caption{GMM PDF ($k=3$) \\ ($\gmm{3}$)}
     \end{subfigure}
    \hfill
    \begin{subfigure}[b]{0.32\textwidth}
         \centering
         \includegraphics[width= \textwidth, trim={1.6cm 0cm 0cm 0.3cm}, clip]{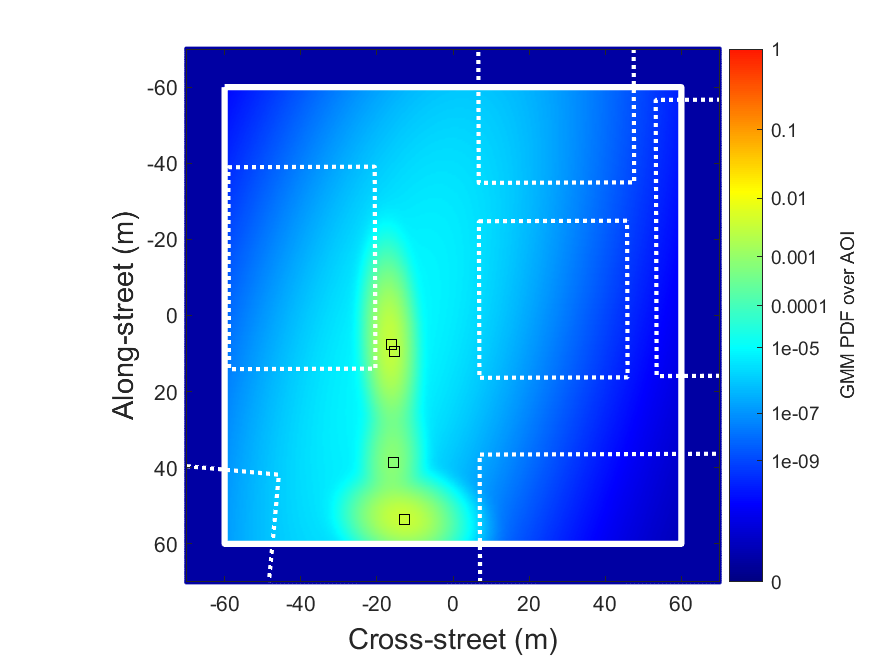}
         \caption{GMM PDF ($k=4$) \\ ($\gmm{4}$)}
     \end{subfigure}
    \hfill
    \begin{subfigure}[b]{0.32\textwidth}
         \centering
         \includegraphics[width= \textwidth, trim={1.6cm 0cm 0cm 0.3cm}, clip]{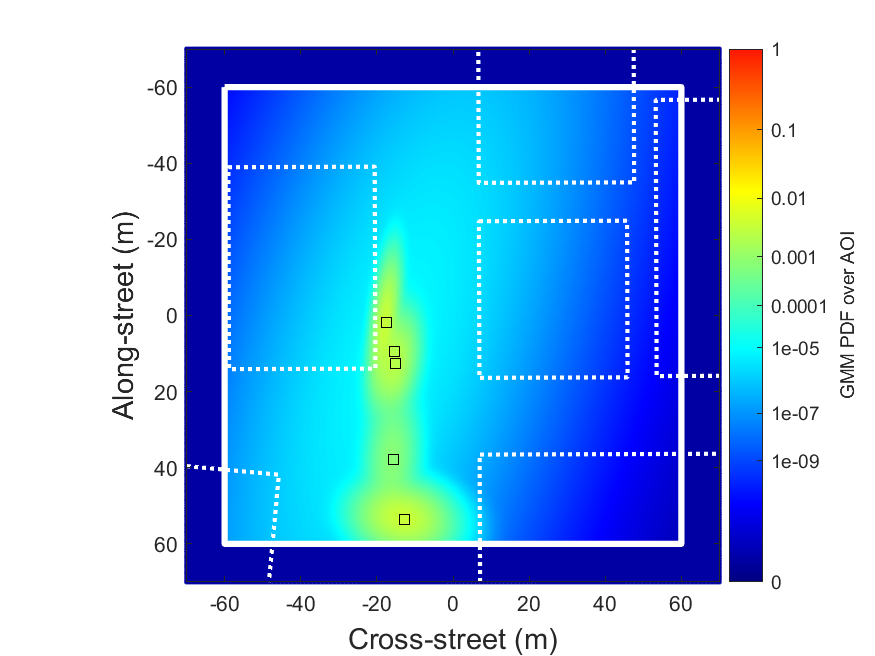}
         \caption{GMM PDF ($k=5$) \\ ($\gmm{5}$)}
    \end{subfigure}
    \hfill
     \begin{subfigure}[b]{0.68 \textwidth}
          \includegraphics[width = \textwidth]{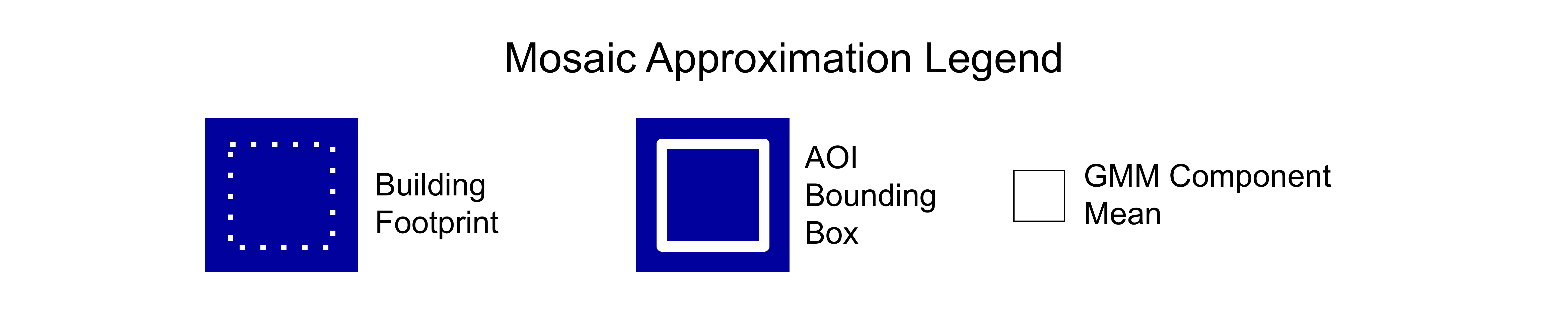}
     \end{subfigure}
    \caption{GMM fits to the MZSM mosaic across a number of mixture components ($k=1,2,3,4,5$) for a 85\% classifier posterior, which matches a typical SVM classifier. 
    Note that, while Figure~\ref{subfig:sensitivity_mosiac_0_85} also discusses a classifier posterior of 85\%, it plots PMF over the mosaic~$ p(\leafnode{l} \mid \AOIvertex )$ from Eq.~\eqref{eq:AOI_prob_normalization}.
    In contrast, we showcase the PDF over the mosaic $\mzsm(x, y)$ conditioned on the AOI, per Eq.~\eqref{eq:mzsm_pdf}, in a).
    The thick white box outline marks the finite support of the mosaic, which is the AOI bounding box. 
    A GMM has infinite support and crosses over the mosaic's finite support, however, we show the truncated GMM to match the AOI information in the mosaic. 
    In (c), the GMMs correctly locate the two modes of the shadow matching mosaic for $k \geq 2$. 
    In comparison to (a), we demonstrate that by increasing the mixture components from $k=1$ in (b) to $k=5$ in (f) we cannot match the preciseness of our proposed MZSM with a GMM. 
    Even with the 5 mixture component GMM ($\gmm{5}$) overestimates the probability in the adjacent streets and alleys while underestimating the mosaic at the polytope edges. 
    }
    \label{fig:gmm_distributions}
\end{figure}



Gaussian error models are highly popular due to modeling simplicity but can be far too restrictive of an assumption. 
Per figure~\ref{fig:sensitivity_mosiac}, the shadow matching mosaic is highly non-Gaussian. 
In Figure~\ref{fig:gmm_distributions}, we show the results of fitting a GMM to the PDF of the shadow matching mosaic ($\mzsm$, defined in Eq.~\eqref{eq:mzsm_pdf}) at a classifier posterior of 85\%. 
As illustrated in Figure~\ref{fig:sensitivity_mosiac}, the shadow matching mosaic ($\mzsmraw$) visually appears to have two modes. 
The two visual modes are more apparent as the red polytopes in the near-perfect classifier case. 
With two mixture components, the optimizer fits one mixture component of roughly equal weight (52\% and 48\%) around each of the two visually identified modes. 
With four mixture components, it adds low-weighted mixture components (weights of 2\% and 8\%) around each visually identified mode. 
The mode with a 2\% weight has a wide covariance to match the spread across the entire street and the mode with an 8\% weight is more centered between the two visually identified modes to match the spread across the intersection. 
With five mixture components, it separates the mode nearer to the center of the AOI into three mixture components with weights of 22\%, 22\%, and 2\%. 
Even then, the Gaussian mixture model with 3 to 5 mixture components appears quite similar to the two mixture component Gaussian mixture model. 
None of the Gaussian mixture models match the PDF of the shadow matching mosaic ($\mzsm$). Since performance plateaus, we only entertain GMMs with up to 5 mixture components. 

We can quantify to what extent the GMM approximates the shadow matching mosaic at a given classifier posterior as a percent error. 
Specifically, we can numerically integrate the error between the $k$-mixture-component GMM probability density ($\gmmk$) versus the probability density ($\mzsm$) of shadow matching mosaic ($\mzsmraw$) (Equation~\ref{eq:integral_abs_error}). 
\begin{align}
    \percenterror &= \left(\iint_{\text{AOI Box}} \mzsm(x, y) \ dx \  dy \right)^{-1} \iint_{\text{AOI Box}} |\gmmk(x, y) - \mzsm(x, y)| \ dx \  dy \\
    \percenterror &= \iint_{\text{AOI Box}} |\gmmk(x, y) - \mzsm(x, y)| \ dx \  dy \label{eq:integral_abs_error}
\end{align}
\noindent where $\percenterror$ is the integrated percent error, $\gmmk(x, y)$ is the GMM PDF for $k$ mixture components, 
AOI Box is the AOI bounding box,
and ($x,y$) are the cross-street and along-street dimensions, respectively. 
We condition $\gmmk(x, y)$ on the AOI bounding box to integrate to unity in finite integration bounds. 
Simply, this truncates the GMM to the AOI bounding box as seen in Figure~\ref{fig:gmm_distributions}. 
By construction, $\percenterror$ is bounded to $0\% \leq \percenterror \leq 200\%$ (equivalently  $0 \leq \percenterror \leq 2$). When $\percenterror > 100\%$, the distributions are significantly anti-aligned such that one is high when the other is low. When $\percenterror < 100\%$, the distributions are roughly aligned such that the GMM generally captures the modes of the MZSM distribution. A $\percenterror > 50\%$ would be a poor approximation, a $10\% < \percenterror < 50\%$ would be a decent approximation, and $\percenterror < 10\%$ would be a good approximation. 

\begin{theorem} \label{theo:propositionD}
The integrated percent error ($\percenterror$) is bounded to $0 \leq \percenterror \leq 2$.
\end{theorem}
\begin{proof}
See \ref{app:AppendixD}.
\end{proof}

\begin{figure}
    \centering
    \includegraphics[width = 0.9 \textwidth]{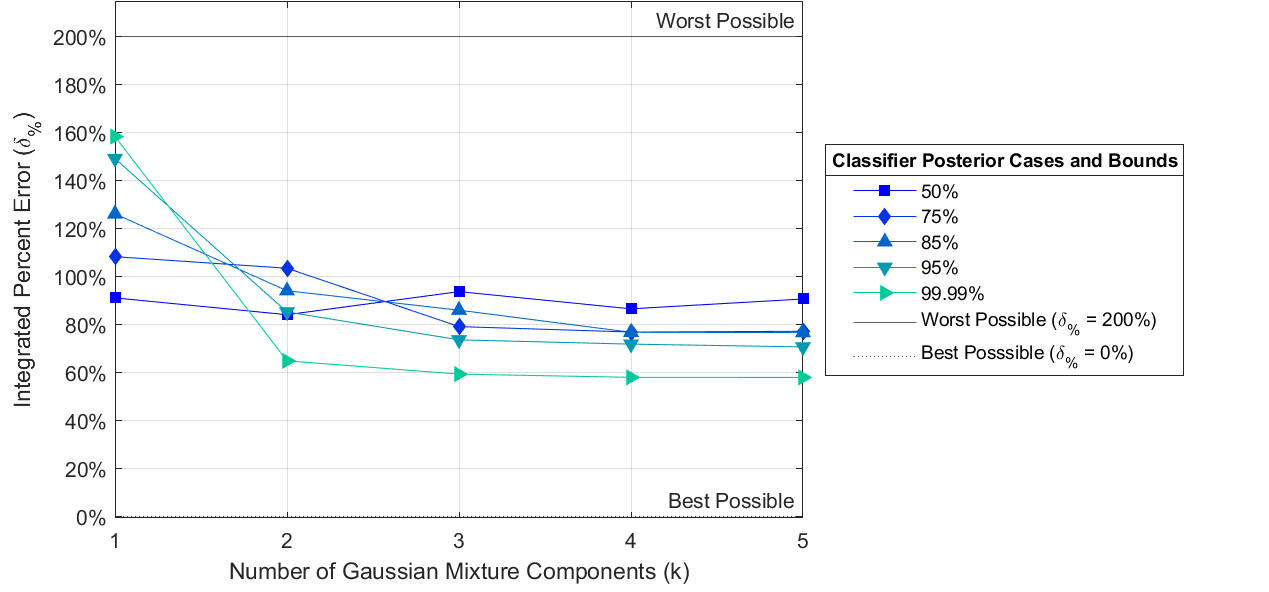}
    \caption{Integrated Percent Error of GMM approximations of the MZSM mosaic to evaluate the GMM approximation accuracy.
    We compare the GMM approximation across mixture components ranging from 1 to 5 and across five classifier posteriors of 50\%, 75\%, 85\%, 95\%, and 99.99\%. 
    Across all classifier posteriors, the GMM approximation greatly improves from a 1 mixture component model to a 2 mixture component model, but the percent error remains consistent thereafter.
    None of the GMM models reach a percent error below 50\% and therefore do not represent a decent approximation of our shadow matching mosaic.
    }
    \label{fig:gmm_integral}
\end{figure}

In Figure~\ref{fig:gmm_integral}, we evaluate the integrated percent error ($\percenterror$) for up to 5 mixture components and classifier posteriors of 50\%, 75\%, 85\%, 95\%, and 99.99\%, where we displayed the GMM and MZSM comparison qualitatively for the 85\% classifier posterior case in Figure~\ref{fig:gmm_distributions}. 
As observed qualitatively in Figure~\ref{fig:gmm_distributions}, the one component Gaussian is largely anti-aligned with the shadow matching mosaic. The GMM is high at its mean is roughly between the two modes of mosaic, but low elsewhere. Correspondingly, $\percenterror > 90\%$ for all the classifiers (Figure~\ref{fig:gmm_integral}). We see an improvement moving from a single Gaussian mixture component to two Gaussian mixture components with $\percenterror$ dropping from between 90\% and 160\% to between 60\% and 110\%. 
However, performance plateaus thereafter. 
A good approximation should have $\percenterror < 10\%$ or better. 
Even with 5 mixture components, the Gaussian mixture model does not even hit $\percenterror = 50\%$. 
Therefore, a Gaussian mixture model can identify all the modes and also provide a rough estimate of the mean of the modes of the shadow matching distribution. 
However, a Gaussian mixture model fails to provide a reasonable approximation of the overall shadow matching mosaic and is thus, not suitable for risk-aware urban localization with shadow matching. 
As an immediate corollary, a Gaussian distribution or a bank of Gaussian distributions in any further filtering framework (i.e., a Kalman filter) is not suitable for use with risk-aware shadow matching. 
An AI decision-making agent using a Gaussian state model with shadow matching is vulnerable to underestimating or overestimating the uncertainty in the scene. 
A set-based uncertainty model is necessary to equip the autonomous agent with the accurate information needed for risk-awareness.

\subsection{Comparison analysis of Set-Based Confidence Bounds with MZSM and SA-GBSM}

\begin{figure}
    \centering
    \begin{subfigure}[b]{0.45\textwidth}
         \centering
         \includegraphics[width= \textwidth]{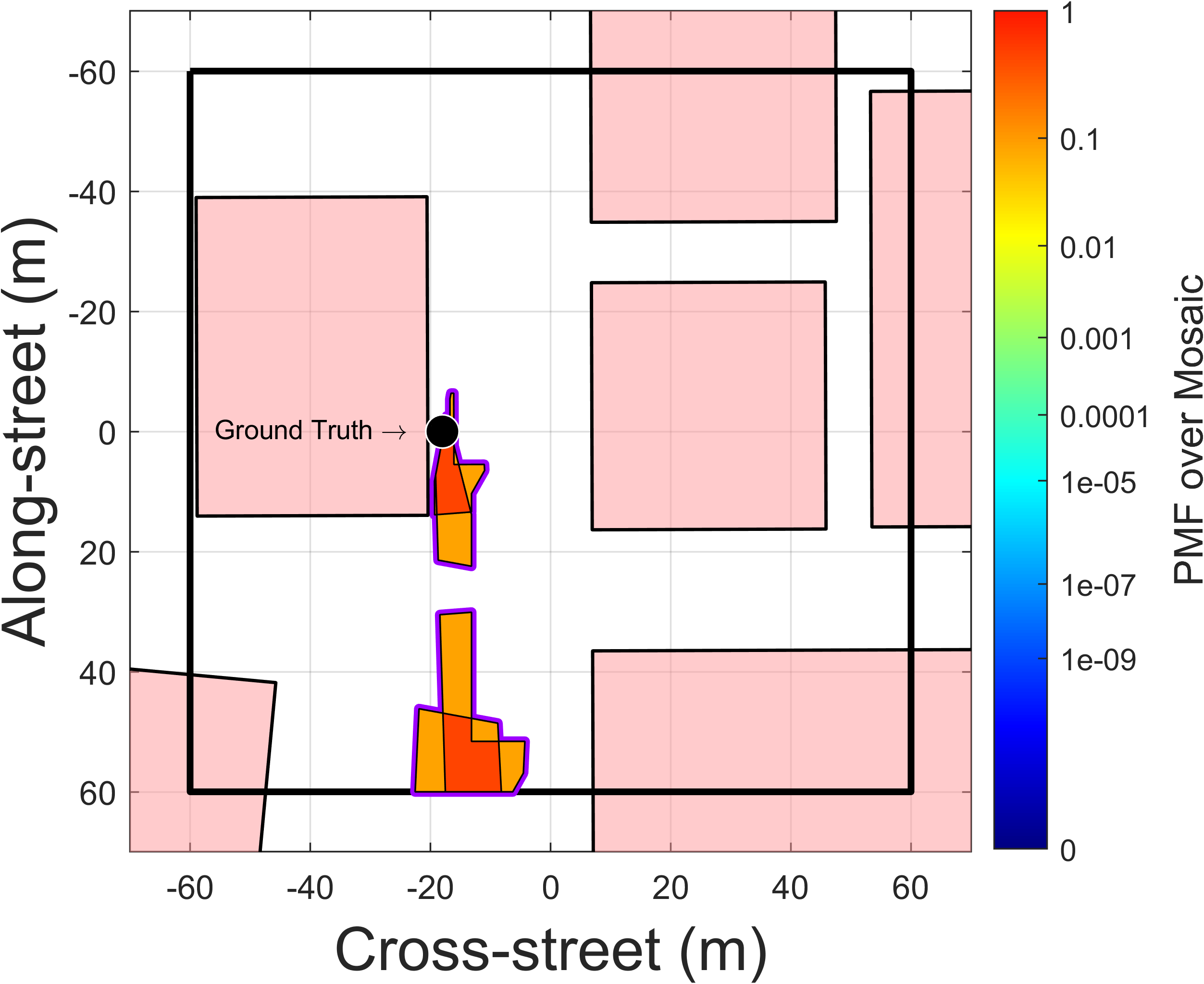}
         \caption{68\% Confidence Collection ($1 \sigma$) \\ $\confcollectval{0.68}$}
     \end{subfigure}
     \hfill
    \begin{subfigure}[b]{0.45\textwidth}
         \centering
         \includegraphics[width= \textwidth]{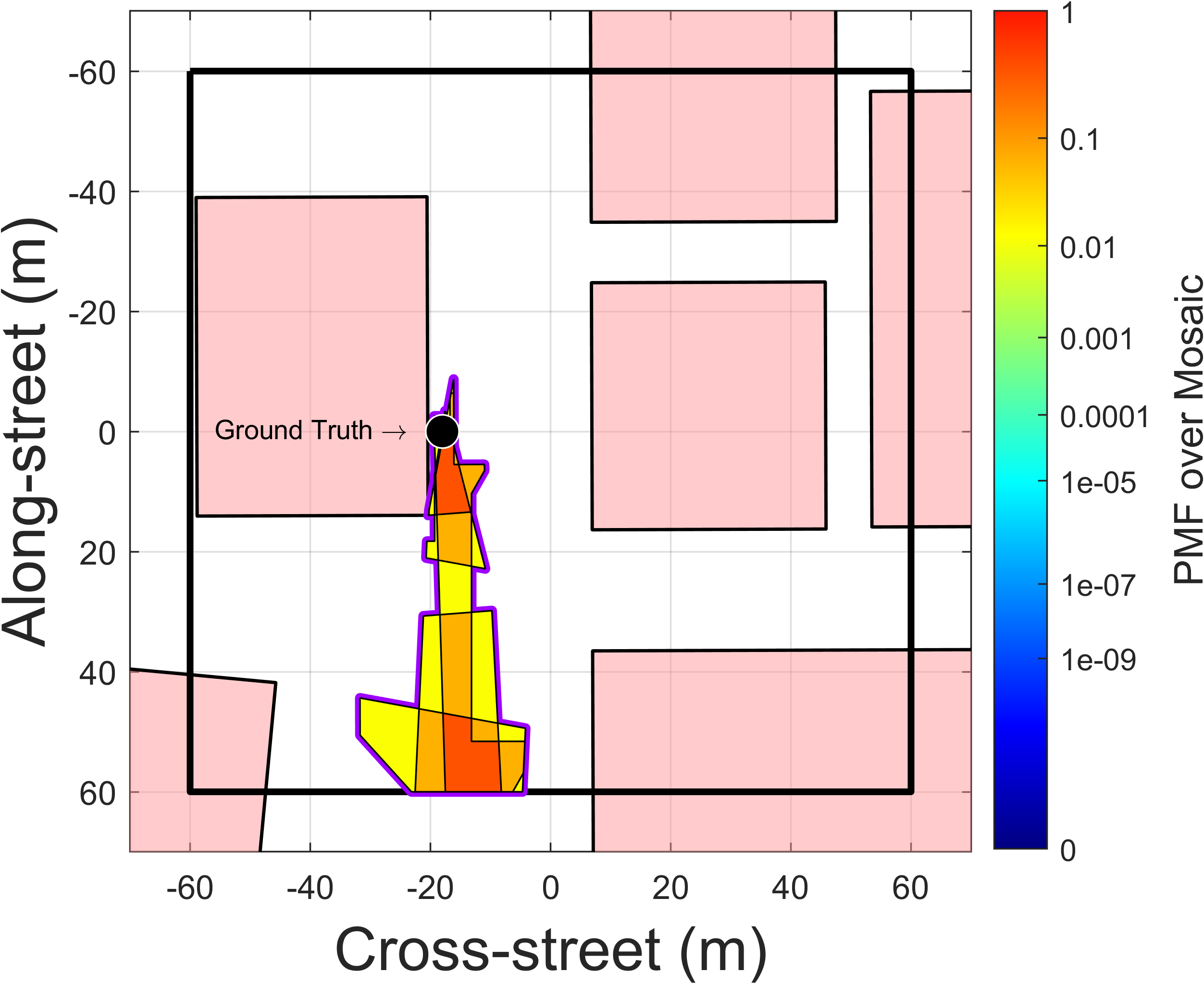}
         \captionsetup{justification=centering}
        \caption{90\% Confidence Collection \\ $\confcollectval{0.90}$}
     \end{subfigure}
     \par\bigskip
    \begin{subfigure}[b]{0.45\textwidth}
         \centering
         \includegraphics[width= \textwidth]{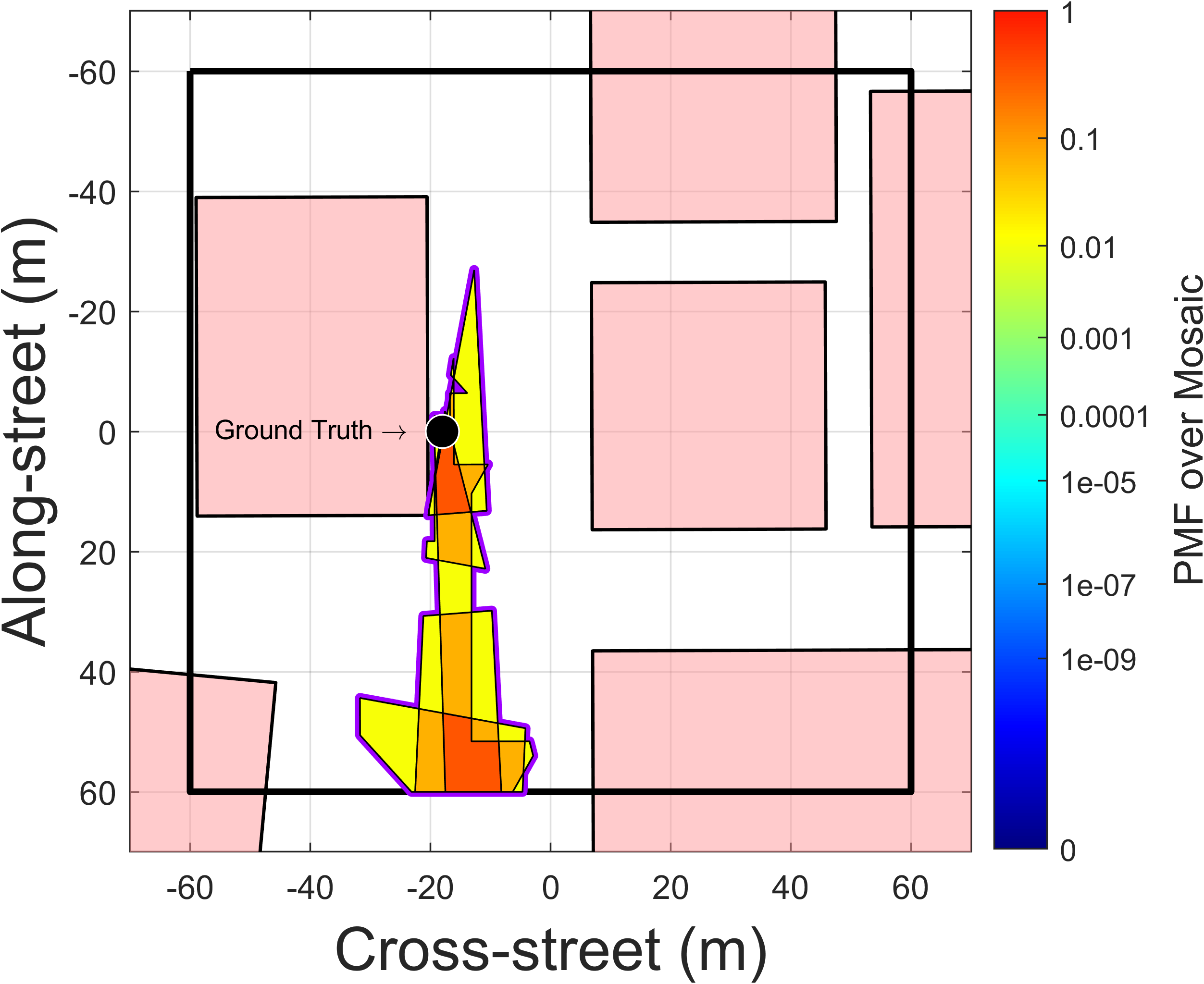}
         \captionsetup{justification=centering}
         \caption{95\% Confidence Collection ($2 \sigma$) \\ $\confcollectval{0.95}$}
         \label{subfig:cc_single_mosaic_0_95}
     \end{subfigure}
     \hfill
    \begin{subfigure}[b]{0.45\textwidth}
         \centering
         \includegraphics[width= \textwidth]{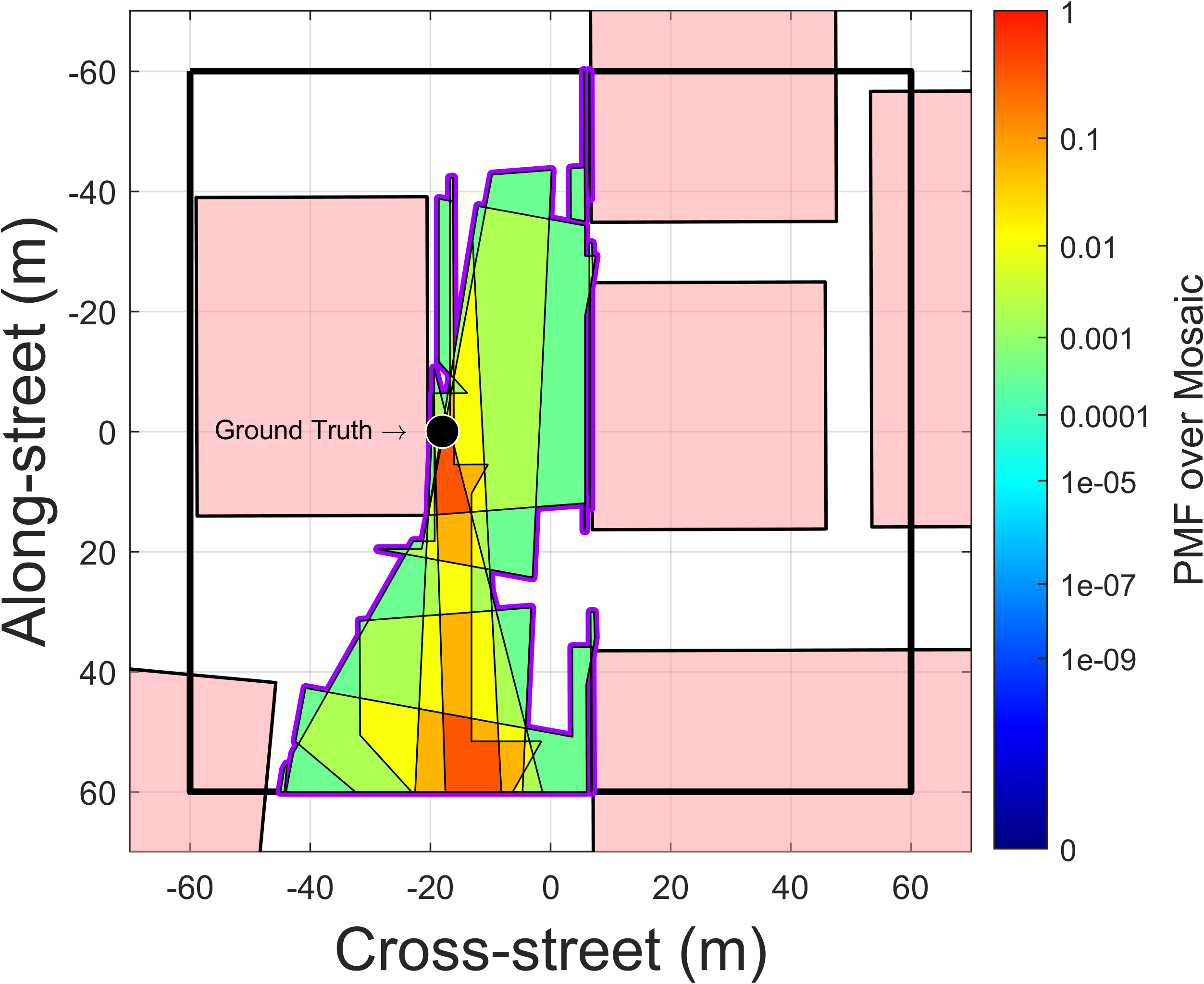}
         \caption{99.7\% Confidence Collection ($3 \sigma$) \\ $\confcollectval{0.997}$}
     \end{subfigure}
     \begin{subfigure}[b]{0.8 \textwidth}
          \includegraphics[width = \textwidth]{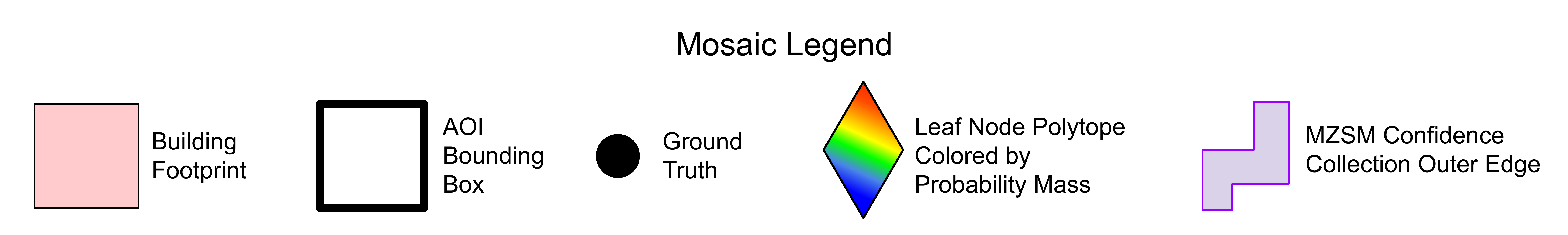}
     \end{subfigure}
    \caption{Confidence collections at different user-defined confidence levels $\gamma=0.68, 0.9, 0.95, \text{ and } 0.997$ for our shadow matching mosaic ($\mzsmraw$) produced using a classifier posterior of 85\%. 
    We showcase the selected set of leaf polytopes in the confidence collection, per Eq.~\eqref{eq:confcollect}, where the colors of different leaf polytopes indicate the AOI-conditioned PMF~(similar to that shown in Figure~\ref{subfig:sensitivity_mosiac_0_85}). 
    In each case, we denote the union of all the sets in the collection, whose outer edge is shown by a thick purple line.
    We demonstrate that our proposed MZSM is confident in the side of street up to 95\% confidence~($2 \sigma$) and confident in the street up to 99.7\% confidence ($3 \sigma$).}
    \label{fig:cc_single_mosiac}
\end{figure}

\begin{figure}
    \centering
    \begin{subfigure}[b]{0.45\textwidth}
         \centering
         \includegraphics[width= \textwidth]{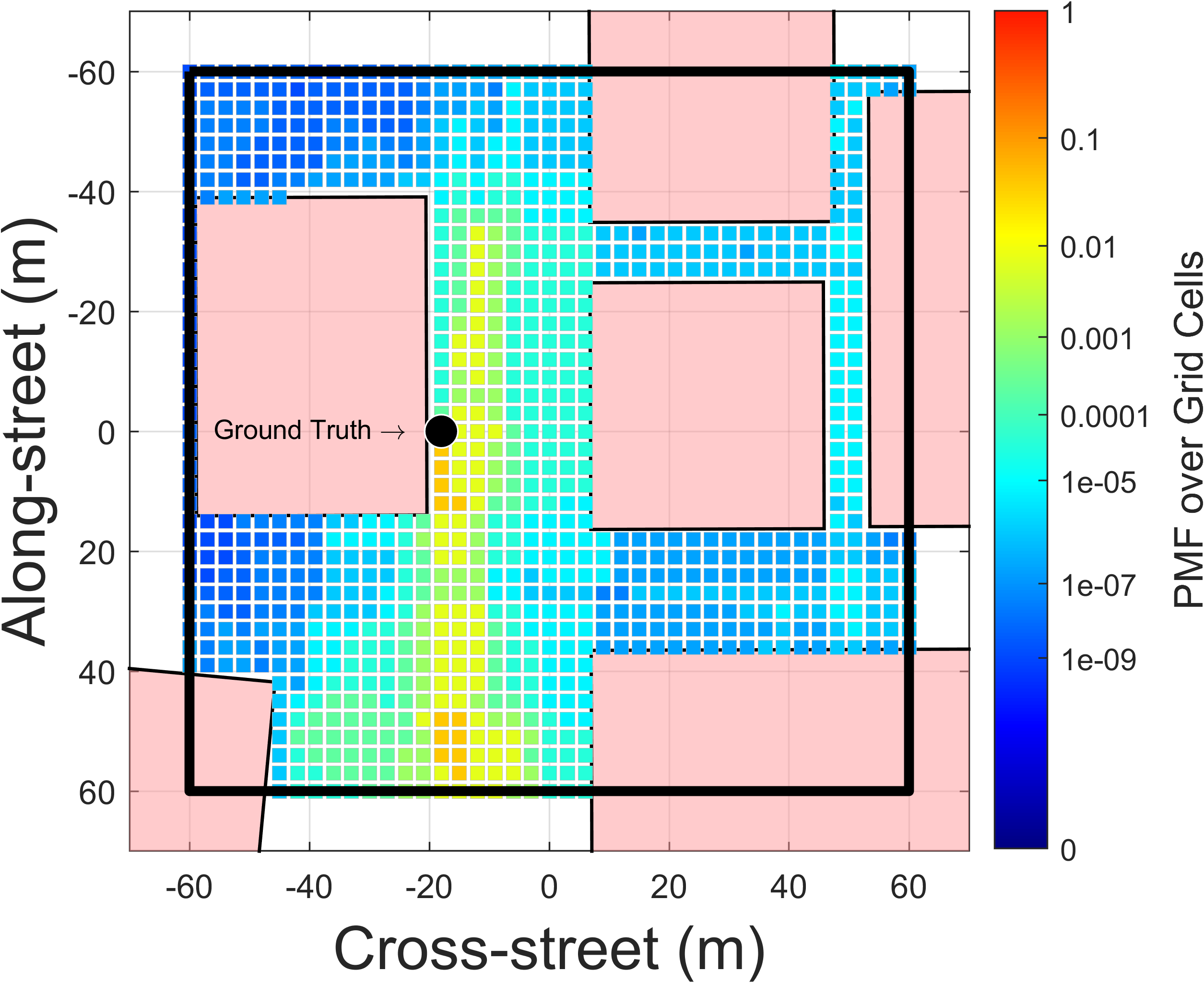}
         \caption{Full GBSM Grid \\ 3~\si{\metre} resolution}
     \end{subfigure}
     \hfill
    \begin{subfigure}[b]{0.45\textwidth}
         \centering
         \includegraphics[width= \textwidth]{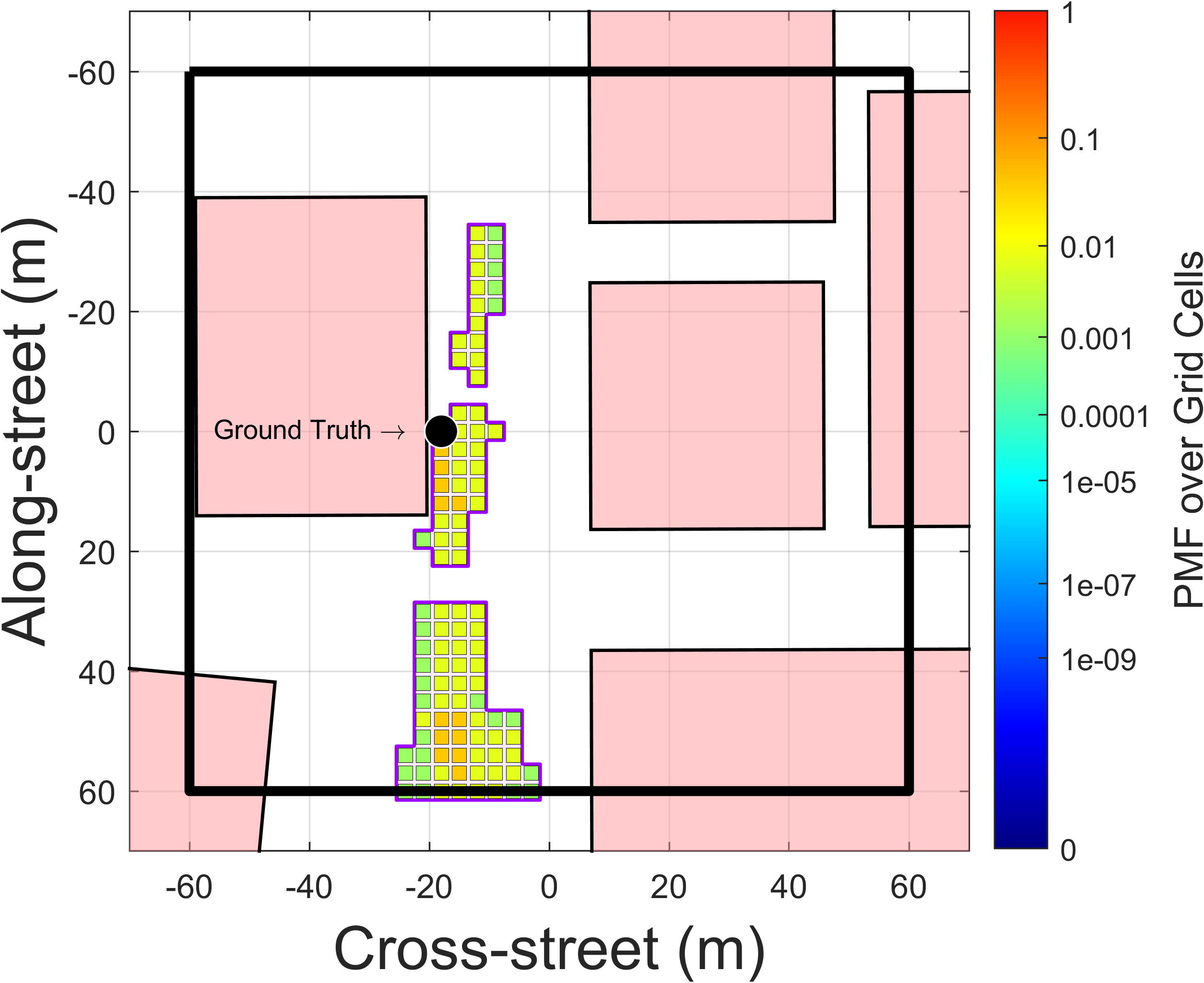}
        \caption{95\% Confidence Collection $\confcollectval{0.95}$ \\ for GBSM at 3~\si{\metre} resolution}
     \end{subfigure}
     \par\bigskip
     \begin{subfigure}[b]{0.45\textwidth}
         \centering
         \includegraphics[width= \textwidth]{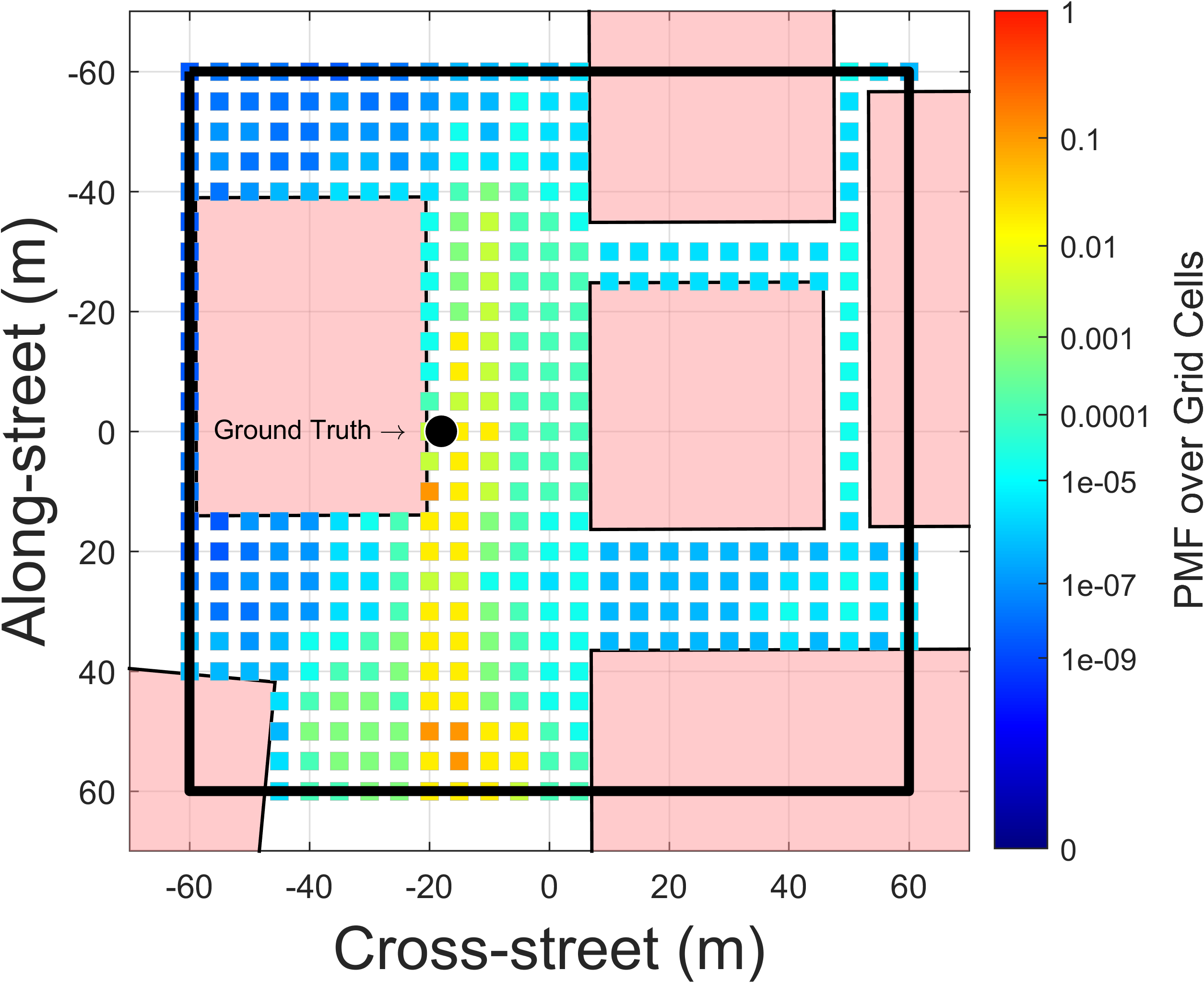}
         \caption{Full GBSM Grid \\ 5~\si{\metre} resolution}
     \end{subfigure}
     \hfill
    \begin{subfigure}[b]{0.45\textwidth}
         \centering
         \includegraphics[width= \textwidth]{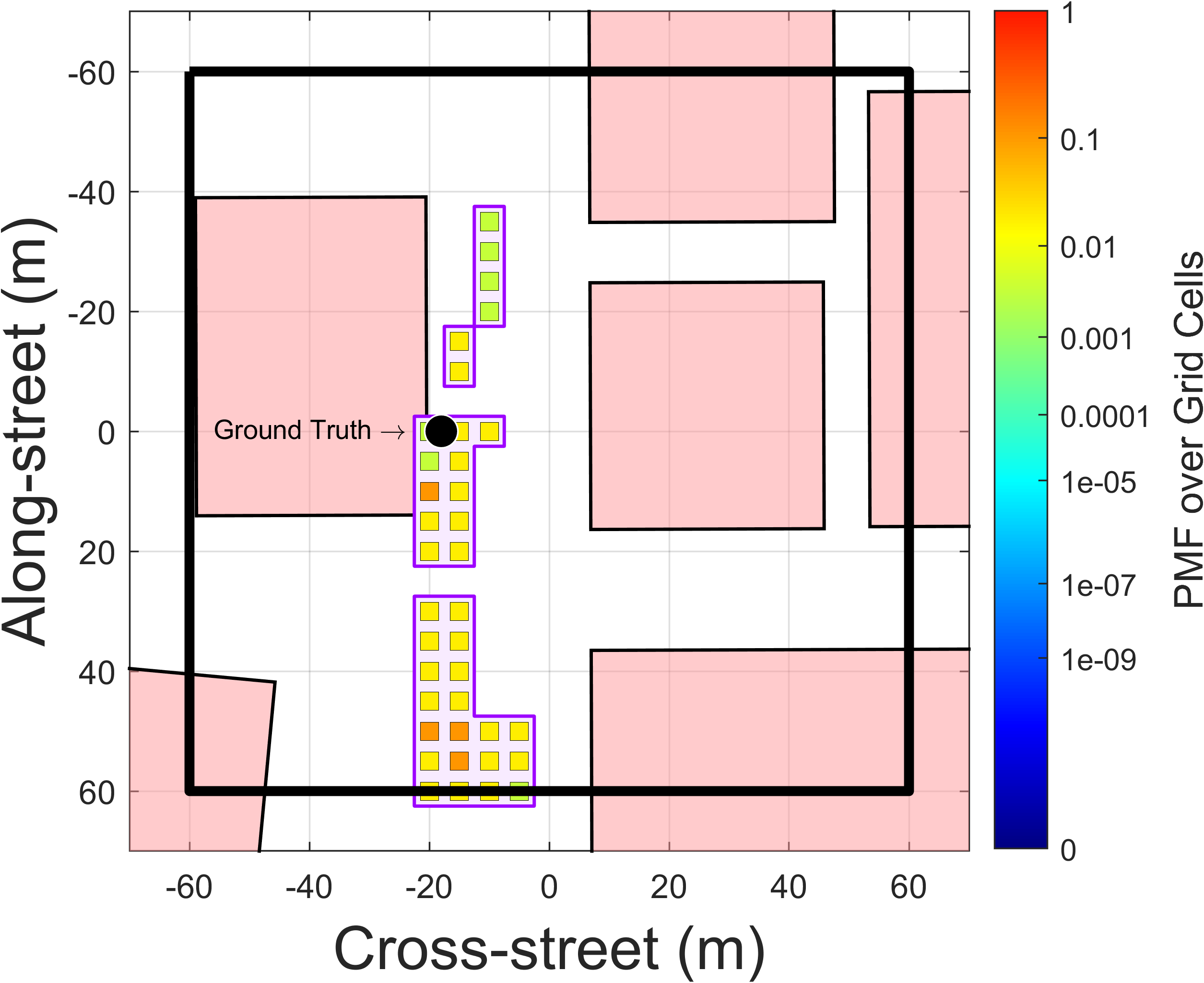}
         \captionsetup{justification=centering}
         \caption{95\% Confidence Collection $\confcollectval{0.95}$ \\ for GBSM at 5~\si{\metre} resolution}
     \end{subfigure}
     \begin{subfigure}[b]{0.8 \textwidth}
          \includegraphics[width = \textwidth]{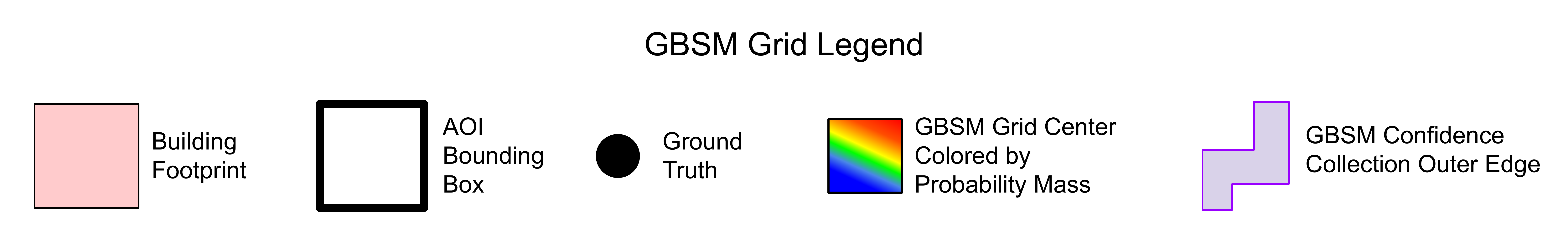}
     \end{subfigure}
    \caption{GBSM (as detailed in \cite{Wang2015probabilistic, Adjrad2016IUP, Adjrad2018IUP}) for a classifier posterior of 85\%, which matches a typical SVM classifier. 
    The subfigures a) and c) denote the GBSM for a grid resolution of 3~m and 5~m, respectively, while the subfigures b) and d) denote the selected grid points that form the confidence collection for GBSM at a user-defined confidence level $\gamma=0.95$, respectively. 
    In b) and d), we denote the outer edge of all the selected grid points by a thick purple line. 
    As the grid resolution refines, the associated confidence collection more closely resembles our proposed MZSM's performance (Figure~\ref{subfig:cc_single_mosaic_0_95}). }
    \label{fig:cc_gbsm_grid_converge}
\end{figure}

In this subsection, we assess the benefits of set-based techniques to compute confidence bounds. 
As detailed in Section~\ref{sec:confidence_collection}, we take a collection of sets such that the summation of probabilities associated with the sets meets the user confidence level. 
We call this collection the confidence collection, whose outer edge is the boundary of the union of all the sets in the collection (purple outline in Figures~\ref{fig:cc_single_mosiac} and \ref{fig:cc_gbsm_grid_converge}). 

In Figure~\ref{fig:cc_single_mosiac}, we illustrate four confidence collections for the shadow matching mosaic. 
At a 68\% ($1 \sigma$) user confidence level, the confidence collection is disjoint in the position domain, clustered around the two visually identified modes. 
By the 90\% user confidence level, the confidence collection is no longer disjoint. 
As the user confidence level increases to 95\% ($2 \sigma$) and 99.7\% ($3 \sigma$) the confidence collection includes more leaf polytopes and expands. 
For a classifier with 85\% accuracy, we find that we can determine the side of the street up to a $2 \sigma$ confidence collection and we can determine the street up to a $3 \sigma$ confidence collection. 

With SA-GBSM, we can also compute confidence collections, as detailed in Section~\ref{sec:gbsm-details}
The precision of the SA-GBSM confidence collection depends on the grid resolution. 
In Figure~\ref{fig:cc_gbsm_grid_converge}, we illustrate the entire grid-based shadow matching output and the 95\% confidence collection for grid resolutions of 3 and 5 meters. 
We use 3 and 5 meter resolutions to match prior GBSM work (e.g., \cite{groves2015gnss, Wang2013samsungbuildingboundaries, Adjrad2016IUP}). 
Qualitatively, the grid-based 95\% confidence collections have a similar shape to the MZSM 95\% confidence collection in Figure~\ref{fig:cc_single_mosiac}. 
However, the grid-based 95\% confidence collections are three disjoint sets and do not fully match any sharp edges in the exact MZSM 95\% confidence collection. 
Quantitatively, we can assess the grid-based approximation with the ``intersection over union" (IOU) metric (also known as the Jaccard index) \cite{Rezatofighi2019IOU}. 
For sets $\mathcal{D}_1$ and $\mathcal{D}_2$, the IOU metric is
\begin{equation}
    \text{IOU}(\mathcal{D}_1, \mathcal{D}_2) = \frac{\text{Area}(\mathcal{D}_1 \cap \mathcal{D}_2)}{\text{Area}(\mathcal{D}_1 \cup \mathcal{D}_2)} \label{eq:iou}
\end{equation}
\noindent where the IOU metric is unity 
if the sets $\mathcal{D}_1$ and $\mathcal{D}_2$ are equal
and zero if the sets are completely disjoint.
We evaluate the IOU metric of the confidence collections produced with MZSM (i.e., the exact shadow matching mosaic $\mzsmraw$) with respect to the confidence collections produced with SA-GBSM (i.e., the grid-based approximation of the shadow matching mosaic) at a variety of specified user confidence levels and classifier posteriors.
Mathematically, $\mathcal{D}_{1}$ and $\mathcal{D}_2$ in Eq.~\eqref{eq:iou} match the union of all the sets in the confidence collection (visually, the purple outer edges in Figures~\ref{fig:cc_single_mosiac} and \ref{fig:cc_gbsm_grid_converge}).
We then determine to what extent the grid-based approximation converges to the shadow matching mosaic as we refine the grid resolution. 

\begin{figure}
    \centering
    \begin{subfigure}[b]{0.45\textwidth}
         \centering
         \includegraphics[width= \textwidth]{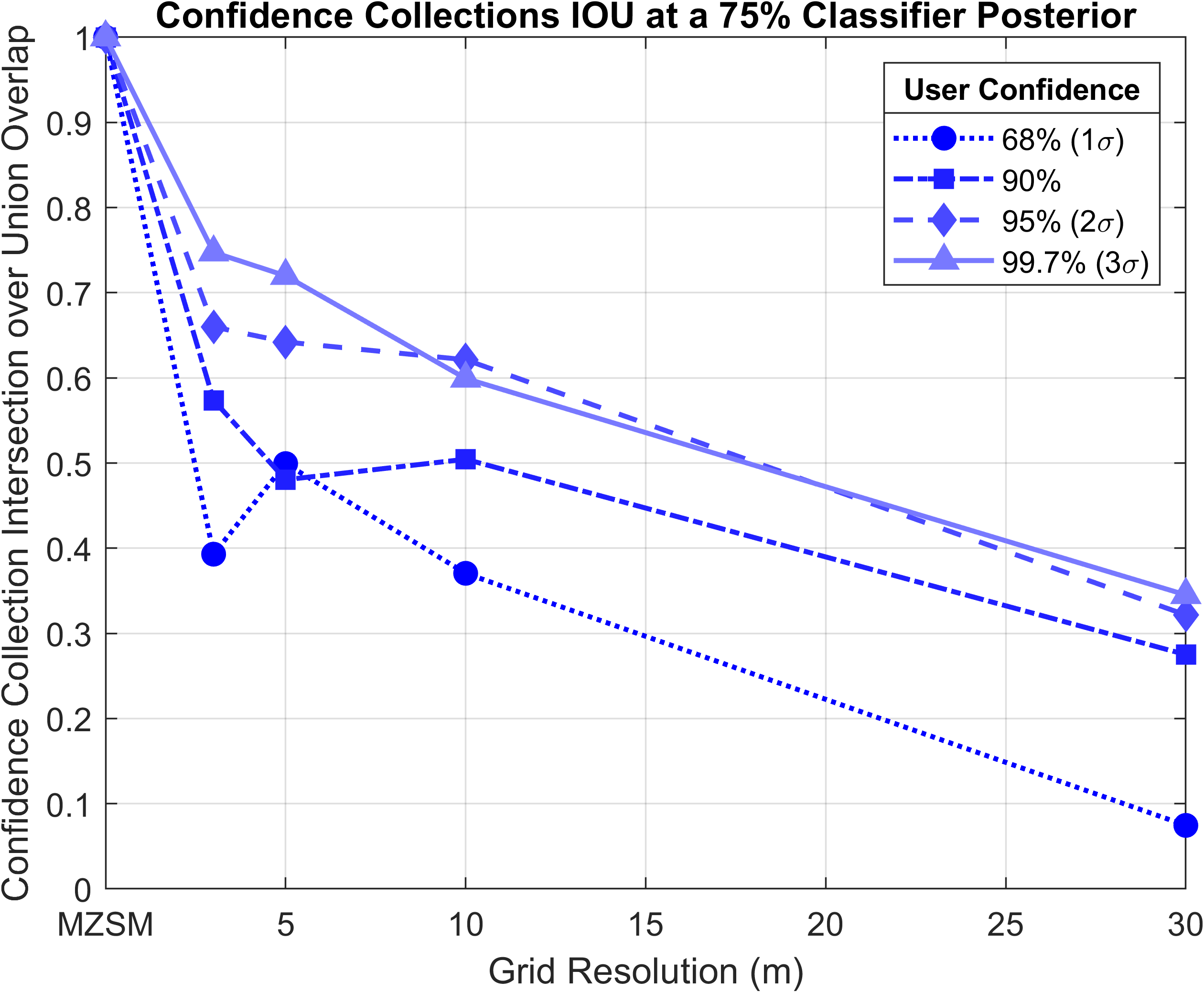}
         \caption{Typical $\cno$ Bayesian Classifier}
     \end{subfigure}
     \hfill
    \begin{subfigure}[b]{0.45\textwidth}
         \centering
         \includegraphics[width= \textwidth]{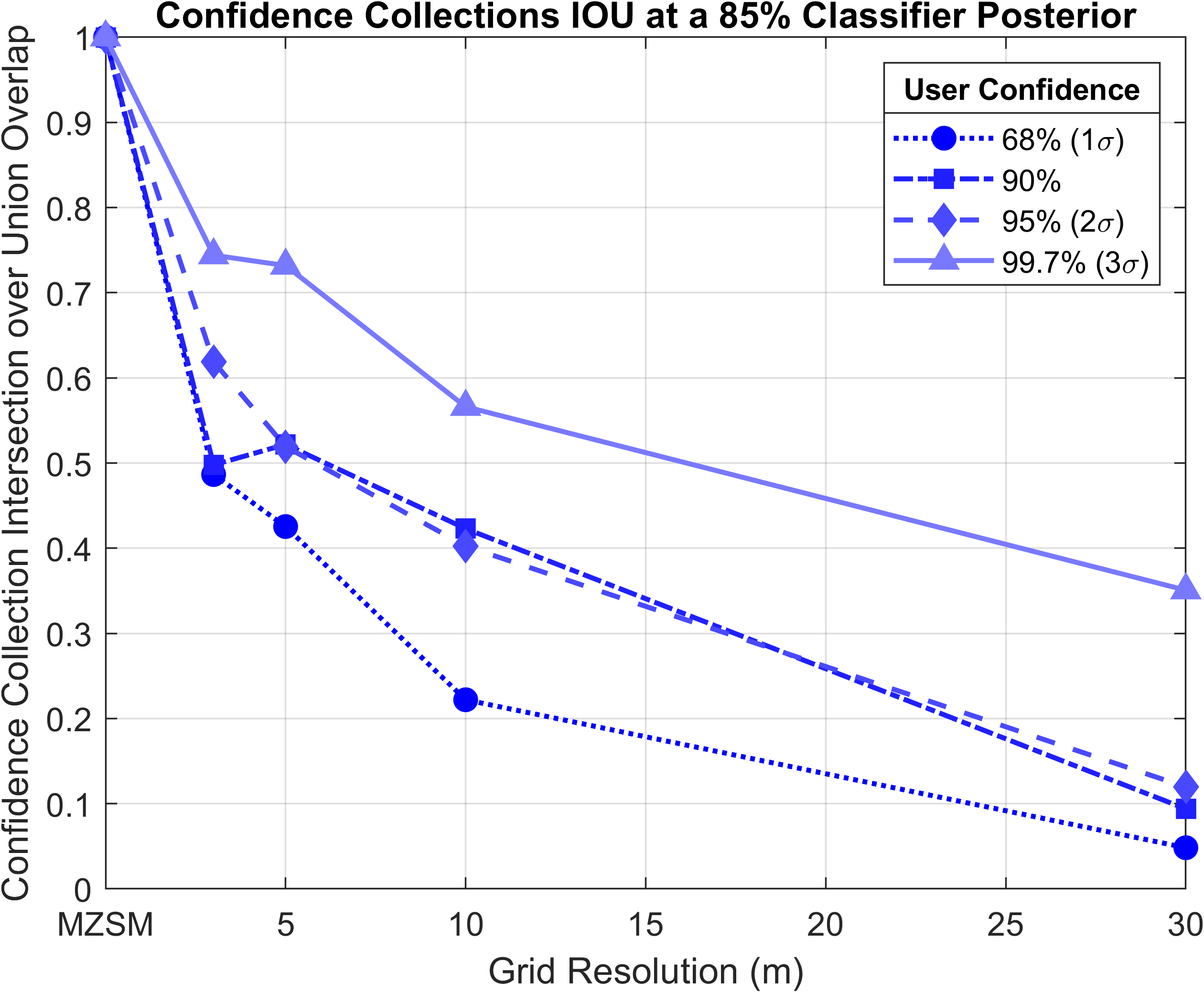}
        \caption{Typical Support Vector Machine Classifier}
     \end{subfigure}
     \par\bigskip
    \begin{subfigure}[b]{0.45\textwidth}
         \centering
         \includegraphics[width= \textwidth]{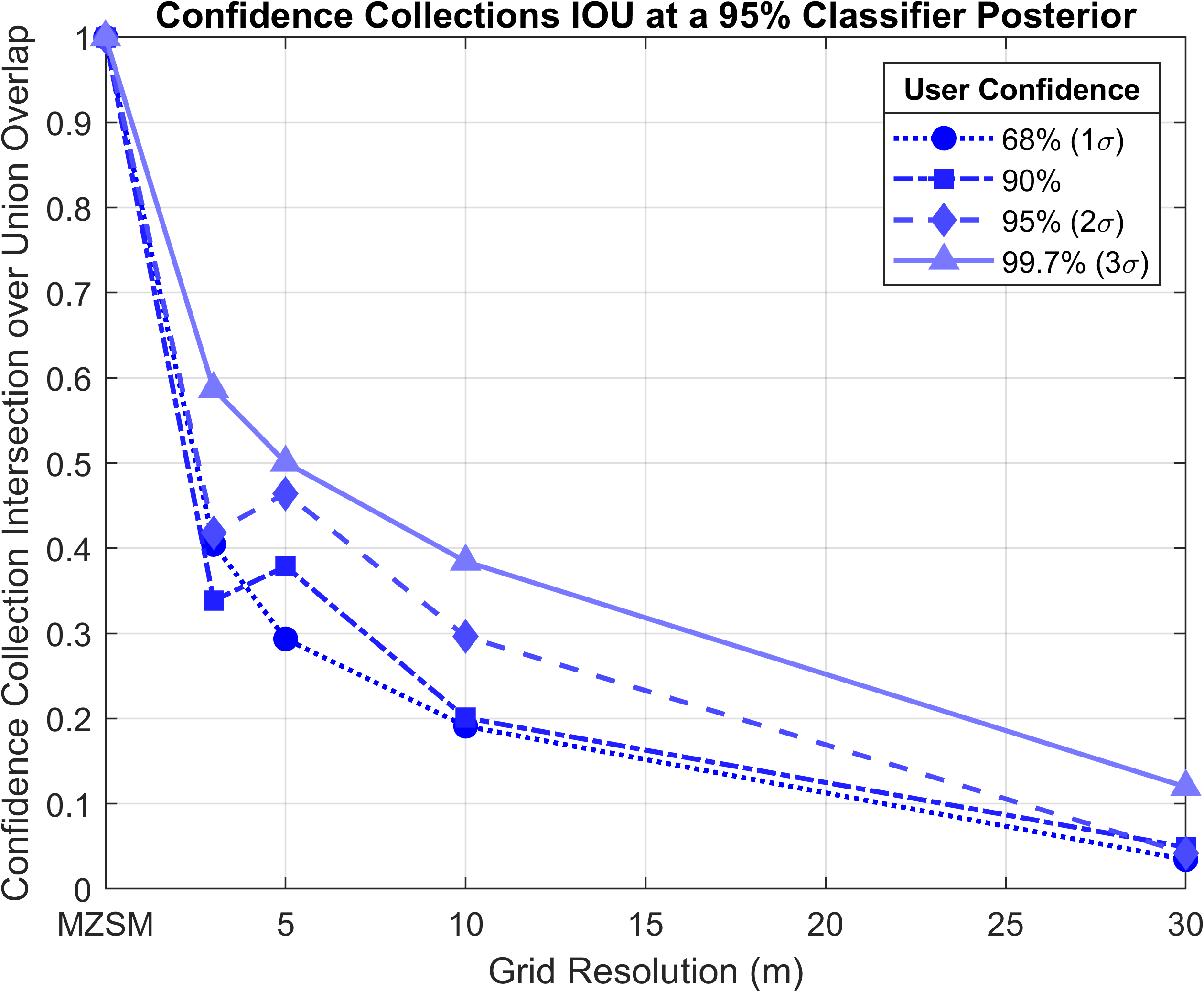}
         \captionsetup{justification=centering}
         \caption{Top Performing Classifiers}
     \end{subfigure}
     \hfill
    \begin{subfigure}[b]{0.45\textwidth}
         \centering
         \includegraphics[width= \textwidth]{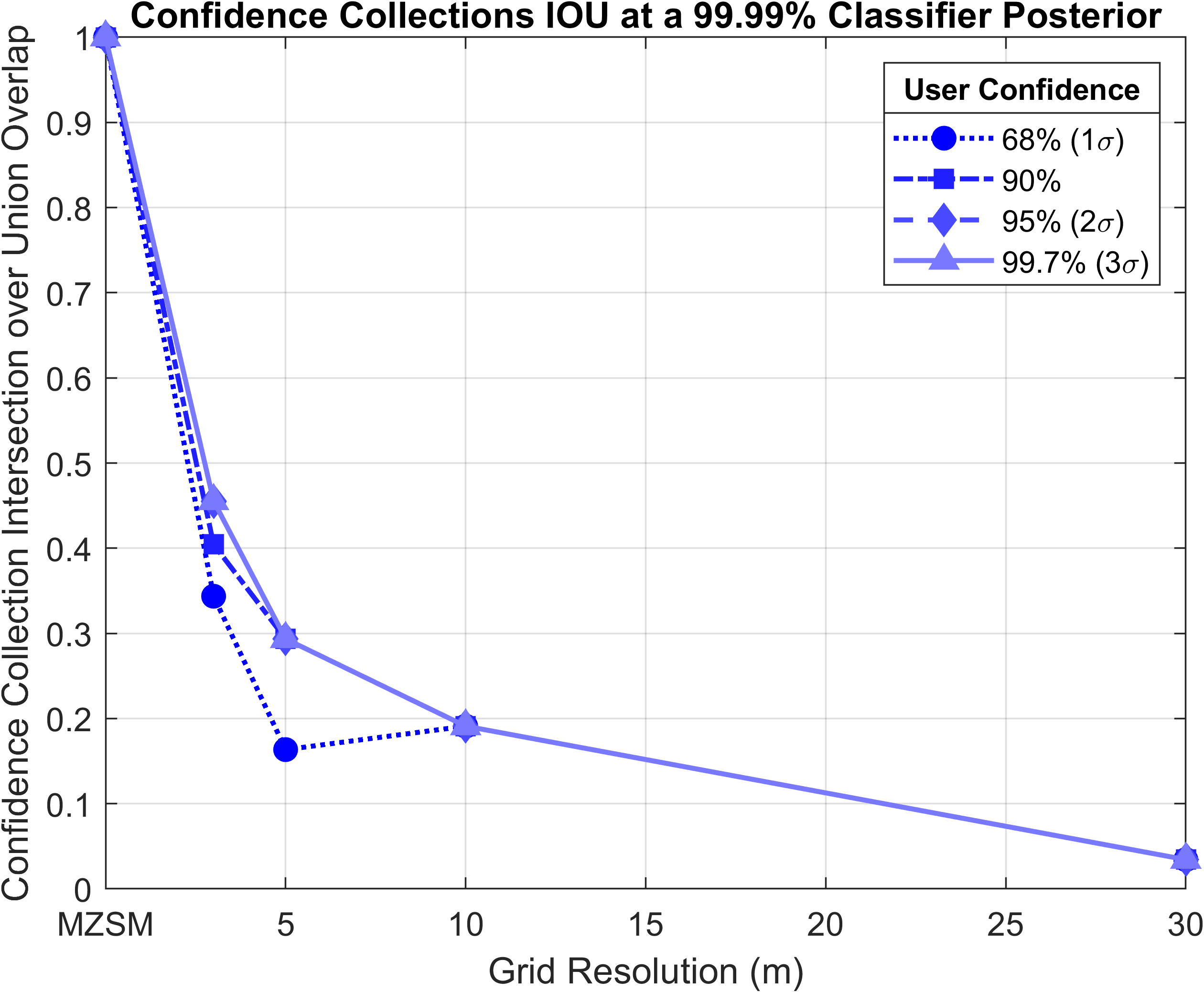}
         \caption{Near-Perfect Classifier}
     \end{subfigure}
    \caption{Convergence of confidence collection (based on IOU metric in Eq.\eqref{eq:iou}) as the grid resolution of SA-GBSM is refined from $30~$m to $3~$m. 
    We analyze the variation of the IOU metric as the classifier posterior is varied from 75\% to 99.99\% and user-defined confidence ($\gamma$) from 68\% to 99.7\%.
    We observe that the SA-GBSM confidence collections converge to the MZSM confidence collections as we refine the grid resolution from 30 m to 3 m.
    However, SA-GBSM more readily converges to MZSM (i.e., the slope is shallower) at lower classifier posteriors (e.g., 75\% classifier posterior in (a)) and higher user confidence levels (e.g., $\gamma = 99.7\%$).
    }
    \label{fig:cc_iou_converge}
\end{figure}

In Figure~\ref{fig:cc_iou_converge}, we compare convergence in grid resolution across 4 classifier posteriors largely matching Figure~\ref{fig:sensitivity_mosiac}. 
Note that a 50\% classifier posterior is uninformative
so, we replace it with a 95\% classifier posterior, which corresponds with a top-performing SVM classifier in a favorable environment \cite{Xu2018classifier, xu2020machine}. 
While past GBSM works use a grid resolution of roughly 3-5 meters, the building boundaries of a 3-5 meter resolution grid cannot be updated sufficiently fast for real-time applications. 
We, therefore, illustrate sparser grid resolutions of 10~\si{\metre} and 30~\si{\metre} that can be updated in less time, though at degraded precision. 
Overall, SA-GBSM confidence collections converge to the MZSM confidence collections as we refine the grid resolution from 30~\si{\metre} to 3~\si{\metre} (Figure~\ref{fig:cc_iou_converge}) going from IOU metrics of 0.4 or less at 30~\si{\metre} to between 0.4 and 0.8 at 3~\si{\metre}. 
The confidence collections cover a larger area at lower classifier posteriors and higher user confidence levels, so a less refined grid resolution better covers the larger area.
Correspondingly, the IOU metric is higher and the convergence slope is shallower at lower classifier posteriors and higher user confidence levels. 
Using our novel MZSM approach, we achieve the performance of highly refined grid resolutions in SA-GBSM while avoiding discretization trade-offs and ensuring compute efficiency. 

\subsection{Self-Monitoring and Efficiency with Tree-based Construction} \label{sec:selfmonitoring}

We demonstrate the benefits of the tree-based structure: how we self-monitor the root of the tree based on the satellite consistency, how we track the tree expansion, and how we quantify the computational load of MZSM.  

\begin{figure}[H]
    \centering
    \includegraphics[width=0.55\textwidth]{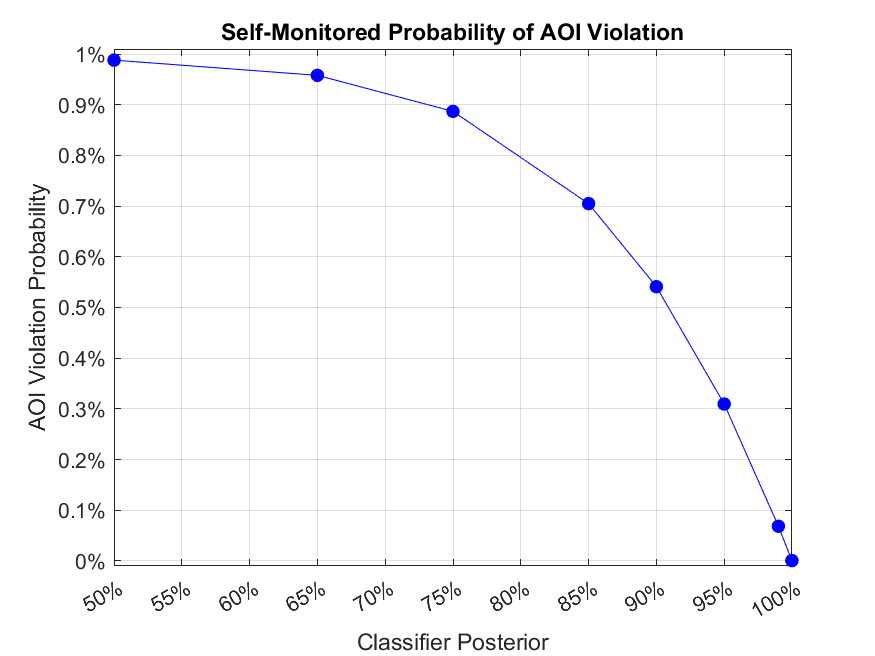}
    \caption{Our MZSM's estimate of AOI violation probability~$p_{\emptyset}$, per Eq.~\eqref{eq:pempty}, for different classifier posteriors, when the true receiver position is in the root AOI. 
    Less accurate classifiers (i.e., classifier posterior of 50\%) provide a less confident assertion that the receiver is in AOI, thus the AOI violation probability is a high value~$\approx 1$. 
    In contrast, more accurate classifiers (i.e., classifier posterior of 99.99\%) provide a highly confident assertion, thus AOI violation probability is very low~$\approx 0$.}
    \label{fig:violation_probability}
\end{figure}

The AOI sets the scale for the MZSM localization task. 
Ideally, we want a smaller AOI without compromising risk-aware localization. 
By design, the tree estimates the probability that the GNSS shadow information violates the assumption that the receiver is in the AOI, which is the root of the MZSM tree (Section~\ref{sec:proposed_method}). 
In Figure~\ref{fig:violation_probability}, we demonstrate how this probability changes with the classifier posterior. 
Specifically, we focus on the current scenario where the true receiver location is in the AOI. 
A highly accurate receiver (i.e., classifier posterior of 99.99\%) concentrates the probability on the leaves that match the true signal designation as LOS or NLOS (such as seen in Figure~\ref{subfig:sensitivity_mosiac_0_9999}). 
In such a case, we accept that the receiver location is in the AOI with the self-monitor logging the AOI violation probability at 0.001. 
A less accurate receiver (i.e.,  classifier posterior of 50\%) spreads the probability mass over many leaves (such as seen in Figure~\ref{subfig:sensitivity_mosiac_0_50}), and thus, the self-monitor alerts that the AOI violation probability is 0.988. 
Practically, the user should consider larger AOIs in expectation when employing a less accurate classifier. 
As illustrated in Figure~\ref{fig:violation_probability}, the relation is nonlinear. 
At moderate classifier accuracies, a small improvement in classifier accuracy yields a noticeably lower AOI violation probability such that the tree provides a noticeably more confident assertion that the receiver is in the root AOI. 
Importantly, our AOI self-monitoring adapts to the classifier and environment, thereby allowing risk awareness without being dependent on a specific classifier. 

\begin{figure}
    \centering
     \begin{subfigure}[b]{0.49\textwidth}
         \centering
         \includegraphics[width= \textwidth]{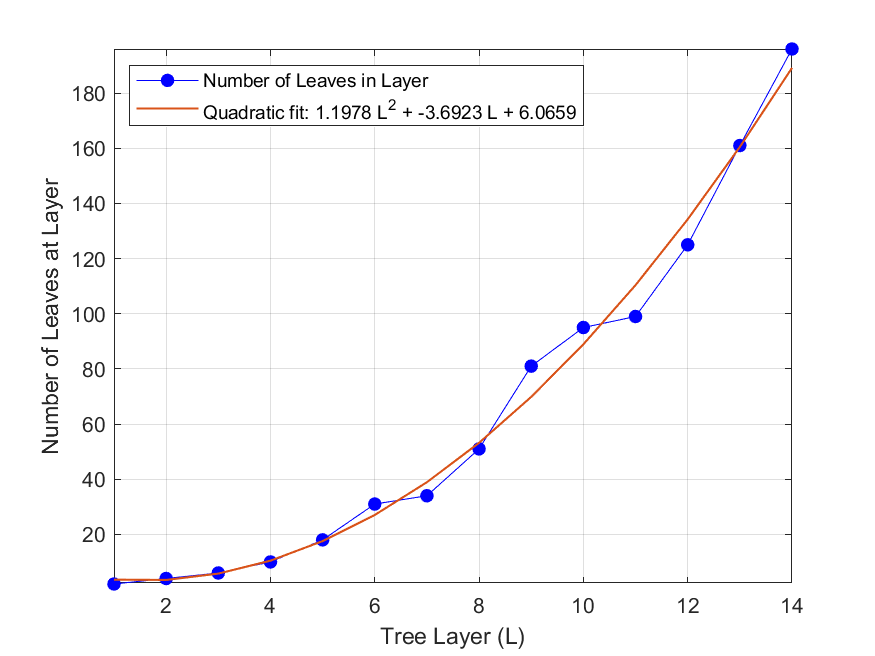}
         \caption{Increase in the number of leaves as the tree expands. }
         \label{subfig:leaf_analysis_num_leaves}
     \end{subfigure}
     \hfill
     \begin{subfigure}[b]{0.49\textwidth}
         \centering
        \includegraphics[width= \textwidth]{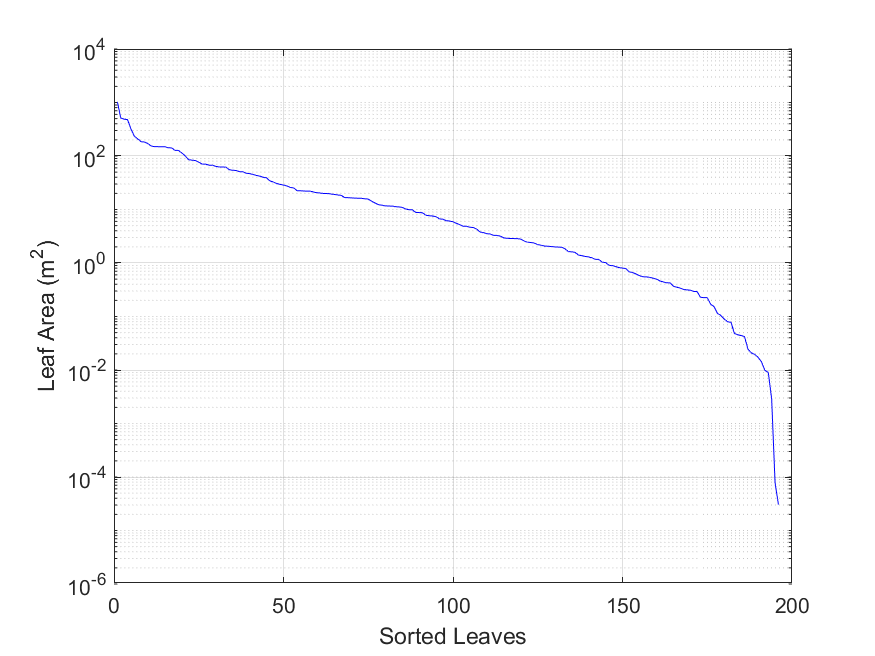}
         \caption{Sorted leaf areas in the final layer of the tree. }
         \label{subfig:leaf_analysis_leaf_areas}
     \end{subfigure}
    \caption{Leaf analysis as the tree expands to a depth of 14 satellites. 
    In (a), we track the number of leaves~(blue line with circular markers) at each iteration of the tree expansion. 
    We fit a quadratic~(orange line) to illustrate the algorithm's quadratic complexity. 
    In (b), we sort the leaves by area and plot them sequentially. 
    Only 102 leaves of 196 total have an area larger than 5~\si{\square\metre} and only 86 leaves have an area larger than 10~\si{\square\metre} (where 10~\si{\square\metre} is just above the area of a sedan's footprint).}
    \label{fig:leaf_analysis}
\end{figure}

Our tree tracks its expansion with new satellite information. 
In Figure~\ref{fig:leaf_analysis}, we compute the number of leaves after each expansion of the tree. 
As noted in Section~\ref{sec:proposed_method}, the tree leaves expand quadratically with the tree depth. 
We experimentally demonstrate this complexity (Figure~\ref{subfig:leaf_analysis_num_leaves}). 
The number of leaves that the tree expansion step generates depends on how much complete overlap there is between the new shadow and any of the past shadows. 
If the new shadow is similar to any of the past shadows, many leaf nodes will be completely in the new shadow or completely out of the new shadow. 
Therefore, the tree will generate fewer new leaves (matching the \nooverlaplower \ and \fulloverlaplower \ cases discussed in Section~\ref{sec:proposed_method}). 
In contrast, a satellite that provides unique geometry will have a shadow that intersects many nodes and produce more new leaves (matching the \partoverlaplower \ case discussed in Section~\ref{sec:proposed_method}). 
In our simulation, the 7th and 11th satellites are blocked throughout most of the AOI. The shadows from these two satellites are both similar to the 3rd satellite. Correspondingly, the 7th and 11th satellites did not generate many new leaves at tree layers 7 and 11, respectively. In contrast, the 9th satellite has a unique shadow and generated many more leaves at tree layer 9. 
Nevertheless, the tree expansion is still quadratic. As the number of shadows increases, some leaf nodes will have a very low area (Figure~\ref{subfig:leaf_analysis_leaf_areas}). 
By the 14th layer, only a few leaves have large areas and many leaves have small areas. 
For example, only 102 leaves of the 196 total leaves have an area larger than 5~\si{\square \metre}, which is roughly half the area of a sedan footprint. 

\begin{figure}
    \centering
    \includegraphics[width = 0.8 \textwidth, trim={3cm 0cm 14cm 0.5cm}, clip]{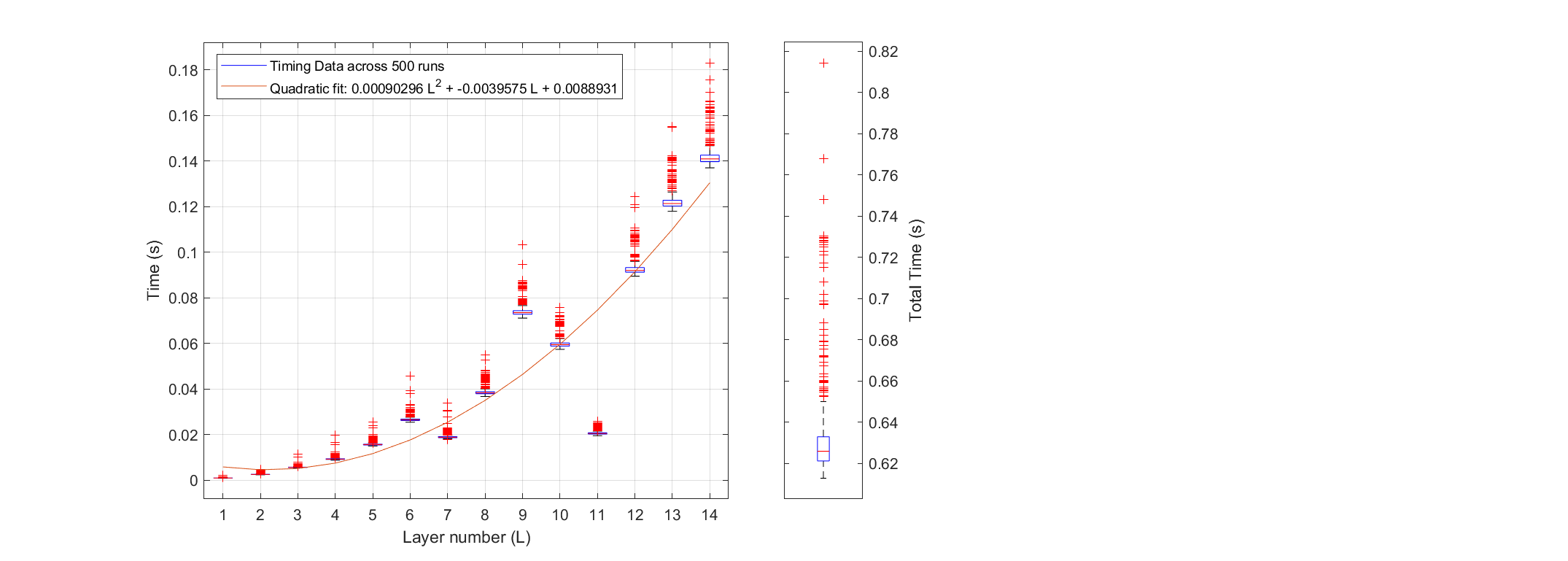}
    \caption{Timing results to complete each new layer of the tree. As the number of layers increases, the algorithm must traverse more nodes. However, we maintain the quadratic complexity of the tree expansion in the timing results. Some satellites are very informative and require many intersections yielding larger times to complete the layer (e.g., layer number 9). Other satellites are not as informative and predominantly update the probabilities (e.g., layer number 11). 
    }
    \label{fig:tree_timing}
\end{figure}

Figure~\ref{fig:tree_timing} illustrates the time to complete each tree expansion as the tree depth increases, wherein our proposed algorithm produces the full tree and the mosaic in a low median time of $0.63~$\si{s}. 
As observed in Figure~\ref{subfig:leaf_analysis_num_leaves}, the shadows from the 7th and 11th satellites were not as informative as other satellites, such as the 9th satellite. 
We see this trend repeat in the timing analysis.
When expanding to layers 7 and 11, we identify the redundant information with the 3rd satellite very early in the tree. So, we compute fewer intersections and instead quickly recursively propagate the LOS and NLOS scores to the leaves (Algorithm~\ref{alg:tree_expansion}). 
When expanding to layer 9, the satellite shadow is highly informative and we can split many leaf nodes. 
Therefore, we produce many more leaves and we incur higher timing costs in computing the intersections. 
By design, our algorithm spends more time on the more informative signals that improve the risk-aware localization and much less time on signals that have less impact on the risk-aware localization. 



\section{Conclusion} \label{sec:conclusions_and_future_work}

We proposed a novel Mosaic Zonotope Shadow Matching (MZSM) algorithm that advanced shadow matching to enable risk-aware 3DMA-GNSS localization. 
We specifically targeted two risk-aware objectives: 
(1)~formulate precise models of the shadow matching uncertainty in a computationally efficient manner while tractably scaling with increasing numbers of available signals and
(2)~provide guarantees on the uncertainty bounds of the urban localization solution across a vast space of AI-driven LOS classifier designs.
At a high level,
we met these objectives through set-based representations of the user location, set-based uncertainty bounds on the user location, and efficient tree-based construction methods for organizing the set-based information. 
We started with the exact set-based representations of the GNSS shadows from our prior work \cite{bhamidipati2021set}, where our set-based shadow extraction method avoids the trade-off between discretization and computational load. 
We expanded upon this set-based shadow matching paradigm to produce a shadow matching mosaic. 
We recursively expanded a binary tree
that exhaustively considered all possible LOS and NLOS misclassifications.
As the tree expands, we refined the AOI (i.e., the coarse initial estimate) into smaller polytopes with associated probability masses.
We used special cases of overlap between the GNSS shadow and tree node polytope to design a memory-efficient binary tree that grows quadratically with the number of satellites rather than the exponential growth of a full binary tree.
With a probability mass function over the tree's leaves, we incorporated self-monitoring into the tree to check the AOI violation probability. 
Finally, we derived confidence collections that provide guarantees on the uncertainty bounds of the urban localization solution.

We validated that our MZSM algorithm meets our risk-aware objectives via our high-fidelity experiments. 
First, we demonstrated that we can assess the expectation of the shadow matching mosaic with data on the LOS/NLOS classifier's true positive and true negative rates (Figure~\ref{fig:sensitivity_mosiac}). 
We discussed how our classifier posterior method can enable a system designer to ascertain the benefits of different classifiers and enable AI decision-making agents to forecast how the shadow matching mosaic could evolve along a trajectory.
We employed our classifier posterior method in the role of a system designer to perform a sensitivity analysis of the expectation of our shadow matching mosaic across classifier accuracy. 
Second, we demonstrated that the Gaussian uncertainty models, such as Gaussian mixture models (GMMs), can capture 
the multimodality in the shadow matching mosaic, but they cannot provide sufficient approximation of the uncertainty in the exact shadow matching mosaic (Figure~\ref{fig:gmm_distributions}).
Quantitatively, we observed that the integrated percent error between the GMMs and the mosaic is 50\% or worse (Figure~\ref{fig:gmm_integral}), which is far too high for a risk-aware approximation.
Such high approximation error impedes downstream AI agents from making risk-aware decisions and would yield vulnerability in safety-critical applications.
Hence, our exact set-based model of the uncertainty advances shadow matching to incorporate risk awareness. 
Third, we demonstrated that grid-based shadow matching approaches can provide meaningful approximations of the uncertainty at high grid resolutions (Figure~\ref{fig:cc_single_mosiac} and \ref{fig:cc_gbsm_grid_converge}). 
Quantitatively, we used the intersection over union (IOU) metric to compute the similarity in the overlap between uncertainty bounds from our MZSM method and existing work. 
We found that the IOU metric is between 0.5 and 0.8 for most confidence levels for typical AI-driven classifiers with a 3-meter grid resolution (Figure~\ref{fig:cc_iou_converge}). 
Reaching better overlap between the grid-based approximation and exact set-based mosaic would require much finer grid resolutions. 
However, such fine grid resolution would not be computationally efficient. 
Hence, our exact set-based model is better suited to provide a risk-aware understanding of the uncertainty in a computationally efficient matter than existing methods.
Lastly, we demonstrated the self-monitoring, tractability, and timing of our algorithm. 
We showed how our algorithm accurately estimates the self-monitoring violation probability to be a low value of 0.001 for a near-perfect classifier and a high value of 0.988 for an uninformative classifier (Figure~\ref{fig:violation_probability}). 
Furthermore, we validated that our tree grows quadratically with the number of satellites with a low quadratic coefficient of 1.2 (Figure~\ref{fig:leaf_analysis}).
We completed the tree in a median of 0.63 seconds with the 3rd quartile at 0.65 seconds for a case with 14 shadows (Figure~\ref{fig:tree_timing}). 
Altogether, we succeeded in meeting our objectives and advancing shadow matching for risk-aware 3DMA-GNSS that is well suited for the safety of autonomous systems in harsh urban environments.
At a macroscopic level, we provide an instance where we formally connect the uncertainty in AI-tools for localization to the state uncertainty information needed for AI decision-making agents. 
At least in the domain of 3DMA-GNSS, we show the value of certifiable and risk-aware localization in enabling risk-aware autonomous systems.


\section{Acknowledgments}

This material is based upon work supported by the National Science Foundation under Grant No.~DGE-1656518. 
The authors would like to thank Adyasha Mohanty for reviewing drafts of this paper. 
We would further like to thank the rest of the Stanford Navigation and Autonomous Vehicles laboratory, Shreyas Kousik, and Erik Herrera for insightful discussions and feedback.

\appendix
\section{: Proof of Proposition~\ref{theo:propositionA}} \label{app:AppendixA}
\renewcommand{\theequation}{A.\arabic{equation}}
\setcounter{equation}{0}

\begin{proof}
First, we prove that the cases are distinct assuming that $\nodepoly{j-1,n}\neq \emptyset$ and $\shadowVertex{j} \neq \emptyset$. The `\nooverlap' case defined in Eq.~\eqref{eq:no_overlap} is the only case that satisfies $\nodepoly{j-1,n} \cap \shadowVertex{j} = \emptyset$, so it is distinct from the rest. Similarly, the `\fulloverlap' case definied in Eq.~\eqref{eq:full_overlap} is the only case that satisfies $\nodepoly{j-1,n} \cap \shadowVertex{j} = \nodepoly{j-1,n}$, so it is distinct from the rest. Lastly, the `\partoverlap' case is necessarily distinct from the `\nooverlap' case via $\nodepoly{j-1,n} \cap \shadowVertex{j} \neq \emptyset$ and necessarily distinct from the `\fulloverlap' case via $\nodepoly{j-1,n} \cap \shadowVertex{j} \neq \nodepoly{j-1,n}$, per Eq.~\eqref{eq:partial_overlap}. Therefore, the three cases are distinct, as desired. Note that if $\nodepoly{j-1,n} = \emptyset$ or $\shadowVertex{j} = \emptyset$, then we have degeneracy. 



Second, we prove that the cases are complete, i.e., that the union over all cases covers all possibilities. Without loss of generality (since unions are commutative), consider the union of the `\nooverlap' case and the `\partoverlap' case:
\begin{align}
    \big(\nodepoly{j-1,n} \cap \shadowVertex{j} = \emptyset \big) \cup \big((\nodepoly{j-1,n} \cap \shadowVertex{j} \neq \emptyset) \cap (\nodepoly{j-1,n} \cap \shadowVertex{j} \neq \nodepoly{j-1,n}) \big) &= \\
    \big((\nodepoly{j-1,n} \cap \shadowVertex{j} = \emptyset) \cup (\nodepoly{j-1,n} \cap \shadowVertex{j} \neq \emptyset) \big) \cap \big((\nodepoly{j-1,n} \cap \shadowVertex{j} = \emptyset) \cup  (\nodepoly{j-1,n} \cap \shadowVertex{j} \neq \nodepoly{j-1,n}) \big) &= \\
    \big((\nodepoly{j-1,n} \cap \shadowVertex{j} = \emptyset) \cup  (\nodepoly{j-1,n} \cap \shadowVertex{j} \neq \nodepoly{j-1,n}) \big) &= \\
    (\nodepoly{j-1,n} \cap \shadowVertex{j} \neq \nodepoly{j-1,n}) &
\end{align}
Now take the union between the above and the `\fulloverlap' case. Simply,
\begin{equation}
    (\nodepoly{j-1,n} \cap \shadowVertex{j} \neq \nodepoly{j-1,n}) \cup (\nodepoly{j-1,n} \cap \shadowVertex{j} = \nodepoly{j-1,n})
\end{equation}
is a tautology. Therefore, the cases are complete, as desired. 
\end{proof}

\section{: Proof of Proposition~\ref{theo:propositionB}} \label{app:AppendixB}
\renewcommand{\theequation}{B.\arabic{equation}}
\setcounter{equation}{0}

\begin{proof}
Our binary search tree consists of branching based on two operations: set subtraction and set intersection. 
Set subtraction is an instance of set intersection. 
Namely, for given sets $A$ and $B$, $A - B = A \cap B^\complement$, where $B^\complement$ denotes the complement of $B$. 
Furthermore, the intersection of sets is commutative: for sets $A$, $B$, and $C$, $A \cap B \cap C = A \cap C \cap B= C \cap A \cap B=\cdots$, etc. Vertex-represented polytopes are closed under set intersection, so these intersection commutativity properties holds for polytopes. 

As seen in Eq.~\eqref{eq:intersectionchain}, we can unravel the recursion to illustrate that any node polytope $\nodepoly{j,n}$ is computed via a chain of set intersections. 
\begin{equation}
    \nodepoly{j,n} = \AOIvertex \cap \shadowVertex{1}^{\ (\complement)} \cap \shadowVertex{2}^{\ (\complement)} \cap \cdots \cap \shadowVertex{j-1}^{\ (\complement)} \cap \shadowVertex{j}^{\ (\complement)} \label{eq:intersectionchain}
\end{equation}
where $^{(\complement)}$ toggles complement for the LOS or NLOS cases per Eqs.~\eqref{eq:loschild} and \eqref{eq:nloschild}. Unfurling the recursive structure means that a position node polytope defined via above Eq.~\eqref{eq:intersectionchain} has a probability score given by Eq.~\eqref{eq:chainprob}.
\begin{equation}
    p(\nodepoly{j}) = p(\AOIvertex) \cdot p(s_1 = (\los)^{(\complement)}) \cdot p(s_2 = (\los)^{(\complement)}) \cdots p(s_{j-1} = (\los)^{(\complement)}) \cdot p(s_j = (\los)^{(\complement)}) \label{eq:chainprob}
\end{equation}
Suppose we have $k$ total GNSS shadows. 
By intersection commutativity, we can equivalently construct the leaf node by any of the following intersection chains
\begin{align}
    \nodeleaf &= \AOIvertex \cap \shadowVertex{1}^{\ (\complement)} \cap \shadowVertex{2}^{\ (\complement)} \cap \cdots \cap \shadowVertex{(k - 1)}^{\ (\complement)} \cap \shadowVertex{k}^{\ (\complement)} \\
    &= \AOIvertex \cap \shadowVertex{k}^{\ (\complement)} \cap \shadowVertex{1}^{\ (\complement)} \cap \cdots \cap \shadowVertex{k-1}^{\ (\complement)} \cap \shadowVertex{2}^{\ (\complement)} \\
    &= \shadowVertex{k}^{\ (\complement)} \cap \shadowVertex{1}^{\ (\complement)} \cap \cdots \cap \shadowVertex{k-1}^{\ (\complement)} \cap \AOIvertex \cap  \shadowVertex{2}^{\ (\complement)} \\
    &= \text{any shuffling thereof} \nonumber
\end{align}
This holds for all the leaves. Consider the fully expanded $2^n$ binary tree at depth $n\sat$. Each leaf has a unique branching path and, therefore, a unique identification via the complement toggle ($^{(\complement)}$) and AOI. When we change the ordering of the satellites or AOI, we simply change the order in which we perform the intersections. We maintain a bijection between the original ordering and the new order such that there is a unique mapping between a leaf in the original ordering and the equivalent leaf in the new ordering. Therefore, the fully expanded $2^n$ binary tree is commutative. As an immediate corollary, if any leaf is empty in the fully expanded $2^n$ binary tree, it is empty in all satellite orderings. 

Next, consider our memory-efficient binary tree. Via the same bijection as the fully expanded $2^n$ binary tree, there is a unique mapping between a (non-empty) leaf in the original ordering and the equivalent (also non-empty) leaf in the new ordering. The only difference is that the identification is compressed. In particular, the `\nooverlap' or `\fulloverlap' cases compress the intersection chain. In the `\nooverlap' case,  the $j$-th GNSS shadow ($\shadowVertex{j}$) does not intersect with the prior chain of length $j$ (where we include the AOI in the chain). Hence, the \texttt{LOS Child} takes the compressed identification of the parent. Combining Eq.~\eqref{eq:no_overlap}, Eq.~\eqref{eq:loschild}, and Eq.~\eqref{eq:intersectionchain} yields
\begin{align}
    \LOSchildnode{j,m} &= \parentnode{j-1,n} \cap \shadowVertex{j}^\complement \\
    \LOSchildnode{j,m} &= \big(\AOIvertex \cap \shadowVertex{1}^{\ (\complement)} \cap \shadowVertex{2}^{\ (\complement)} \cap \cdots \cap \shadowVertex{j-1}^{\ (\complement)} \big) \cap \shadowVertex{j}^{\complement} \\
    \LOSchildnode{j,m} &= \big(\AOIvertex \cap \shadowVertex{1}^{\ (\complement)} \cap \shadowVertex{2}^{\ (\complement)} \cap \cdots \cap \shadowVertex{j-1}^{\ (\complement)} \big) = \parentnode{j-1,n}
\end{align}
The equivalent holds for the `\fulloverlap' case and the \texttt{NLOS Child}. Importantly, the compressed version results in the same polytope at the leaf. Therefore, we have a bijection between the (non-empty) leaves in the memory-efficient binary tree and the equivalent (non-empty) leaves in the fully expanded $2^n$ binary tree. We then also have a bijection between leaves across different satellite ordering for the memory-efficient binary tree. Therefore, since vertex-represented polytopes are closed under set intersection and set intersection is commutative, the leaf polytopes generated from our binary tree are the same for any ordering of the AOI and GNSS shadows. 
\end{proof}

\section{: Proof of Proposition~\ref{theo:propositionC}} \label{app:AppendixC}
\renewcommand{\theequation}{C.\arabic{equation}}
\setcounter{equation}{0}

\begin{proof}
We prove the statement via induction. 
The base case is simply the tree's root (i.e., $0$-th node at $0$-th tree level), which matches the initial AOI. 
We consider the following inductive hypothesis: if the union of the leaf polytopes at the $j$-th tree expansion matches the AOI polytope~$\AOIvertex$, then the union of the leaf polytopes at the $(j+1)$-th tree expansion also matches the AOI polytope~$\AOIvertex$.
By branching the $j$-th tree level to the next $(j+1)$-th tree level, we split each parent node polytope into a maximum of two children nodes: a \texttt{LOS child} (Eq.~\ref{eq:loschild}) and an \texttt{NLOS child} (Eq.~\ref{eq:nloschild}). 
Irrespective of the overlap case, we observe the union of the children node polytopes equal the parent node polytope, which is shown as follows:
\begin{align}
    \text{\nooverlap:} \quad \LOSchildnode{j+1,m} \cup \NLOSchildnode{j+1,l} &= \parentnode{j,n} \cup \emptyset = \parentnode{j,n} \\
    \text{\partoverlap:} \quad \LOSchildnode{j+1,m} \cup \NLOSchildnode{j+1,l} &=\Big(\parentnode{j,n} \cap \shadowVertex{j+1}^c\Big) \cup \Big(\parentnode{j,n} \cap \shadowVertex{j+1}\Big)  = \parentnode{j,n} \\
    \text{\fulloverlap:} \quad \LOSchildnode{j+1,m} \cup \NLOSchildnode{j+1,l} & = \emptyset \cup \parentnode{j,n} = \parentnode{j,n}  
\end{align}
Note that we use the distribute law to analyze the \partoverlaplower \ case. Namely, for polytopes $A$, $B$, and $C$, $(A \cap B)\cup (A \cap C) = A \cap (B \cup C)$.
This result, that the union of the children polytopes matches the parent polytope, holds for any tree expansion since these overlap cases are distinct and complete (Propostition~\ref{theo:propositionA}). 
Moreover, the empty leaves do not geometrically contribute to completing the mosaic. 
By commutativity of union, if the union of the leaf polytopes at the $j$-th tree expansion matches the AOI polytope~$\AOIvertex$ and the union of the children polytopes matches the parent polytopes for all tree expansions, then it must be the case that the union of the leaf polytopes at the $(j+1)$-th tree expansion also matches the AOI polytope~$\AOIvertex$. Therefore, by induction, the union of the leaf polytopes matches the area of interest for any number of shadows (equivalently, any number of satellites).
\end{proof}

\section{: Proof of Proposition~\ref{theo:propositionD}} \label{app:AppendixD}
\renewcommand{\theequation}{D.\arabic{equation}}
\setcounter{equation}{0}

\begin{proof}

First, we explicitly write how we condition $\gmmk(x, y)$ on the AOI:
\begin{align}
    \gmmk(x, y) &= \widetilde{\gmmk}(x, y) \left(\iint_{\text{AOI}} \widetilde{\gmmk}(x, y) \ dx \  dy \right)^{-1}
\end{align}
\noindent where $\widetilde{\gmmk}(x, y)$ is the Gaussian mixture model with infinite support. We do the same with $\mzsm(x, y)$ and $\widetilde{\mzsm}(x, y)$ via Eq.~\eqref{eq:mzsm_pdf}. 

Now, we prove $0 \leq \percenterror \leq 2$. For the lower bound, notice $\gmmk(x, y) > 0$ and $\mzsm(x, y) \geq 0$. 
Therefore, $\percenterror \geq 0$. Equivalently, this lower bound holds from the reverse triangle identity. 
For the upper bound, we use the triangle inequality directly,
\begin{align}
    |\gmmk(x, y) - \mzsm(x, y)| &\leq |\gmmk(x, y)| + |-\mzsm(x, y)| \\
    \iint_{\text{AOI}} |\gmmk(x, y) - \mzsm(x, y)| \ dx \  dy &\leq \iint_{\text{AOI}} |\gmmk(x, y)| \ dx \ dy + \iint_{\text{AOI}} |\mzsm(x, y)| \ dx \ dy \\
    \iint_{\text{AOI}} |\gmmk(x, y) - \mzsm(x, y)| \ dx \  dy &\leq \iint_{\text{AOI}} \gmmk(x, y) \ dx \ dy + \iint_{\text{AOI}} \mzsm(x, y) \ dx \ dy \\
    \iint_{\text{AOI}} |\gmmk(x, y) - \mzsm(x, y)| \ dx \  dy &\leq 1 + 1 \\
    \percenterror &\leq 2
\end{align}

Hence, $0 \leq \percenterror \leq 2$, as desired.
\end{proof}


\bibliography{bibs/mybibfile, bibs/IUP-UCL-groves-adjrad, bibs/SM-UCL-wang, bibs/Classifier-HK-xu-ng-lita, bibs/other-bib, bibs/raytracing}

\end{document}